\documentclass{article} 
\usepackage[preprint]{colm2025_conference}
\usepackage{graphicx}
\usepackage{microtype}
\usepackage{hyperref}
\usepackage{url}
\usepackage{booktabs}

\usepackage{lineno}
\usepackage{enumitem}
\usepackage{amssymb}
\definecolor{darkblue}{rgb}{0, 0, 0.5}
\hypersetup{colorlinks=true, citecolor=darkblue, linkcolor=darkblue, urlcolor=darkblue}

\usepackage{graphicx}               
\usepackage{subcaption}  
\usepackage{xspace}
\usepackage{tcolorbox}

\title{Inference-Time Scaling for Complex Tasks:\\ Where We Stand and What Lies Ahead} 

\author{Vidhisha Balachandran\quad
 Jingya Chen\quad
 Lingjiao Chen \quad
Shivam Garg \quad \\
\bfseries{Neel Joshi} \quad
\bfseries{Yash Lara} \quad
\bfseries{John Langford} \quad
\bfseries{Besmira Nushi} \quad \\
\bfseries{Vibhav Vineet} \quad
\bfseries{Yue Wu} \quad
\bfseries{Safoora Yousefi}
  \vspace{0.4em}
    \\
  {Microsoft Research}
\vspace{0.4em}
  \\
  {\small 
  \includegraphics[width=1.2em, trim=0 0 0 0, clip]{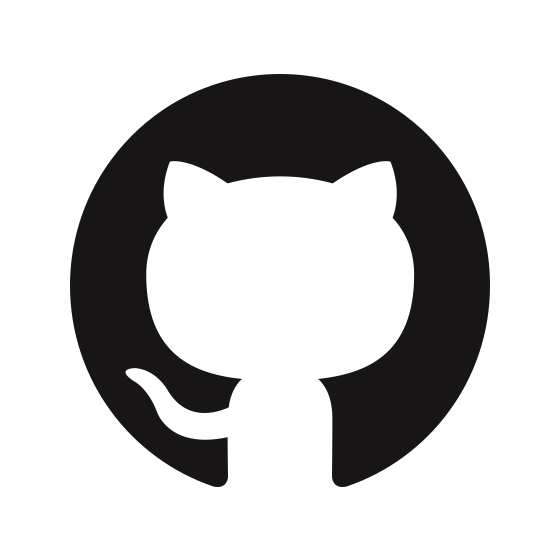} Code: \texttt{\url{https://github.com/microsoft/eureka-ml-insights}}
  } 
}

\usepackage{xcolor}
\definecolor{cadmiumgreen}{rgb}{0.2, 0.82, 0.24}
\definecolor{forestgreen}{rgb}{0.13, 0.55, 0.13}
\definecolor{redfrontier}{HTML}{B50575}
\definecolor{greenfrontier}{HTML}{188814}
\definecolor{bluefrontier}{HTML}{4C66DD}

\newcommand{\ClaudeSonnet}{{Claude 3.5 Sonnet}\xspace}  
\newcommand{\ClaudeSonnetThinking}{{Claude 3.7 Sonnet}\xspace}  
\newcommand{\ROne}{{DeepSeek R1}\xspace} 
\newcommand{\GeminiPro}{{Gemini 2.0 Pro}\xspace}  
\newcommand{\GeminiFlash}{{Gemini 2 Flash Thinking}\xspace}  
\newcommand{\LlamaThreeOneLarge}{{Llama 3.1 405B}\xspace} 
\newcommand{\GPTFourO}{{GPT-4o}\xspace}  
\newcommand{\OOne}{{O1}\xspace}  
\newcommand{\OThree}{{O3-mini}\xspace}  

\begin{document}

\ifcolmsubmission
\linenumbers
\fi

\maketitle

\begin{abstract}
Inference-time scaling can enhance the reasoning capabilities of large language models (LLMs) on complex problems that benefit from step-by-step problem solving. Although lengthening generated scratchpads has proven effective for mathematical tasks, the broader impact of this approach on other tasks remains less clear. In this work, we investigate the benefits and limitations of scaling methods across nine state-of-the-art models and eight challenging tasks, including math and STEM reasoning, calendar planning, NP-hard problems, navigation, and spatial reasoning. We compare conventional models (e.g., GPT-4o) with models fine-tuned for inference-time scaling (e.g., o1) through evaluation protocols that involve repeated model calls, either independently or sequentially with feedback. These evaluations approximate lower and upper performance bounds and potential for future performance improvements for each model, whether through enhanced training or multi-model inference systems. Our extensive empirical analysis reveals that the advantages of inference-time scaling vary across tasks and diminish as problem complexity increases. In addition, simply using more tokens does not necessarily translate to higher accuracy in these challenging regimes. Results from multiple independent runs with conventional models using perfect verifiers show that, for some tasks, these models can achieve performance close to the average performance of today’s most advanced reasoning models. However, for other tasks, a significant performance gap remains, even in very high scaling regimes. Encouragingly, all models demonstrate significant gains when inference is further scaled with perfect verifiers or strong feedback, suggesting ample potential for future improvements.

\end{abstract}
\section{Introduction}
Inference-time scaling refers to allocating increasing computational resources during inference of machine learning models to enhance their performance on complex tasks. Recently, this approach has encompassed post-training techniques that encourage models to generate longer and step-by-step solutions, explore different alternatives at each step, or even backtrack to previous steps when an inference path does not appear promising. 
Several models to date~\citep{O3mini,jaech2024openai,Claude37Sonnet,guo2025deepseek,GeminiFlash} exhibit one or more aspects of such desirable behavior at inference time and improve the state of the art on complex tasks. While the exact training techniques and data that enabled major model releases are not always shared, earlier studies and replication surveys~\citep{lightman2023let,wang2023math,zelikman2022star} as well as open source releases~\citep{guo2025deepseek,wang2024openr}, introduce techniques for lengthening generation traces via strong verifiers, self-reflection, chain-of-thought finetuning and reinforcement learning (RL). 

These recipes have shown to be effective for math problems, which remain the main testbed for understanding the impact of inference-time scaling. In this work, we present an extensive empirical analysis of inference-time scaling for complex tasks, that studies both conventional models and reasoning models (i.e. models tuned for inference-time scaling), and measures their current abilities and future potential, if inference were to be scaled further. 

\begin{figure}
    \centering
    \includegraphics[width=0.5\linewidth]{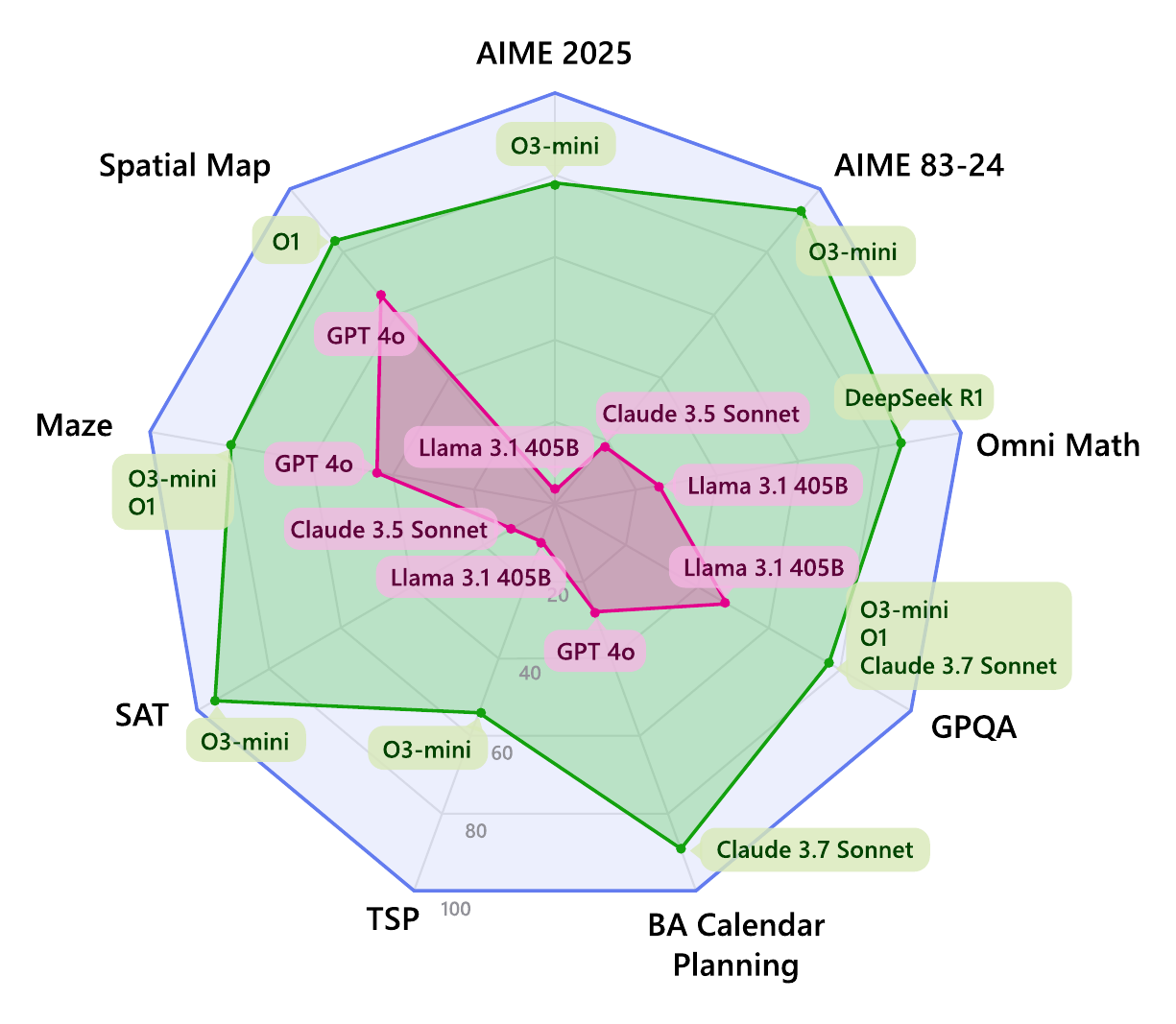}
    \caption{Performance of best and worse models on different reasoning benchmarks. The \textcolor{redfrontier}{red} frontier shows the performance of the worse model. The \textcolor{greenfrontier}{green} frontier shows the performance of the best model, indicating the best known result with current technology. The \textcolor{bluefrontier}{blue} horizon between the best model and the maximum performance shows the room for improvement for mastering the capability. The best performance sets indicated in the green border include all models that perform within 2\% of the best observed result.}
    \label{fig:summary}
\end{figure}

First, we present a comprehensive study of reasoning capabilities of nine state-of-the-art foundation models for a rich set of tasks. 
We utilize existing open source benchmarks for evaluating problems on math and STEM reasoning, calendar planning, navigation, and spatial reasoning and introduce two new benchmarks for evaluating NP-hard problems. We present performance-cost tradeoffs as well as common failure patterns within and across benchmarks, beyond single-score measurements and leaderboards.
Through our analysis, we find that, while all the chosen tasks can benefit from scratchpad-like, step-by-step problem solving (often referred to as reasoning), this paradigm i) does not serve all domains and tasks equally, and ii) improvements diminish with increased problem difficulty. Surprisingly, we also observe that longer generations relative to the same model can sometimes be an indicator of models struggling, rather than improved reflection. Similarly, when comparing different reasoning models, higher token usage is not always associated with better accuracy. These findings motivate the need for more purposeful and cost-effective scaling approaches.

Further, we simulate two different types of inference-time scaling approaches. \emph{Independent parallel generations} sample multiple answers from the same model at a high temperature, and then aggregate to obtain a final result through different operators. If a model reacts positively to such an approach, it shows that there exists at least one correct inference path and that there is still space for improving the models (including those tuned for inference-time scaling) through stronger verification methods. Another lens on these results is to use them for estimating model reliability and variance across different attempts, or its expected performance in real-world pipelines that implement redundant answer sampling. 

\emph{Sequential generations} iteratively leverage the feedback of the same model when the original model's answer is incorrect, and pass that feedback to the model under test to give it another opportunity to improve its answer. This setup helps understand the model's ability to leverage feedback and also its potential for being involved in generating synthetic data for fine-tuning or RL techniques that may be used offline or online for improving the same model or another weaker model~\citep{gulcehre2023reinforced,hosseini2024v}. Results of these simulations measure the potential and limitations of models for further improvement and estimate the possible benefits of future training and RL techniques for improved reasoning.  

We summarize our top findings below 
\ifcolmsubmission
\footnote{Reusable implementations of the benchmarks and scaling approaches will be made available upon publication, including data, code, and evaluation logs.}
\fi
\ifcolmfinal
\footnote{Reusable implementations of the benchmarks and scaling approaches are available at \url{https://github.com/microsoft/eureka-ml-insights}, including data, code, and evaluation logs.}
\fi
\ifcolmpreprint
\footnote{Reusable implementations of the benchmarks and scaling approaches will be made available at \url{https://github.com/microsoft/eureka-ml-insights}, including data, code, and evaluation logs.}
\fi:
\begin{enumerate}[leftmargin=*, itemsep=0em]
    \item All studied tasks benefit from using models trained for scaling inference-time compute. Although inference-time scaling improves performance, its effectiveness varies between domains and tasks, with diminishing returns as task complexity increases.
    \item There is high variability in token use, even across models with similar accuracies on a task, indicating space for improving token efficiency and that higher token consumption does not indicate higher accuracy across models.
    Repeated queries to the same model can yield highly variable token usage, introducing cost nondeterminism for developers and users - even when the model consistently provides correct answers.
    \item Continued scaling with perfect verifiers consistently improves performance across benchmarks for both reasoning and conventional models, indicating further potential for model improvement. This emphasizes the importance of building improved and generalizable verifiers that can be used for further development.
    \item Experiments with superscaling (up to 50× more inference calls) further improve performance across reasoning and conventional models. Conventional models can leverage this additional computation to approach reasoning model performance in some cases, although gains diminish in highly complex settings. 
\end{enumerate}

\section{Benchmarks and Methods}
\label{benchmeth}
\paragraph{Inference-time scaling approaches and aggregators.}Throughout this paper we focus on three test-time scaling approaches. The first is the \emph{standard CoT} approach, which simply asks a model to answer a question in a step-by-step fashion. The second approach is the \emph{parallel scaling method}: for each question, we independently sample $N$ generations from a model, and then use an \emph{aggregator} to extract the final answer from these candidates (e.g., majority vote, average, best-of-n etc.). Finally, the \emph{sequential scaling} approach iteratively generates an answer and asks the model to refine its answer via feedback provided by a \emph{critic}. 
 
A key question is how to instantiate \emph{aggregator} and \emph{critic}. We consider four common instances of the aggregator, namely, average, majority vote, best-of-n and worst-of-n, which return the average, the mode, the best and worst answers from the candidate answers, respectively. The last two aggregators measure upper and lower bounds on model performance.

For the critic, we use a hybrid approach: the critic knows the ground-truth, and then uses it to offer textual feedback about the latest solution without revealing the ground truth. In our simulations, the same model is used to critique its own answer (i.e. self-critique), although other settings and combinations are also interesting to explore.

\paragraph{Benefits of evaluating inference-time scaling.} There are several reasons why a deeper analysis using the above scaling approaches is important. First, comparing the average performance of models across a diverse set of reasoning tasks, enables a broader perspective on how well the current training methods for reasoning generalize to different types of reasoning. Evaluating best-of-n performance for conventional models approximate an upper bound on the potential of these models to be adapted as reasoning models by lengthening their generations via simple verifier-in-the-loop RL or fine-tuning methods that teach the model how to pick the best answer from a set of candidates. We specifically study the gap between the best-of-n performance for conventional models and the average performance of reasoning models, which we refer to as the \emph{conventional-to-reasoning gap}. This serves as an estimate of the gap that needs to be addressed either via sampling beyond N candidates or via more sophisticated RL that introduces feedback and backtracking in a more fine-grained manner, rather than at the end of a generation. Estimates of best-of-n performance for reasoning models demonstrate the untapped potential of current methods, showing that better inference paths in such models are still possible but need to be better extracted to serve the best possible reasoning capability. 

\paragraph{Evaluation metrics.} Compared to standard inference, test-time scaling aims at improving performance with additional computation at test time (i.e. longer generations). Therefore, our evaluation metrics include both the performance accuracy and the amount of computation in terms of the number of tokens generated, including both completion and reasoning tokens. Associating accuracy with token usage of inference-time scaling approaches portrays the Pareto trade-off between accuracy and compute as an assessment of token efficiency. Unless otherwise specified, for all benchmarks, accuracy is defined as how often a given scaling approach leads to the correct answer. 

\paragraph{Models and data sourcing.} In this study, we work with four conventional models (\ClaudeSonnet, \GeminiPro, \GPTFourO, \LlamaThreeOneLarge) and five models tuned for inference-time scaling (\ClaudeSonnetThinking, \ROne, \GeminiFlash, \OOne, \OThree), therefore providing guidance for practitioners that aspire to tune their current models for better reasoning capabilities or those interested in opportunities to extend the state-of-the-art. Table~\ref{tab:models} lists all models and their corresponding sampling parameters used at test time.

We aim at studying a diverse set of complex problems that could potentially benefit from step-by-step solutions and extended scratchpads. Among these benchmarks, AIME and GPQA Diamond are most commonly used in recent technical reports associating major model releases~\citep{O3mini,jaech2024openai}. AIME is a set of problems from the American Invitational Mathematics Examinations, held yearly from 1983 to 2025. GPQA consists of graduate-level problems written by domain experts in biology, physics, and chemistry. Although we include these two benchmarks in this study for comparability with recent studies, we acknowledge that evaluating only these two sources is insufficient for studying the various aspects of reasoning. Besides, given the high popularity of these benchmarks, it is important to evaluate other data sources and problem types, to investigate generalization properties of current models to other algorithmic and planning problems, tasks that require spatial reasoning, or a broader range of math problems.

Therefore, we also experiment with six additional benchmarks shown in Table~\ref{tab:datasets}. Omni-MATH~\citep{gao2024omni} is a large collection of over 4000 olympiad-level math problems with rigorous human annotation, offering a diversity of mathematical topics and difficulty levels, as well as open-ended problems. 3SAT (3-literal Satisfiability Problem) and TSP (Traveling Salesman Problem) are new benchmarks\footnote{The benchmarks and respective code for data generation will be open sourced upon publication.} that this work contributes for studying the ability of models to solve NP-hard problems~\citep{papadimitriou2003computational,hartmanis1982computers}. To create these benchmarks, we synthetically generate controlled questions on different difficulty levels and compute exact solutions for them (see Appendix~\ref{sec:tsp} and~\ref{sec:3st} for more details). In 3SAT, each clause contains three binary literals (variables), the difficulty level corresponds to the ratio of clauses to variables, and the model is tasked to search for a valid assignment. In TSP, the generated graphs are fully connected with positive weights only, the difficulty level corresponds to the number of nodes in the graph, and the model is tasked to find an optimal minimal path. 
BA-Calendar is a calendar planning task~\citep{butt2024benchagents} that requires models to find a common time slot among participants while considering constraints beyond availability, such as time zones, buffer time, priority, etc. Difficulty level in BA-Calendar corresponds to \emph{constrainedness}, which is defined as the complement of the ratio of feasible slots to total slots. The availability of difficulty tags for Omni-MATH, TSP, 3SAT, and BA-Calendar enables us to analyze how accuracy and token usage scale with difficulty  in inference-time scaling, which is a perspective that is still underexplored.

Finally, Maze and SpatialMap are two benchmarks that test for navigation and spatial reasoning skills~\citep{wang2024picture}. Maze consists of multiple choice questions regarding a given maze (see Figure~\ref{fig:vl_maze})\footnote{We use the 10x10 maze version of the benchmark.}. The questions include counting the number of turns or determining the spatial relationships between two points in the maze. SpatialMap tests for spatial reasoning (see Figure~\ref{fig:vl_spatialmap}) by first introducing a set of objects with unique names, providing a set of pairwise relationships between those objects (e.g., A is to the southeast of B), and asking about the spatial relationships between two objects (which were not directly mentioned in context) or the number of objects that meet certain spatial criteria. 

Several of the above benchmarks (TSP, 3SAT, BA-Calendar, Maze, SpatialMap) are procedurally generated, offering the possibility to generate new or more difficult versions of them in the future to address concerns on benchmark memorization or saturation. 

\begin{table}[t]
  \centering
  \scriptsize
  \caption{List of benchmarks.}
    \begin{tabular}{lrccrrr}
    \toprule
    \bfseries{Benchmark} & \bfseries{\#prompts} & \bfseries{Domain} & \multicolumn{1}{c}{\bfseries{Answer space}} & \bfseries{Results}\\
    \midrule
    AIME 25, 83-24~\citep{AIME25,AIME8324} & 30, 933    & Math & \multicolumn{1}{c}{integer} & Appendix~\ref{sec:aime} \\
    \hline
    Omni-MATH~\citep{gao2024omni} & 4428    & Math  & \multicolumn{1}{c}{open ended} & Appendix~\ref{sec:omnimath} \\
    \hline
    GPQA$\mathbin{\Diamond}$~\citep{rein2024gpqa} & 198    & Natural Sciences & \multicolumn{1}{c}{mult. choice} & Appendix~\ref{sec:gpqa} \\
    \hline
    3SAT-Search (new benchmark)    &    800   &   NP-hard   &   \multicolumn{1}{c}{open ended} & Appendix~\ref{sec:3st} \\
    \hline
    TSP-Opt (new benchmark)    &   960    &  NP-hard    &   \multicolumn{1}{c}{open ended} & Appendix~\ref{sec:tsp} \\
    \hline
    BA-Calendar~\citep{butt2024benchagents}    &   2000    &  Planning    &  \multicolumn{1}{c}{open ended} & Appendix~\ref{sec:bacalendar} \\
    \hline
    Maze~\citep{wang2024picture}     &    1500   &  Navigation    &  \multicolumn{1}{c}{mult. choice} & Appendix~\ref{sec:maze} \\
    \hline
    SpatialMap~\citep{wang2024picture}     &    1500   &   \multicolumn{1}{c}{Spatial Reasoning}  &  \multicolumn{1}{c}{mult. choice} & Appendix~\ref{sec:spatialmap} \\
    \bottomrule
    \end{tabular}%
  \label{tab:datasets}%
\end{table}%

\section{Experiments and Findings}
\paragraph{Model performance and generalization.}
\begin{figure}[t]
    \centering
    \begin{subfigure}[b]{0.24\textwidth}
        \centering
        \includegraphics[width=\textwidth]{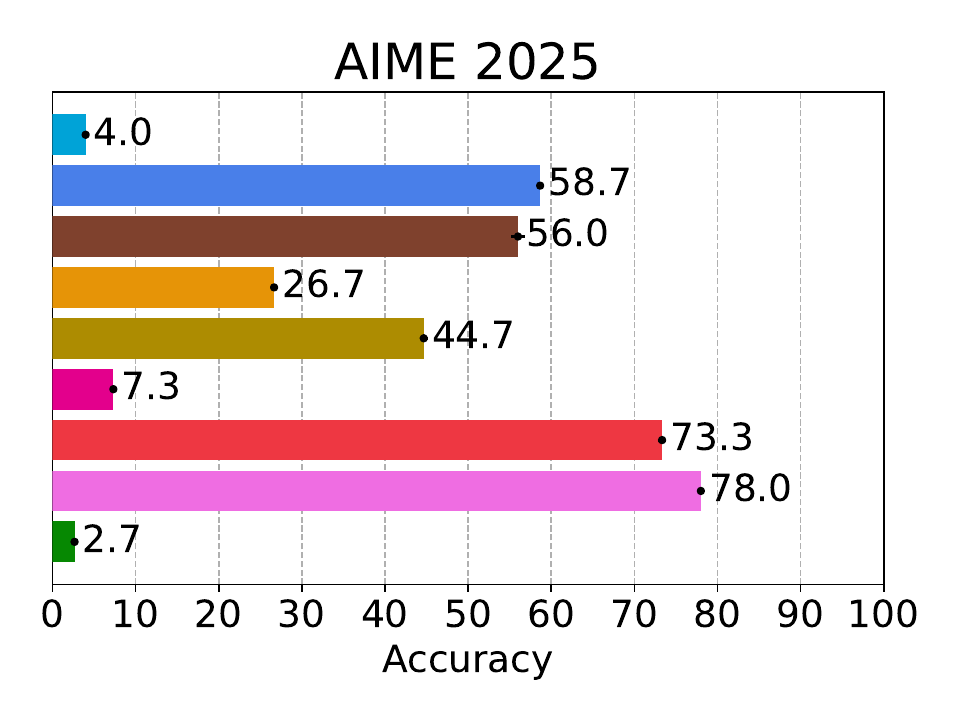}
    \end{subfigure}
    \begin{subfigure}[b]{0.24\textwidth}
        \centering
        \includegraphics[width=\textwidth]{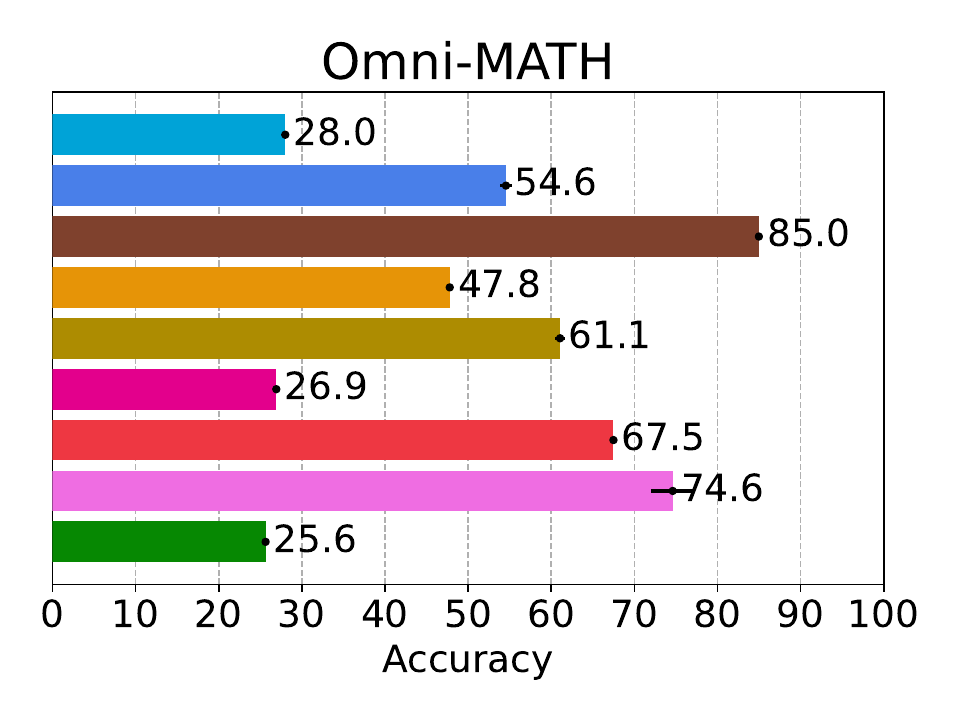}
    \end{subfigure}
    \begin{subfigure}[b]{0.24\textwidth}
        \centering
        \includegraphics[width=\textwidth]{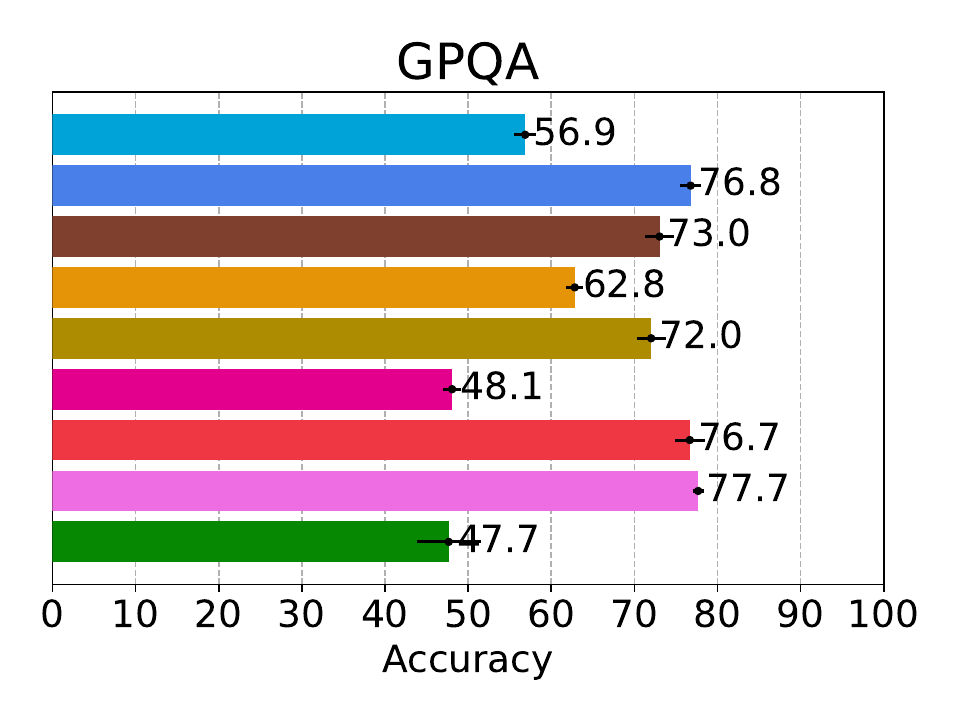}
    \end{subfigure}
    \begin{subfigure}[b]{0.24\textwidth}
        \centering
        \includegraphics[width=\textwidth]{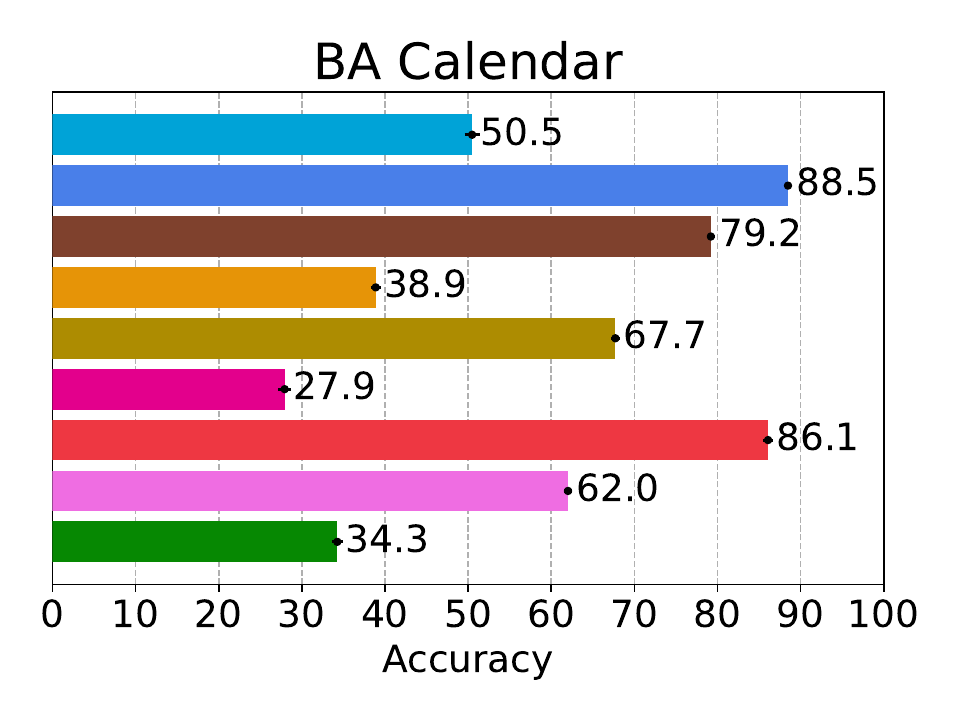}
    \end{subfigure}
        \begin{subfigure}[b]{0.24\textwidth}
        \centering
        \includegraphics[width=\textwidth]{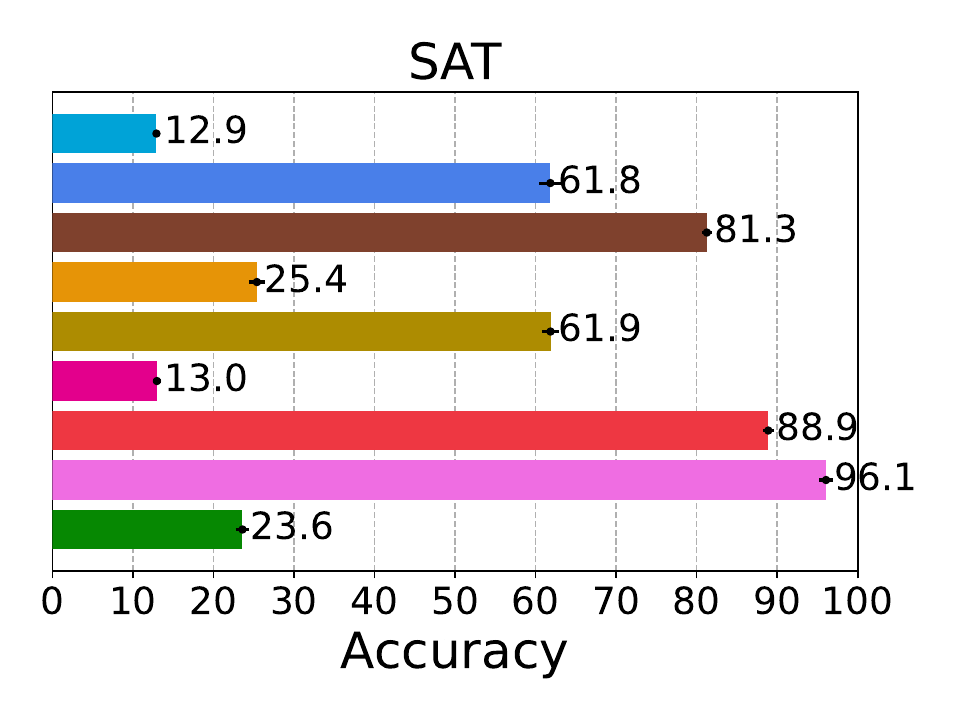}
    \end{subfigure}
    \begin{subfigure}[b]{0.24\textwidth}
        \centering
        \includegraphics[width=\textwidth]{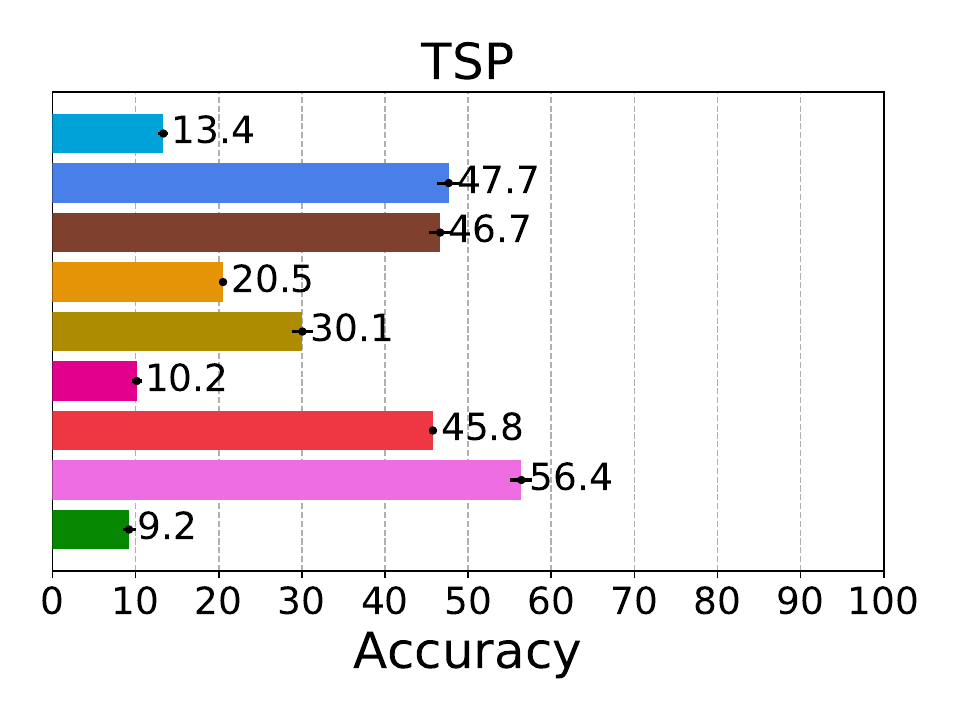}
    \end{subfigure}    
    \begin{subfigure}[b]{0.24\textwidth}
        \centering
        \includegraphics[width=\textwidth]{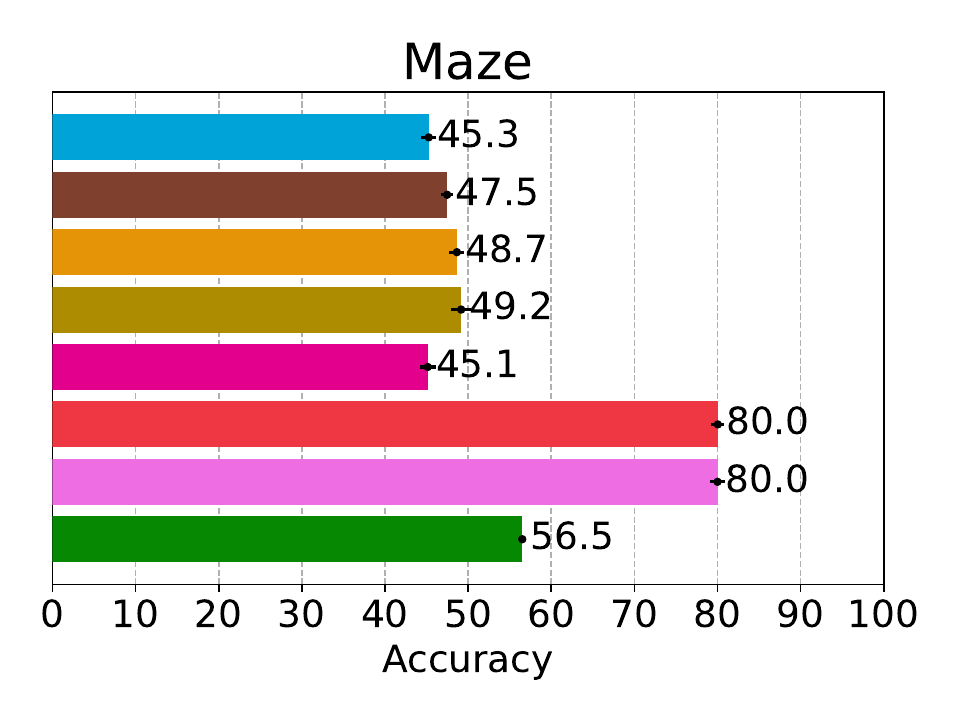}
    \end{subfigure}    
    \begin{subfigure}[b]{0.24\textwidth}
        \centering
        \includegraphics[width=\textwidth]{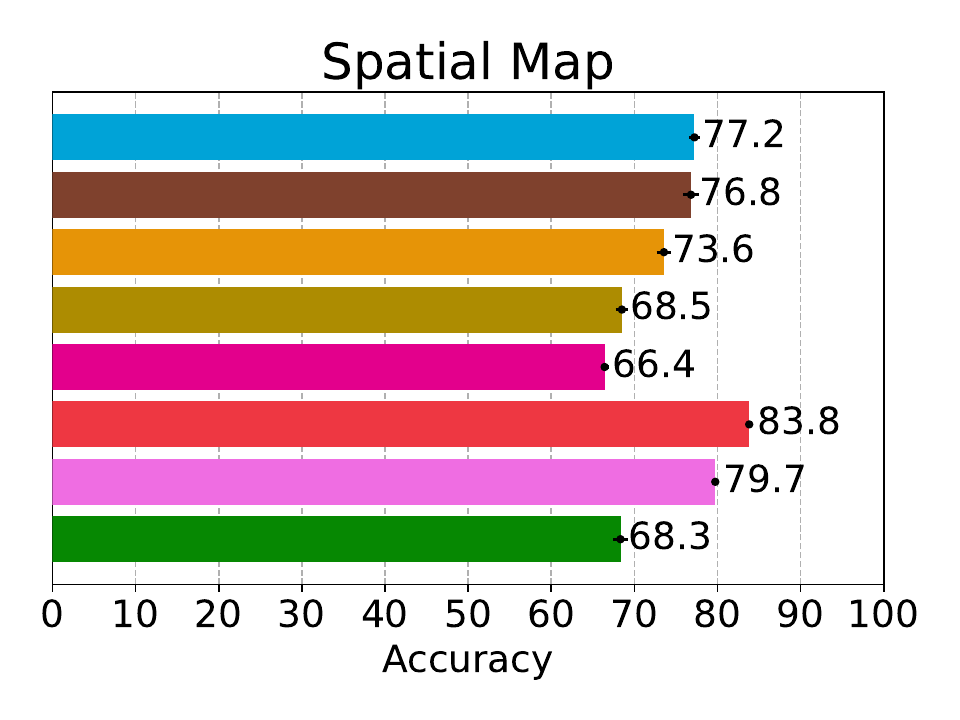}
    \end{subfigure}     
    \begin{subfigure}[b]
    {\textwidth}
        \centering
\includegraphics[width=\textwidth]{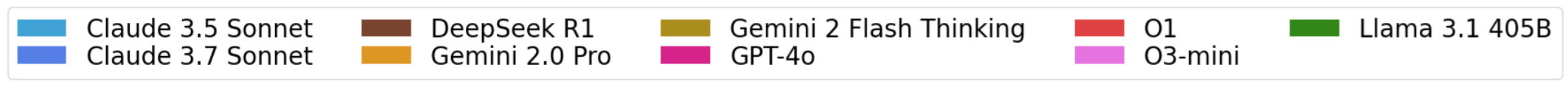}
    \end{subfigure}
    \caption{Overall Avg Pass@1 model performance across eight reasoning tasks. }
    \label{fig:all_in_one}
\end{figure}

\begin{figure}[t]
    \centering
    \begin{subfigure}[b]{0.49\textwidth}
        \centering
        \includegraphics[width=\textwidth]{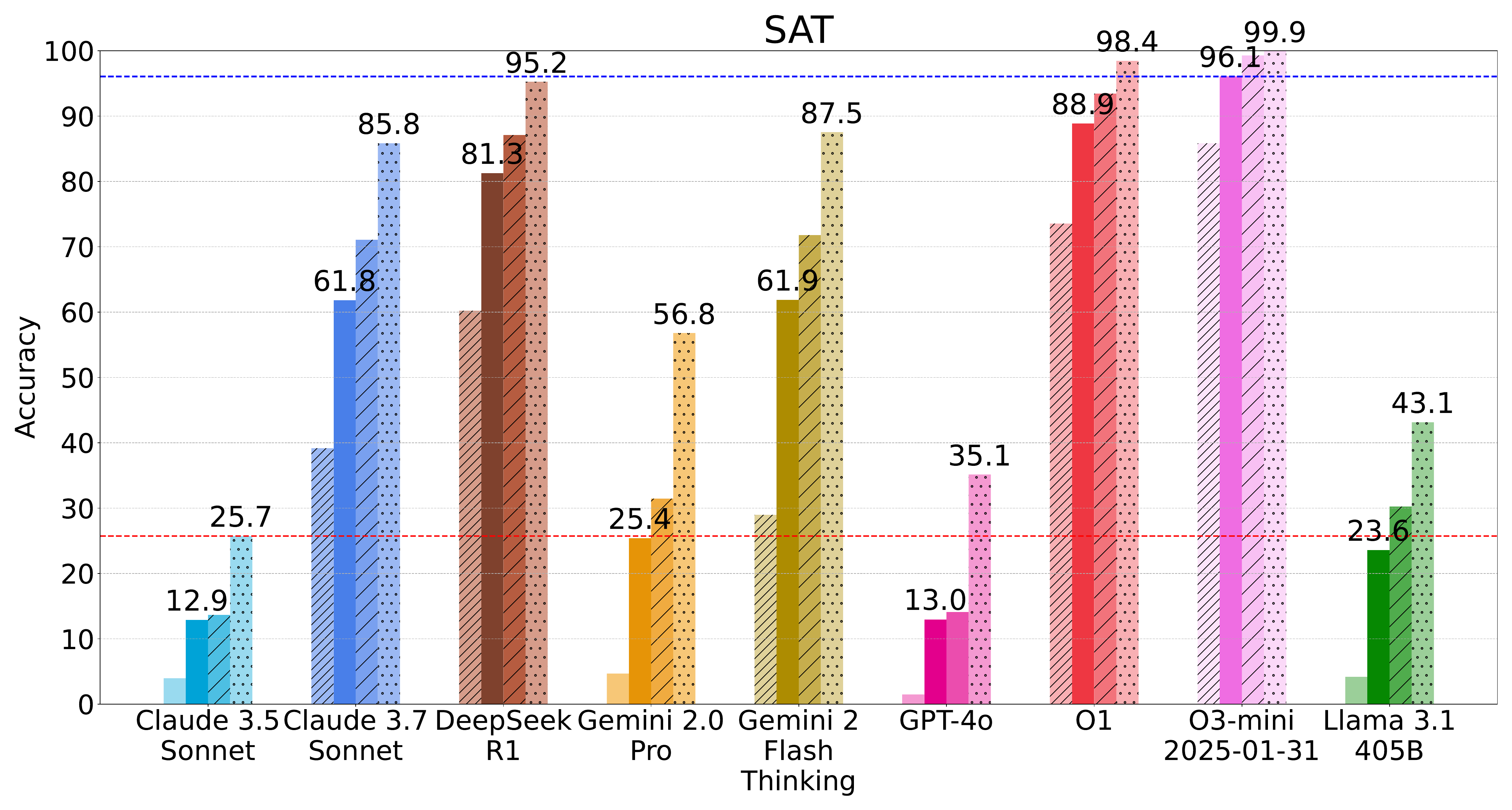}
    \end{subfigure}
    \begin{subfigure}[b]{0.49\textwidth}
        \centering
        \includegraphics[width=\textwidth]{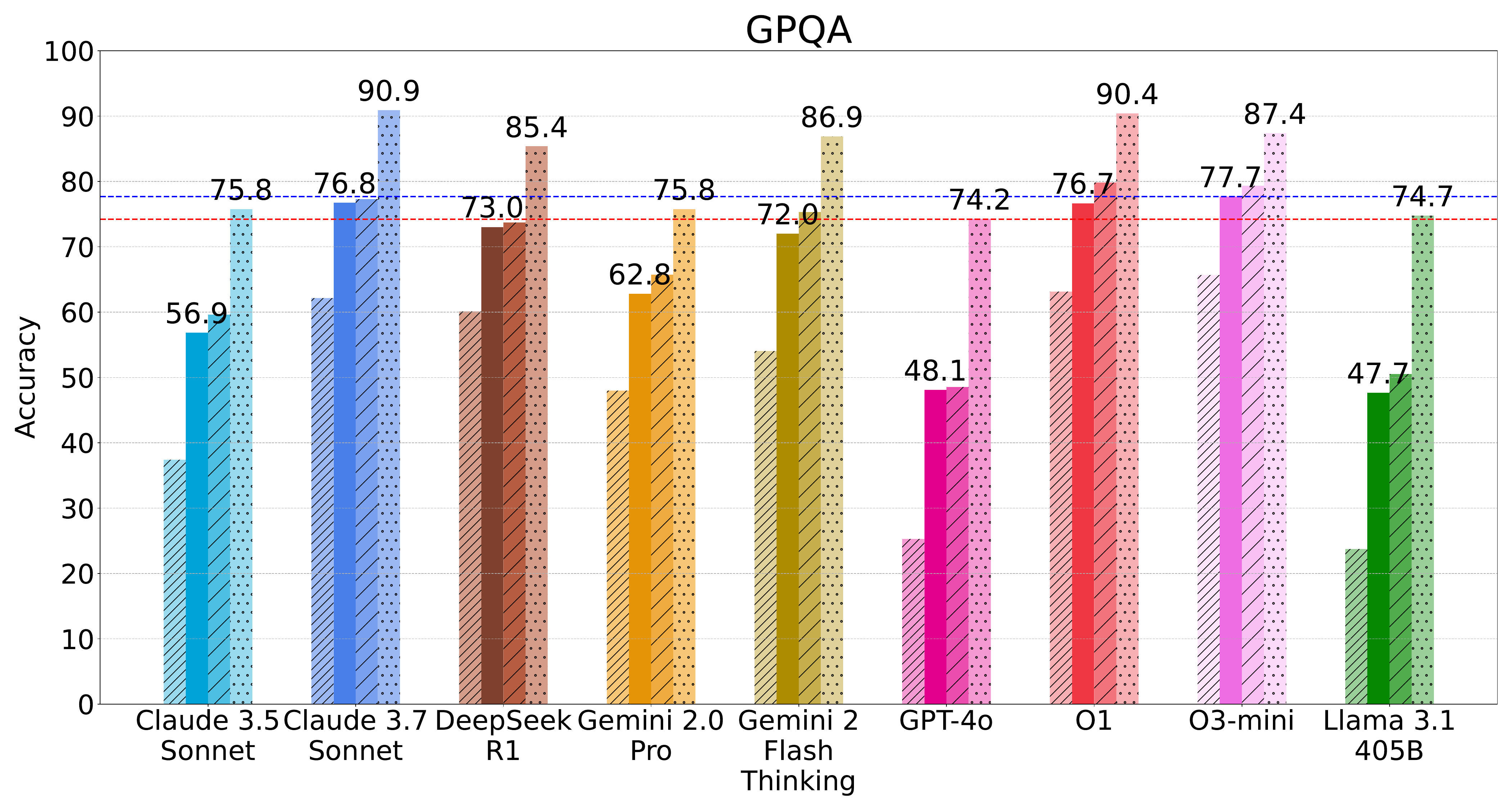}
    \end{subfigure}
    \begin{subfigure}[b]{0.49\textwidth}
        \centering
        \includegraphics[width=\textwidth]{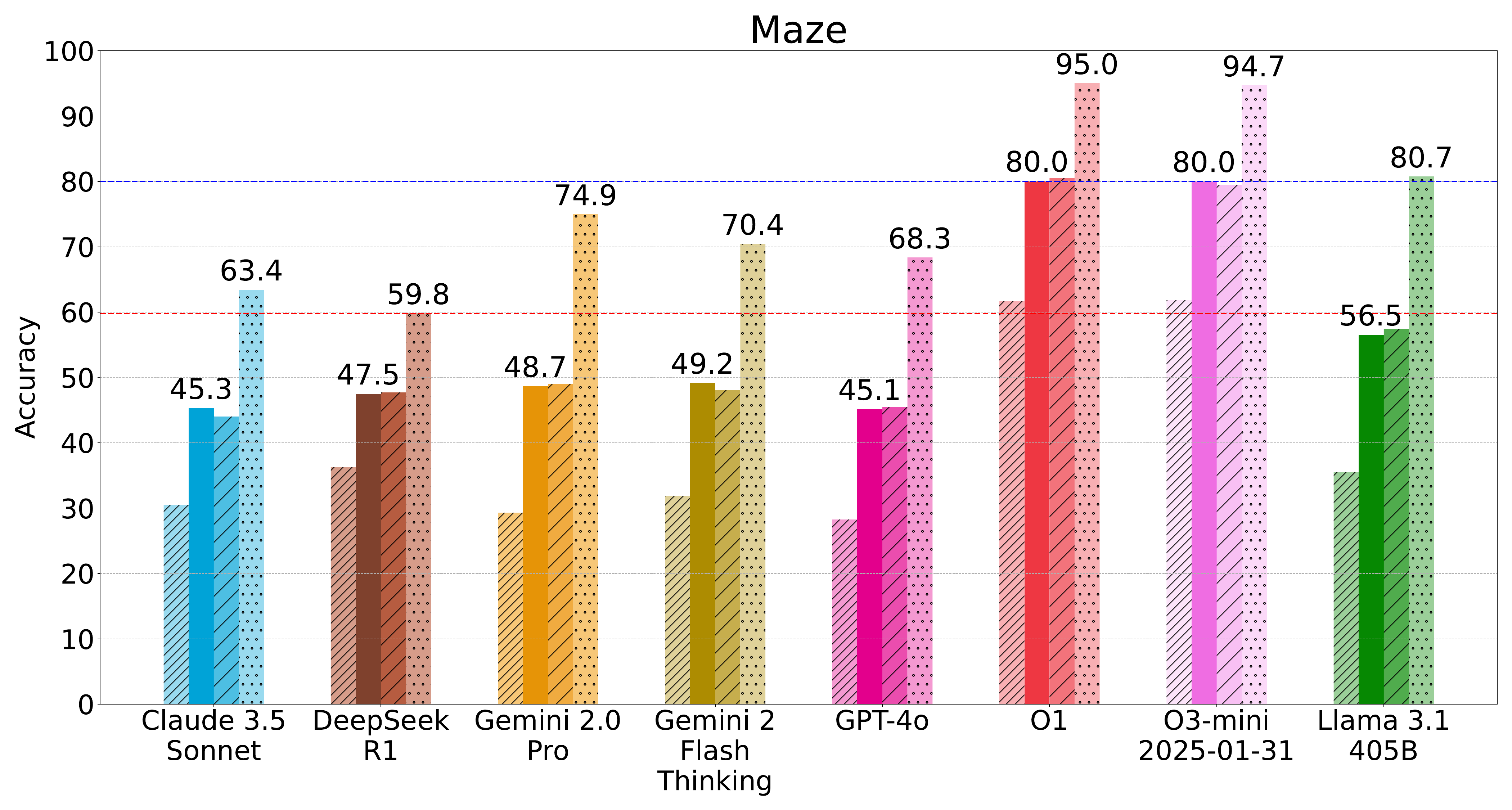}
    \end{subfigure} 
    \begin{subfigure}[b]{0.49\textwidth}
        \centering
        \includegraphics[width=\textwidth]{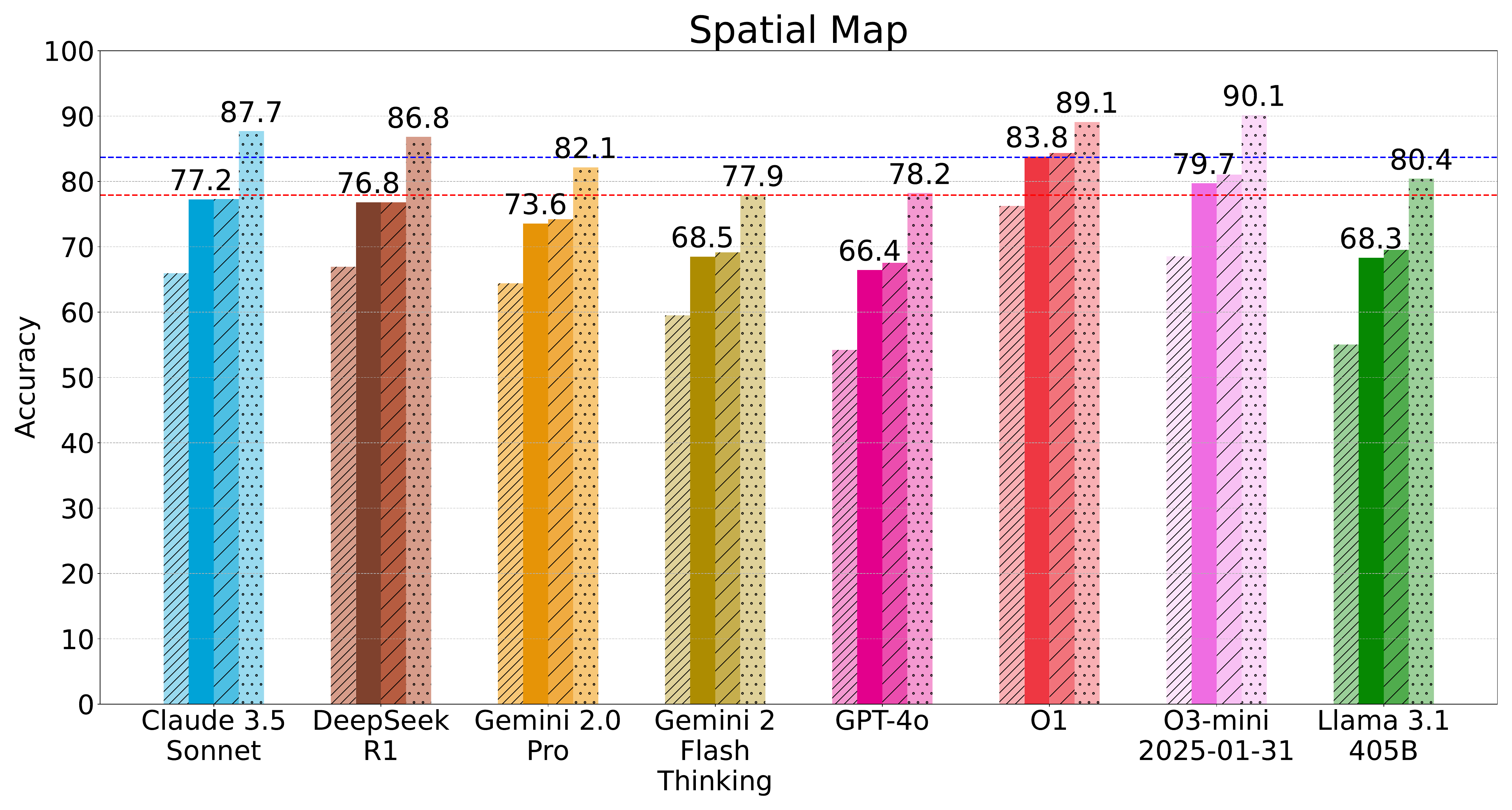}
    \end{subfigure}    
    \vspace{-15pt}
    \begin{subfigure}[b]
    {\textwidth}
        \centering
\includegraphics[width=0.5\textwidth]{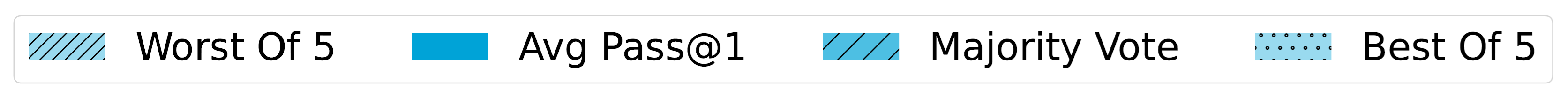}
        \label{fig:9}
    \end{subfigure}
    \vspace{-5pt}
    \caption{Results on 3SAT, GPQA, Maze, and SpatialMap with different aggregations by parallel scaling over 5 runs. The \textcolor{red}{red} line indicates the lowest best-of-5 accuracy observed across all models, while the \textcolor{blue}{blue} line represents the highest average pass@1 accuracy. Notably, there is a significant performance gap in 3SAT, where some models achieve high accuracy with aggregation, whereas others struggle even at their best-of-5 runs. The narrow conventional-to-reasoning gap between the two  on GPQA (3.5\%) and SpatialMap (5.5\%) shows that the best reasoning model is only slightly more accurate than a hypothetical model that can potentially be trained to verify and select the best outcome from the model with the lowest best-of-5 (i.e. GPT-4o).}
    \label{fig:main_inf_scaling}
\end{figure}
Figure~\ref{fig:all_in_one} presents an overview of average model performance over 5 independent runs across eight reasoning tasks, illustrating the generalization capability of various models when tested across diverse datasets. Results indicate that reasoning models like \ROne, \OOne and \OThree have consistently high performance across different tasks, suggesting strong reasoning capabilities. 
However, their performance varies significantly depending on the dataset, highlighting task-specific strengths and weaknesses. For example, while \ClaudeSonnetThinking performs on par with \OOne on some datasets, it underperforms in NP-hard tasks and Omni-MATH showing that its capabilities do not generalize to algorithmic hard problems or broader math. Even within the same domain, we observe variance in model performance across datasets. For example, for math reasoning, while \OOne and \OThree outperform \ROne on AIME, the opposite is the case for Omni-MATH, which is a larger and more diverse benchmark. Moreover, all models show a performance drop on AIME 2025 compared to AIME 83-24.

In addition, we also conduct a disaggregated analysis for certain benchmarks on meaningful subcategories of their data. Related to generalization, we observe that based on GPQA measurements (Appendix~\ref{sec:gpqa} Figure~\ref{fig:gpqa_domain}) all reasoning models perform worse on Chemistry and Biology, despite spending more tokens on such problems. This shows that inference-time scaling methods do not benefit all domains equally. A similar analysis on different math topics in Omni-MATH (Appendix~\ref{sec:omnimath} Figure~\ref{fig:omni_math_const_acc}) shows that all reasoning models have a lower accuracy on problems in geometry and discrete math.

Figure~\ref{fig:main_inf_scaling} further analyzes model accuracy using different aggregation methods, such as best-of-5, majority voting, and worst-of-5 accuracy for four of the benchmarks. The gap between the red line (worst best-of-5) and the blue line (best average pass@1) shows the \emph{conventional-to-reasoning gap}. For tasks like 3SAT, TSP, AIME, and Omni-MATH there is a large gap, suggesting that for these problems simple outcome-based verification is not sufficient. For other tasks like GPQA and SpatialMap, the best-of-5 scaling approach already gets the models close to the best reasoning model.  Across all benchmarks and models we observe additional gains when having a perfect verifier for aggregation (best-of-n), providing an encouraging signal that there is further potential for improvement.

From a model reliability perspective, it is also useful to look at the difference between worst-of-5 and average performance, which varies between 10\%-20\% across models and tasks.

\paragraph{Performance vs. token usage tradeoffs.}
\begin{figure}[t]
    \centering
    \begin{subfigure}[b]{0.24\textwidth}
        \centering
        \includegraphics[width=\textwidth]{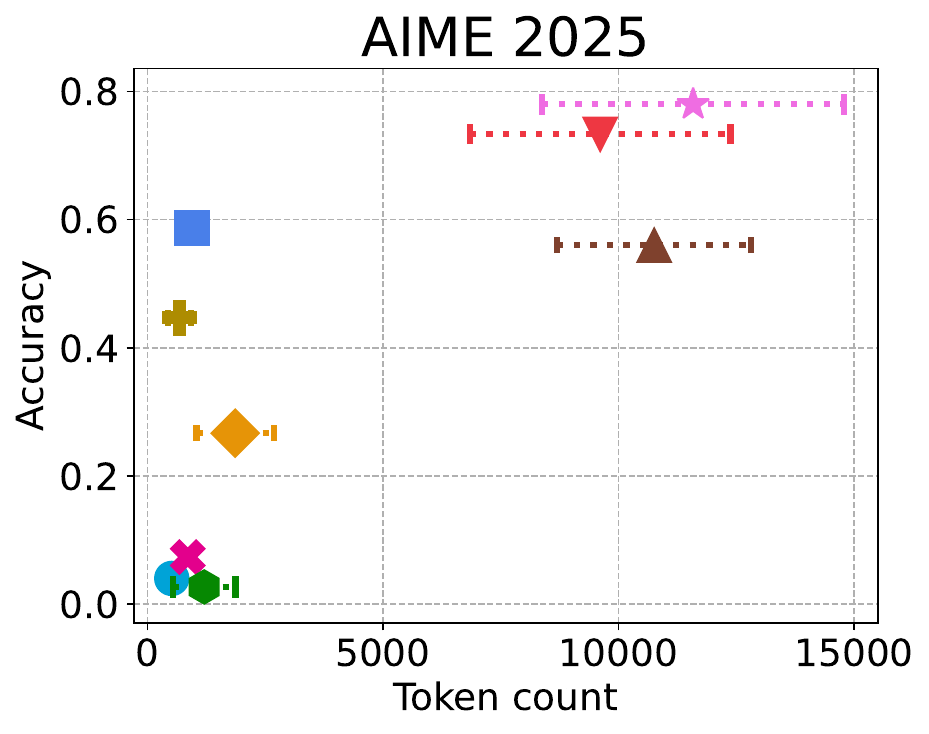}
    \end{subfigure}
    \begin{subfigure}[b]{0.24\textwidth}
        \centering
        \includegraphics[width=\textwidth]{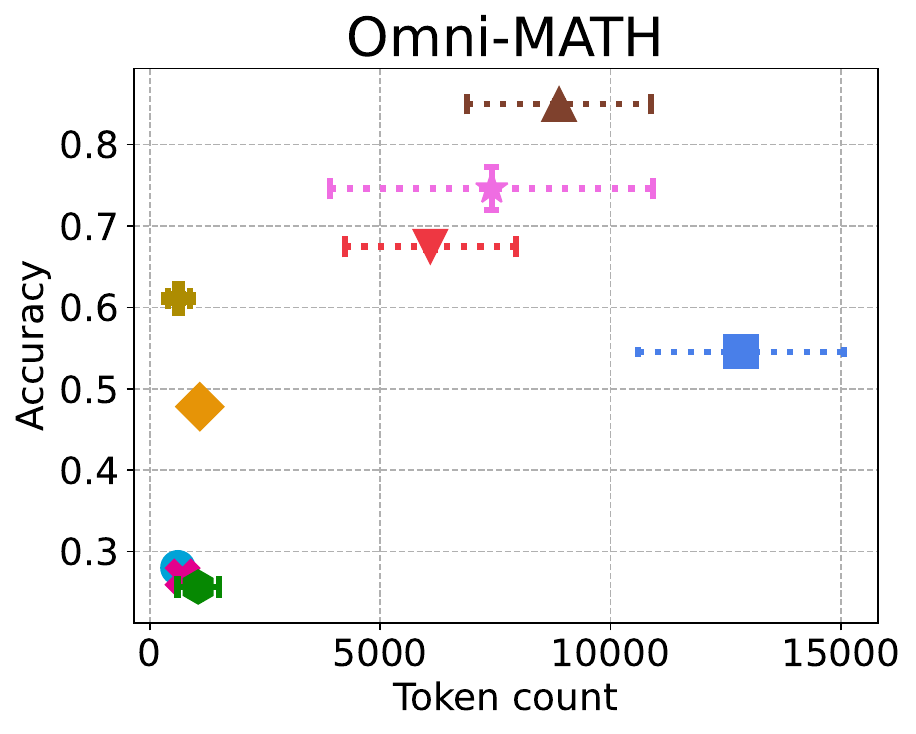}
    \end{subfigure}
    \begin{subfigure}[b]{0.24\textwidth}
        \centering
        \includegraphics[width=\textwidth]{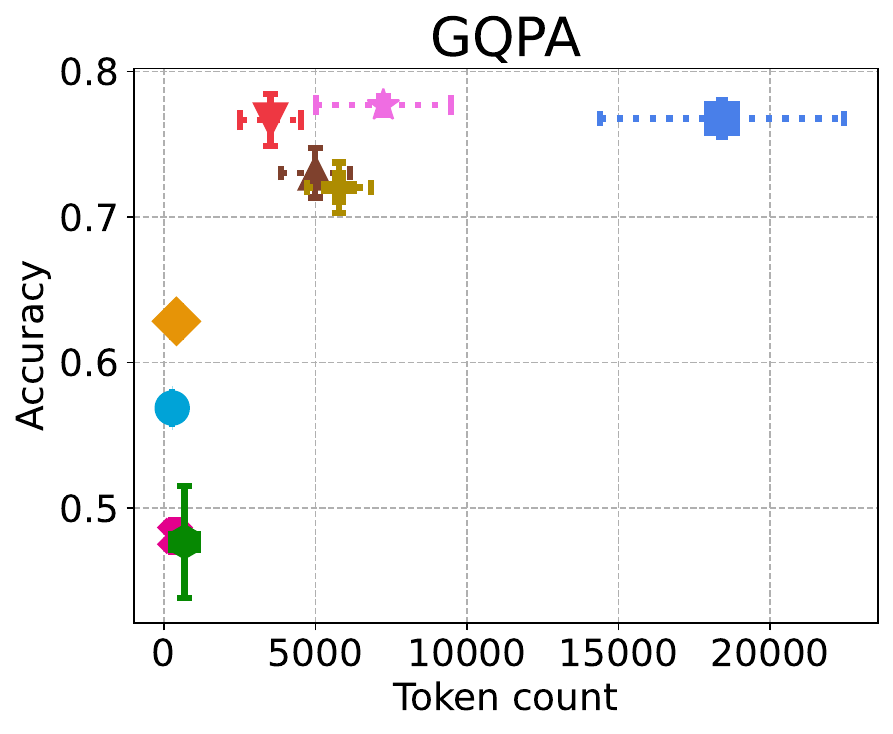}
    \end{subfigure}
    \begin{subfigure}[b]{0.24\textwidth}
        \centering
        \includegraphics[width=\textwidth]{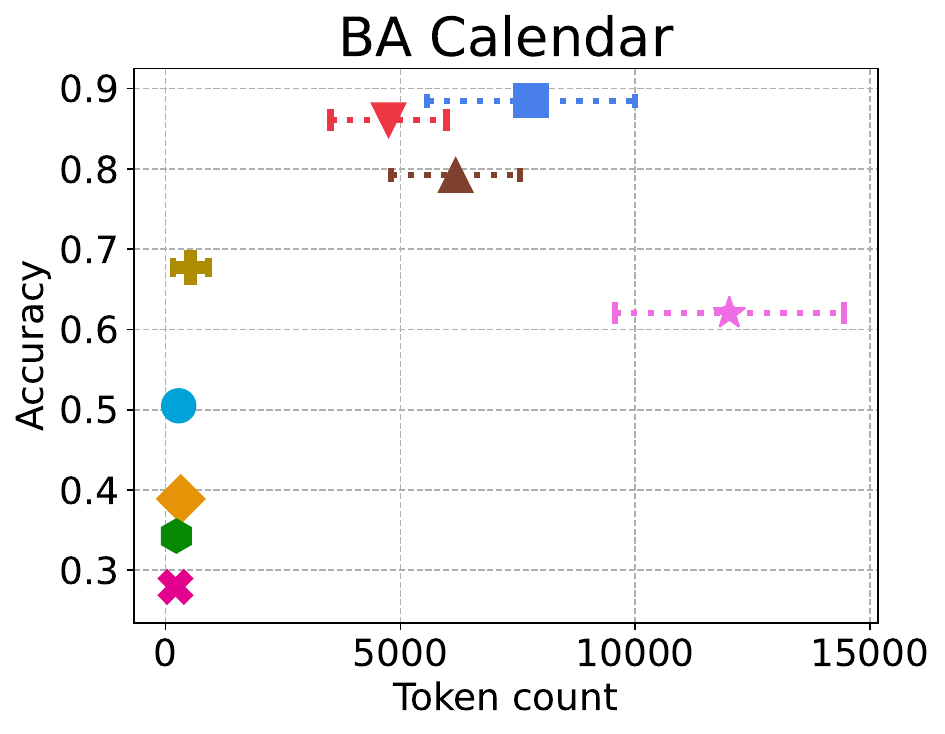}
    \end{subfigure}
    \begin{subfigure}[b]{0.24\textwidth}
    \centering
        \includegraphics[width=\textwidth]{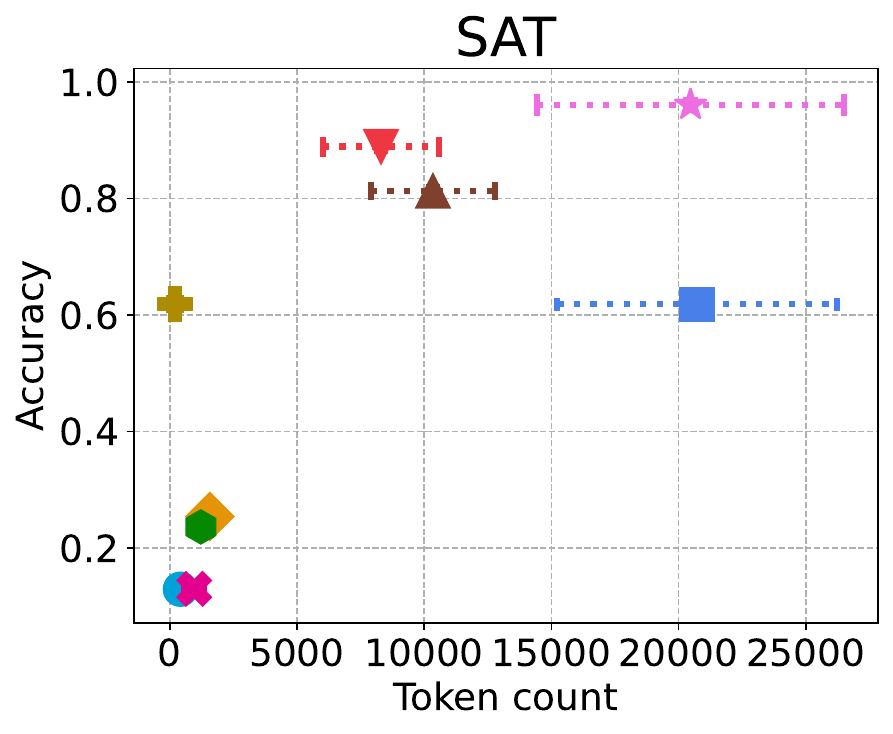}
    \end{subfigure}
    \begin{subfigure}[b]{0.24\textwidth}
        \centering
        \includegraphics[width=\textwidth]
        {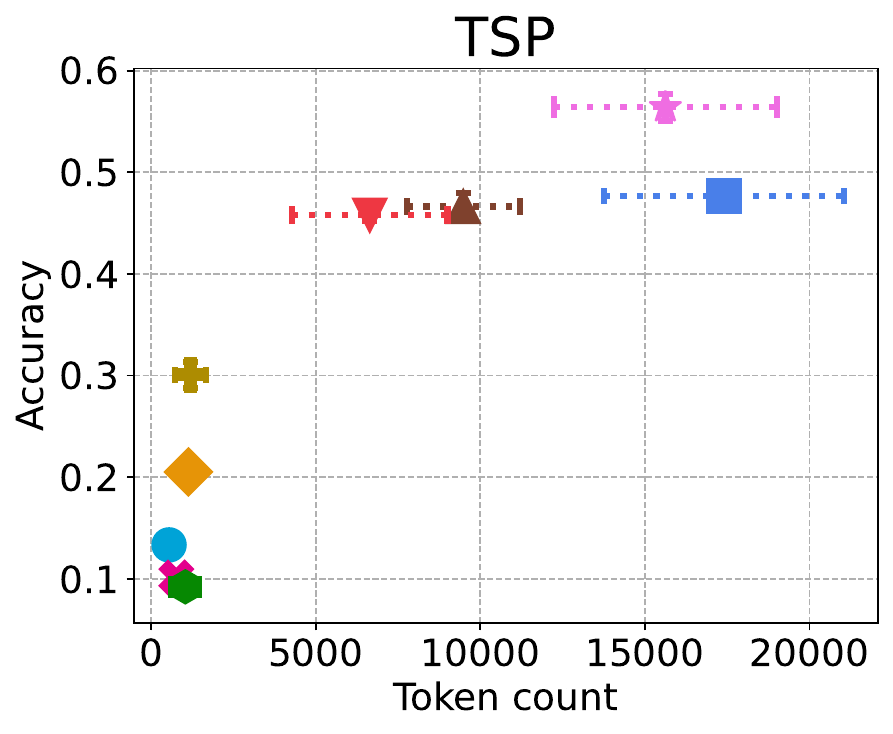}
    \end{subfigure}
    \begin{subfigure}[b]{0.24\textwidth}
        \centering
        \includegraphics[width=\textwidth]{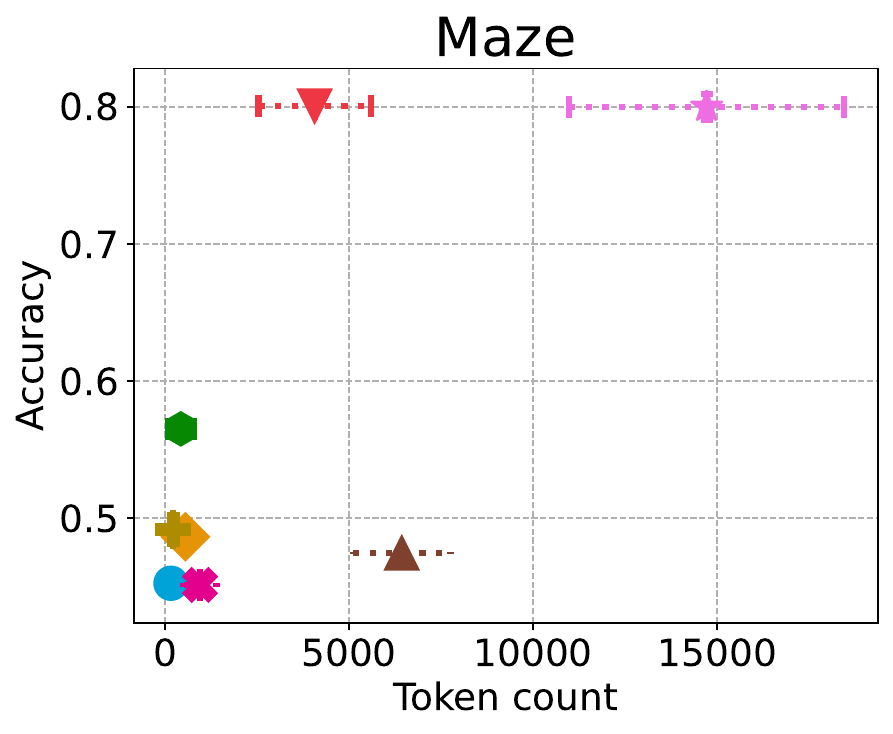}
    \end{subfigure}
    \begin{subfigure}[b]{0.24\textwidth}
        \centering
        \includegraphics[width=\textwidth]{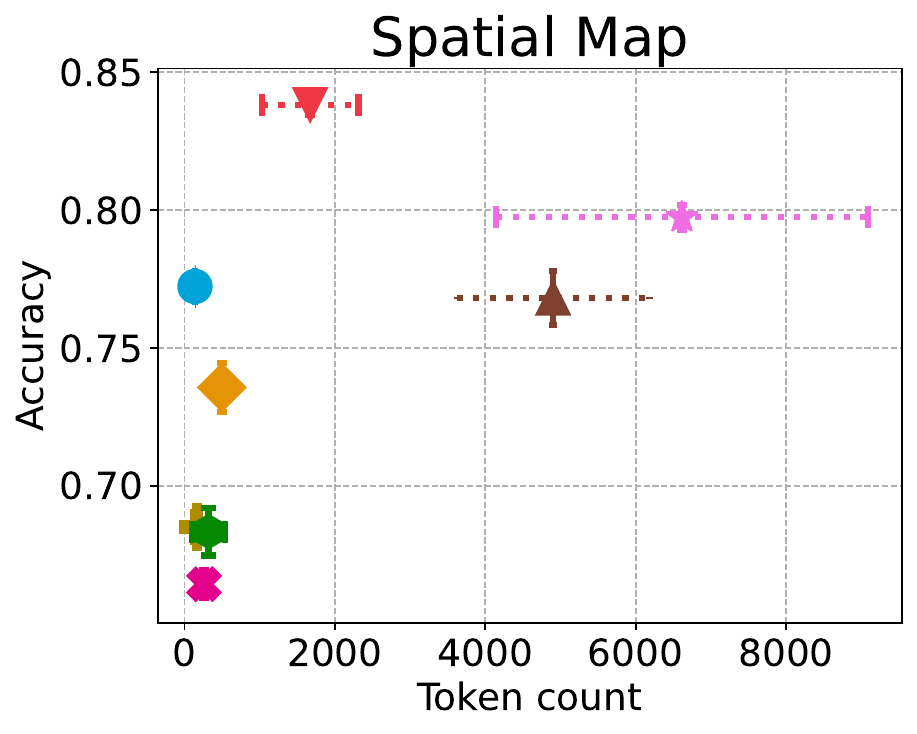}
    \end{subfigure}    
     \begin{subfigure}[b]{\textwidth}
        \centering
\includegraphics[width=1.0\textwidth]{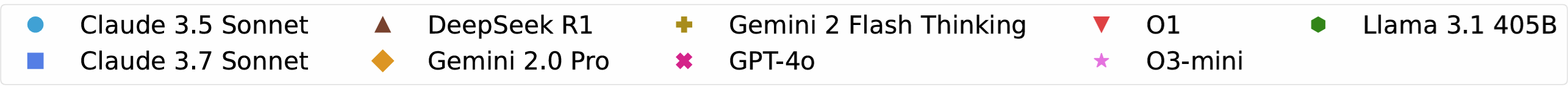}
    \end{subfigure}
    \caption{Pareto tradeoff between accuracy and token usage for all benchmarks. The standard deviation for accuracy (vertical, filled line) is computed across 5 different repetitions. The standard deviation for token usage (horizontal, dotted line) is computed by first taking the standard deviation per data instance, and then averaging by the size of the benchmark, to show the variability per instance.
    }
    \label{fig:all_in_one_accuracy_tokens}
\end{figure}

\begin{figure}[t]
    \centering
    \begin{subfigure}[b]{0.24\textwidth}
    \centering
    \includegraphics[width=\textwidth]{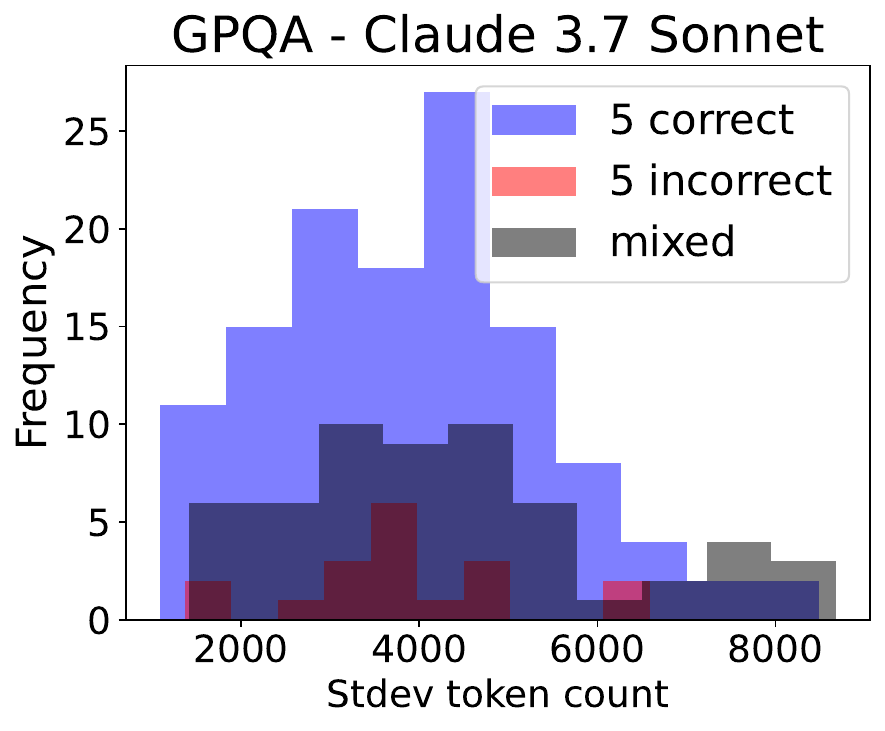}
    \end{subfigure}
    \begin{subfigure}[b]{0.245\textwidth}
        \centering
        \includegraphics[width=\textwidth]{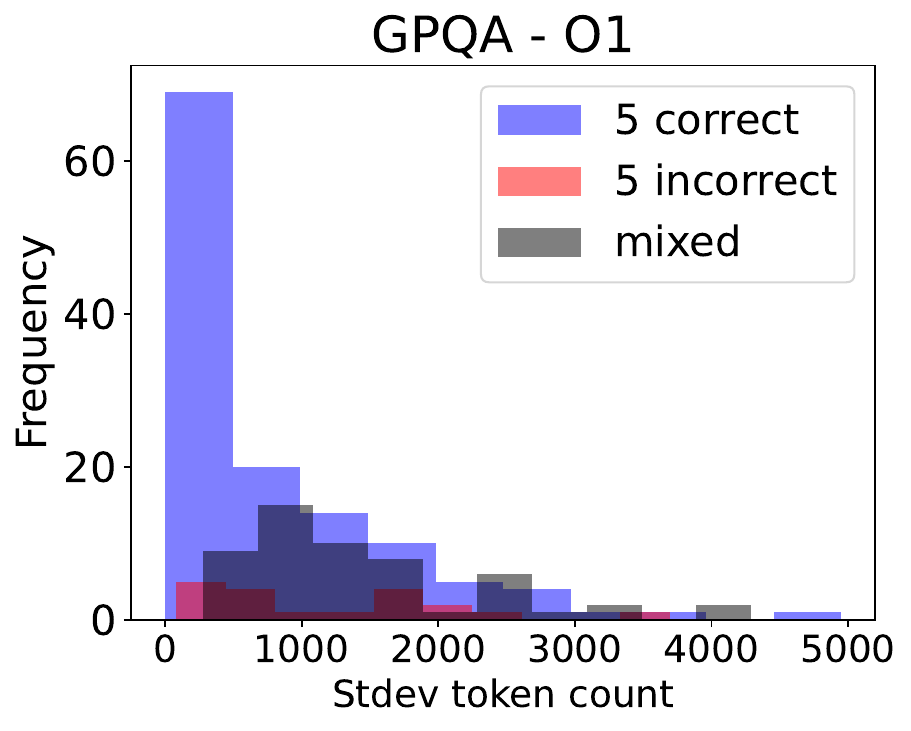}
    \end{subfigure}
    \begin{subfigure}[b]{0.25\textwidth}
    \centering
    \includegraphics[width=\textwidth]{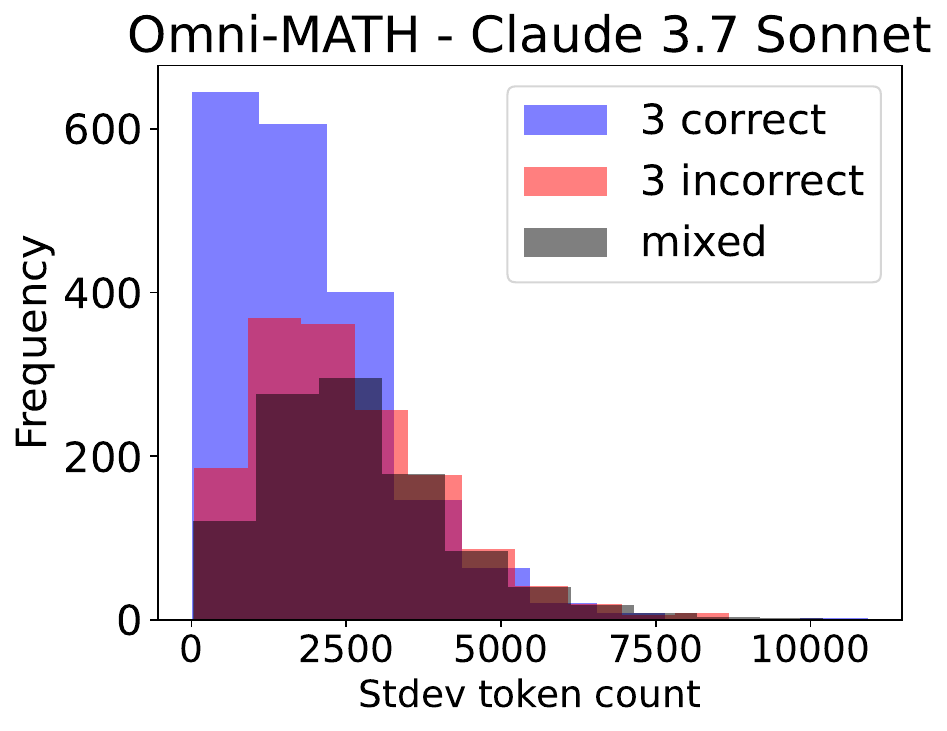}
    \end{subfigure}
    \begin{subfigure}[b]{0.245\textwidth}
        \centering
        \includegraphics[width=\textwidth]{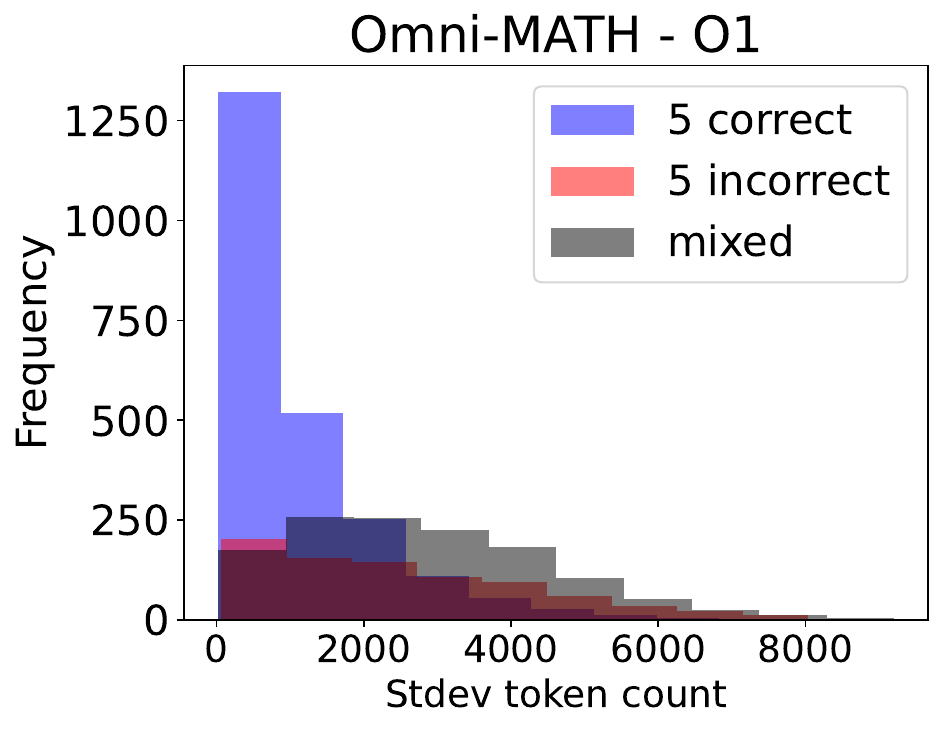}
    \end{subfigure}
    
    \caption{Distributions of the standard deviations of token usage within the same instance (5 repeats), shown for instances where the models are always correct, always incorrect, or mixed (figure is continued in Figure~\ref{fig:correct_incorrect_token_usage_stdev_appendix} for more models). Models often have high standard deviation of token usage even when all the retrieved answers are correct.}
\label{fig:correct_incorrect_token_usage_stdev}
\end{figure}

\begin{figure}[t]
    \centering
    \begin{subfigure}[b]{0.24\textwidth}
        \centering
        \includegraphics[width=\textwidth]{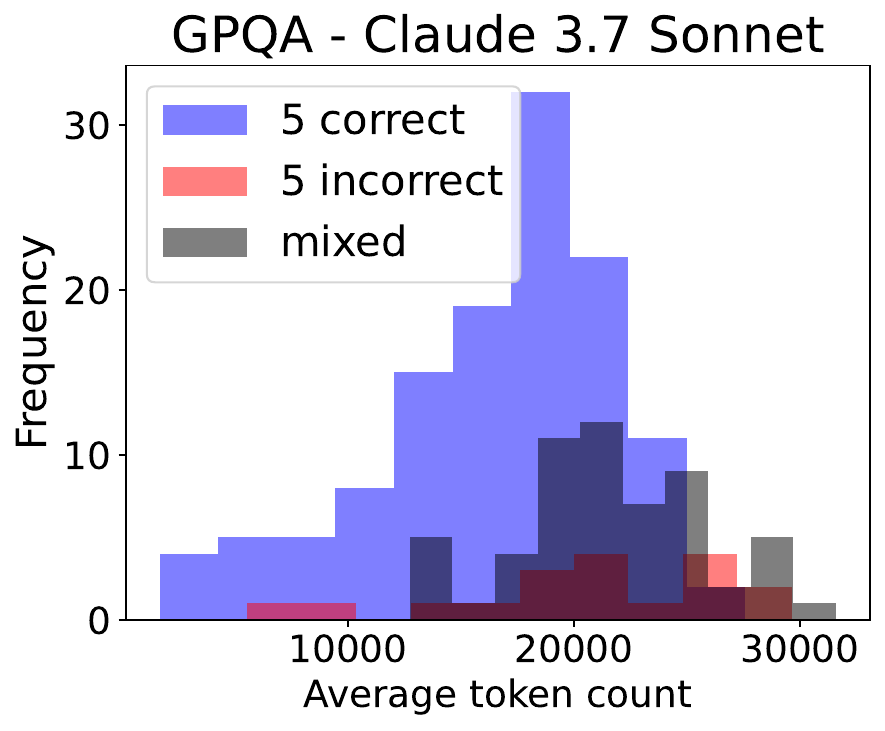}
    \end{subfigure}
    \begin{subfigure}[b]{0.24\textwidth}
        \centering
        \includegraphics[width=\textwidth]{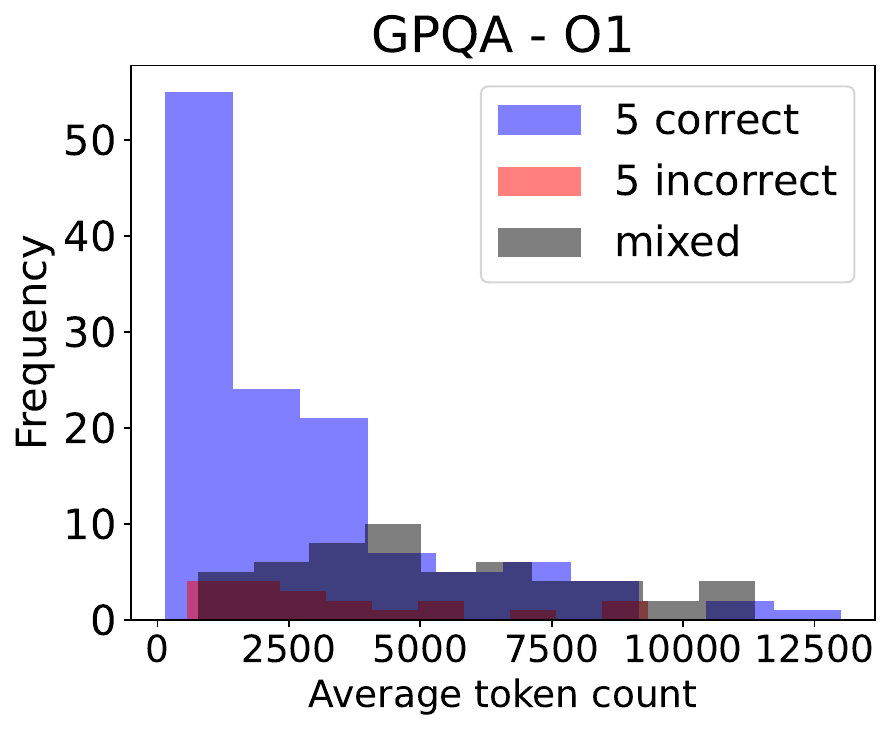}
    \end{subfigure}
   \begin{subfigure}[b]{0.24\textwidth}
       \centering
       \includegraphics[width=\textwidth]{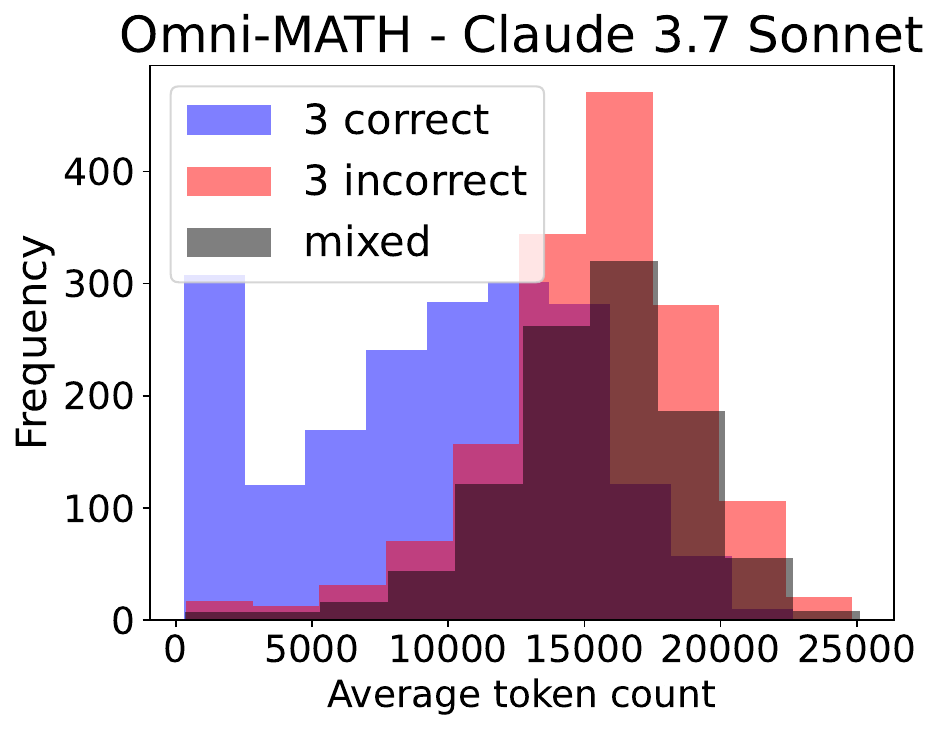}
   \end{subfigure}
    \begin{subfigure}[b]{0.24\textwidth}
        \centering
        \includegraphics[width=\textwidth]{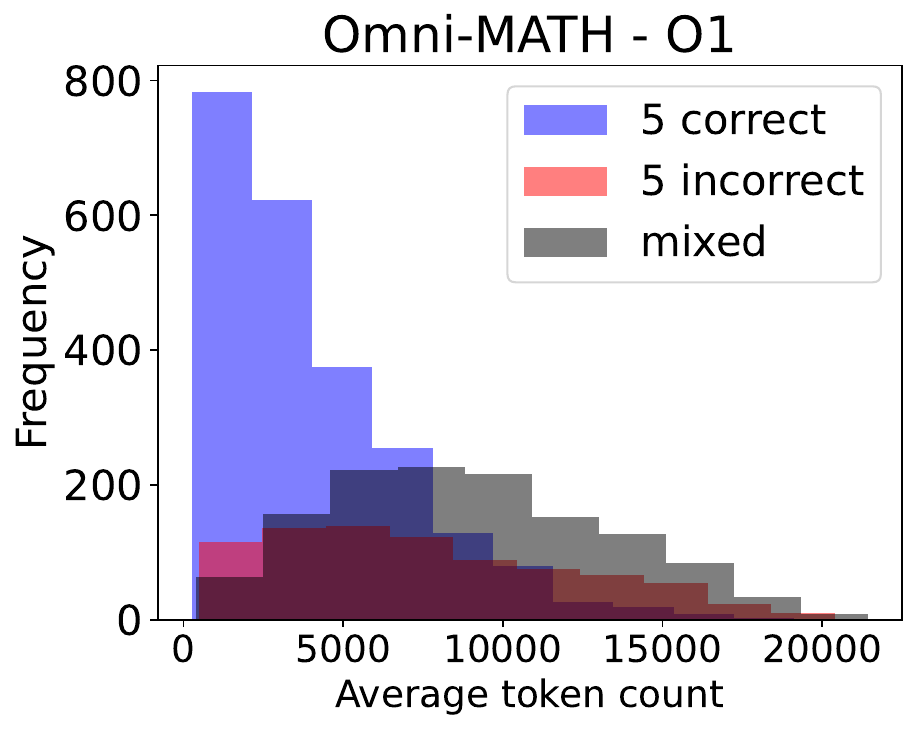}
    \end{subfigure} 
    \caption{Distributions of average token usage, shown for instances where the models are always correct, always incorrect, or mixed (figure is continued in Figure~\ref{fig:correct_incorrect_token_usage_mean_appendix} for more models). \OOne 
    has a higher concentration of ``all correct'' instances towards the shorter lengths, while for other models the ``all correct'' instances are more spread out indicating more unpredictability of token usage across instances even when the model is always correct.}
    \label{fig:correct_incorrect_token_usage_mean}
\end{figure}

\begin{figure}[t]    
     \begin{subfigure}[b]{\textwidth}    
     \centering
    \includegraphics[width=0.47\linewidth]{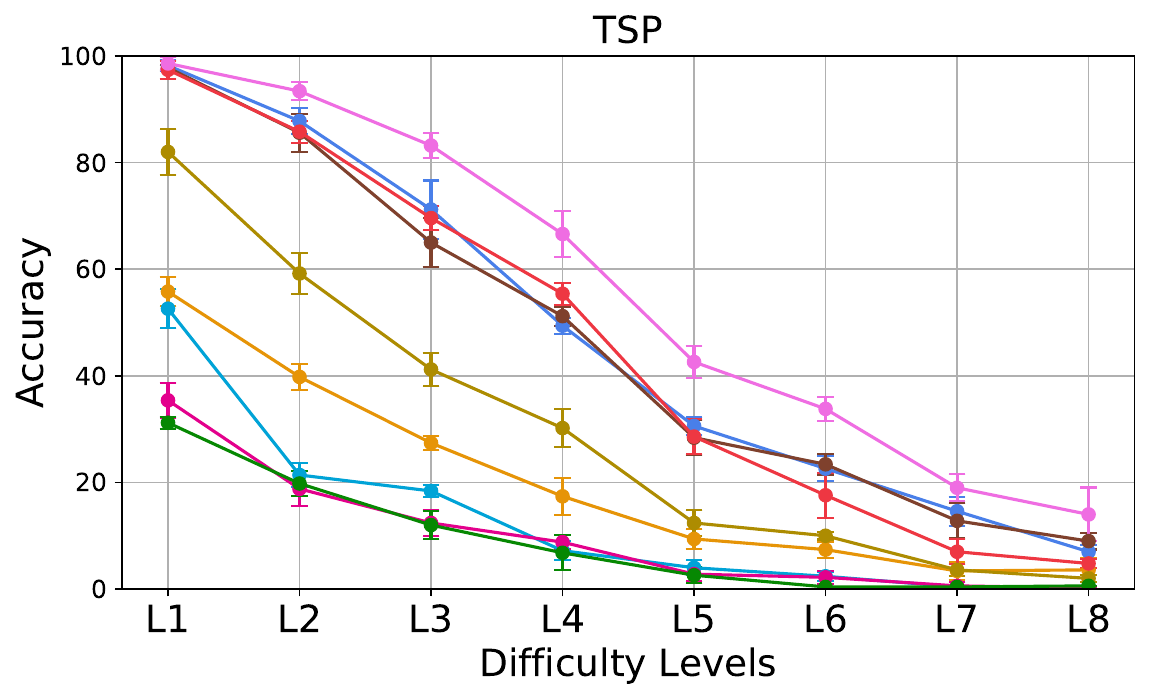}
    \includegraphics[width=0.47\linewidth]{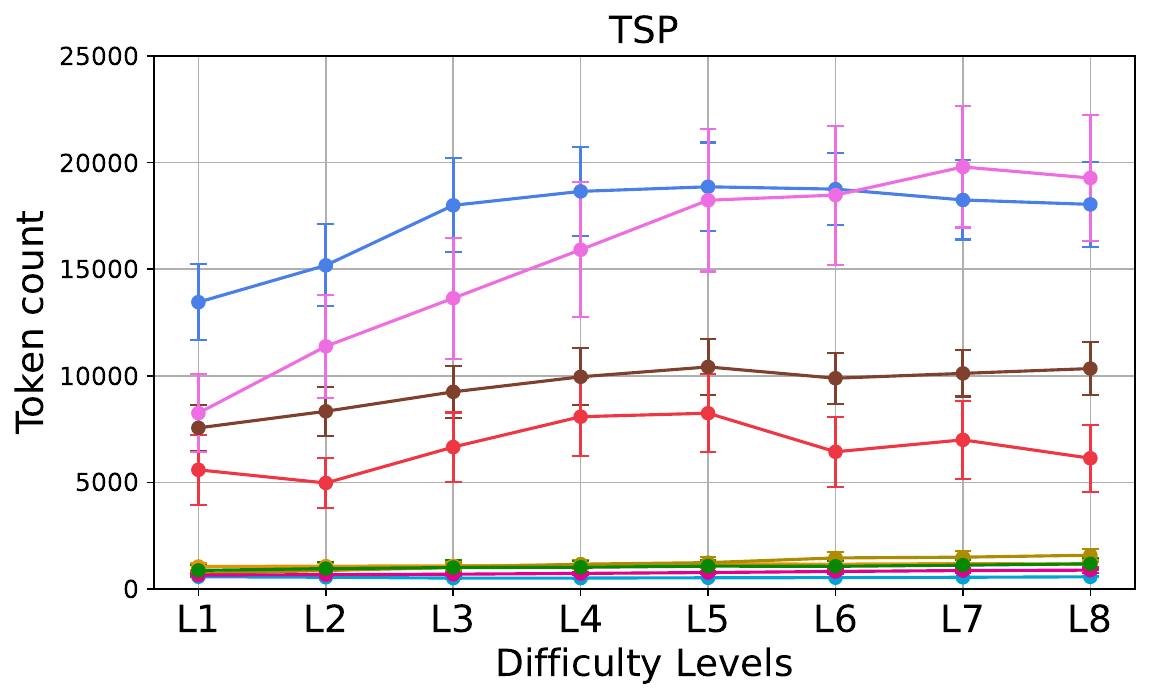}
\end{subfigure}
    \begin{subfigure}[b]{\textwidth}
        \centering
        \includegraphics[width=0.47\textwidth]{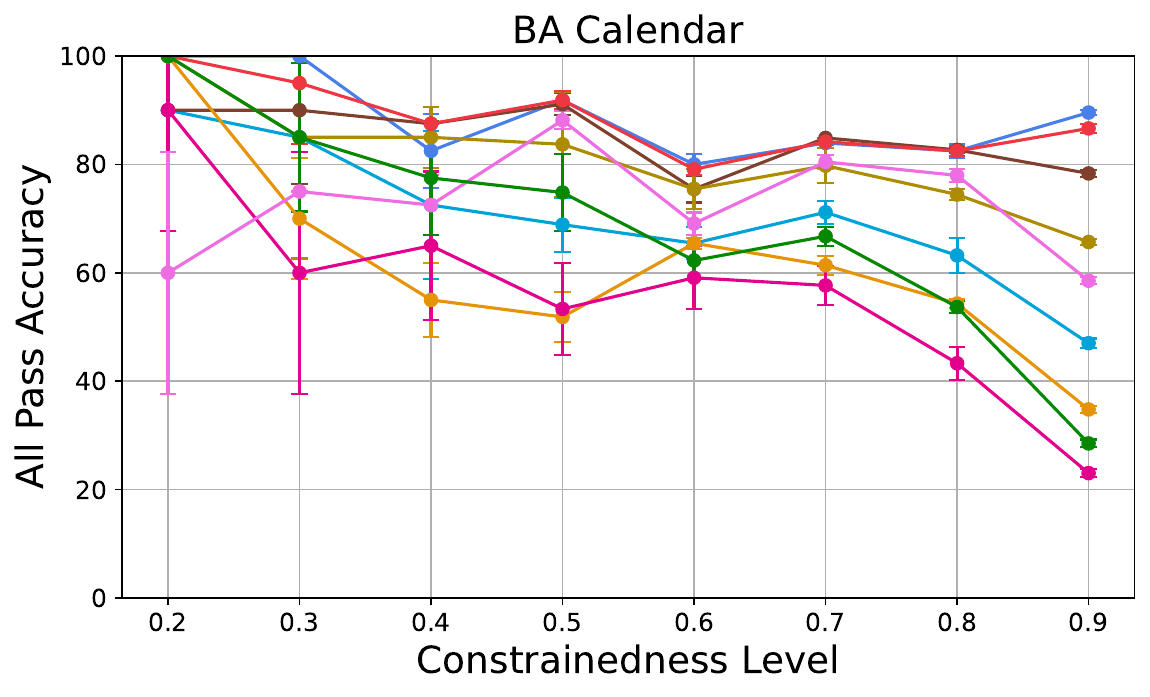}
        \includegraphics[width=0.47\textwidth]{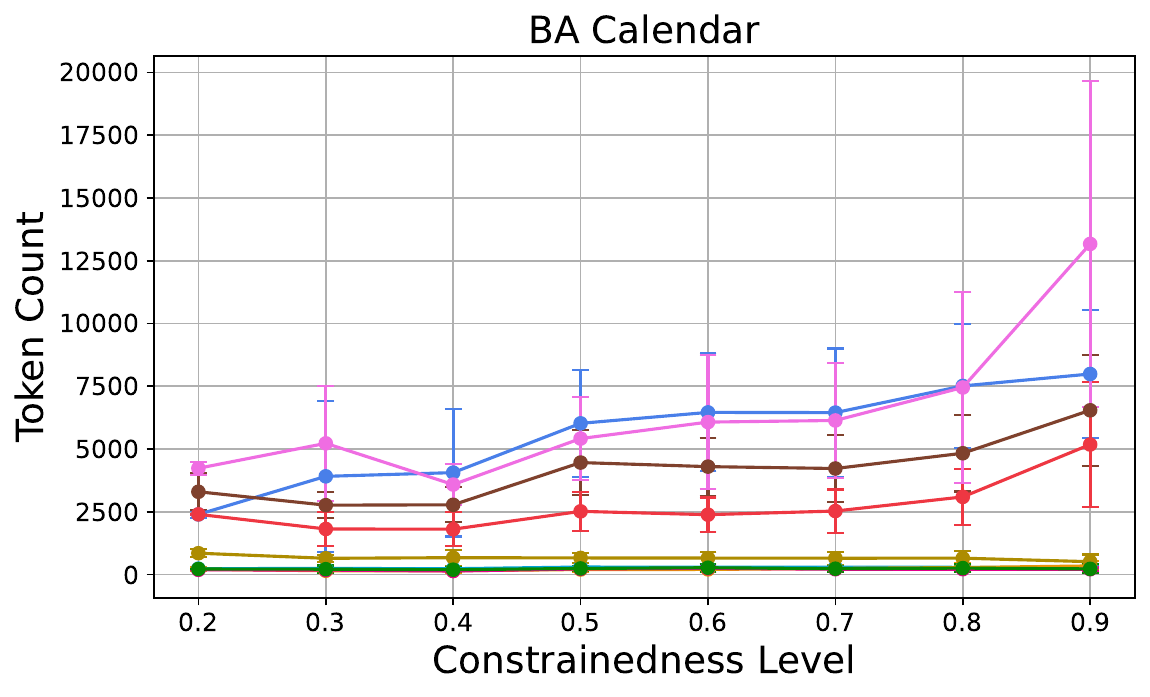}
    \end{subfigure}
     \begin{subfigure}[b]{\textwidth}
        \centering
\includegraphics[width=1.0\textwidth]{figures/model_legend.png}
    \end{subfigure}
    
    \caption{TSP and BA-Calendar accuracy and token usage with difficulty levels. Standard deviation for token usage is computed across different parallel repeats.}
    \vspace{-2pt}
    \label{fig:tsp_token_accuracy_difficulty}
\end{figure}

Next, we study the tradeoffs between accuracy and token usage. Throughout these results, token usage corresponds to the total number of tokens that the model uses for both output and reasoning. There are three important aspects to these tradeoffs: variability in token usage (i) \emph{across models}, (ii) within the \emph{same instance and the same model}, and (iii) within the \emph{same model but across different data instances}.

Figure~\ref{fig:all_in_one_accuracy_tokens} shows the average accuracy of each model vs. the average number of tokens used. Here, we can see trends and tradeoffs across models. For example, we can observe that often there exist pairs of models that have similar accuracy but one of them uses a lot more tokens (e.g. for AIME 25, \ROne and \ClaudeSonnetThinking have an average accuracy across five repeats within a $\leq$3\% range, but \ROne uses at least 5 times more tokens). This indicates that the same task can be solved with the same level of accuracy but more efficiently, and that higher token consumption does not indicate higher accuracy across models. While there isn't a model that provides the best Pareto tradeoff (top left corner of these charts) consistently for all tasks, \OOne is the model that most frequently provides the best tradeoff (at least five out of eight benchmarks).

The standard deviations for token usage in Figure~\ref{fig:all_in_one_accuracy_tokens} (horizontal dotted lines) are computed by first taking the standard deviation per data instance, and then averaging by the size of the benchmark, to show the \emph{variability per instance}. Semantically, these standard deviations show how much cost nondeterminism one should expect for posing the same query multiple times to the same model. While accuracy and outcome nondeterminism is expected at repeats with high temperature, or even with temperature zero as shown by \cite{balachandran2024eureka} for conventional models, cost nondeterminism is a new behavior that is specific to reasoning models and can impact real-world usability and user preferences. Ideally, developers and users would prefer models for which the standard deviation on token usage per instance is low for cost predictability. We further delve into this behavior in Figures~\ref{fig:correct_incorrect_token_usage_stdev} and~\ref{fig:correct_incorrect_token_usage_stdev_appendix} by splitting the standard deviations per instance for cases where the model is always correct, always incorrect or mixed. These results show that cost nondeterminism exists even when the model is always correct and is more prominent in \ClaudeSonnetThinking.

Further, we also investigate how accuracy and token usage changes with problem difficulty (i.e., same model on different instances) for benchmarks that have a notion of problem difficulty: TSP (Figure~\ref{fig:tsp_token_accuracy_difficulty}), 3SAT (Figure~\ref{fig:sat_accuracy_tokens_difficulty_ver1}), BA-Calendar (Figure~\ref{fig:tsp_token_accuracy_difficulty}), Omni-MATH (Figure~\ref{fig:omni_math_tok_perf}). Overall, reasoning models have higher average token usage and lower accuracy on more difficult problems. However, the growth rate of token usage vs. problem difficulty varies between benchmarks. In BA Calendar, models seem to be better at maintaining their accuracy despite increased difficulty, and token usage continues to increase consistently with problem difficulty. This indicates that for these problems the models are better at utilizing inference-time scaling and lengthening their scratchpads effectively with increased difficulty. However, for TSP, token usage saturates approximately after level 6, while accuracy drops much faster, even for the best models. 

Finally, Figure~\ref{fig:correct_incorrect_token_usage_mean} shows the distribution of average token usage for cases that are all correct, all incorrect, and mixed responses. We observe that \OOne has a higher concentration of instances that are all correct towards shorter generations, which is a desirable behavior, while other models are more unpredictable across different instances.
\paragraph{Scaling effects with number of calls (parallel and sequential).}
\begin{figure}[t]
    \centering
    \begin{subfigure}[b]{0.32\textwidth}
        \centering
        \includegraphics[width=\textwidth]{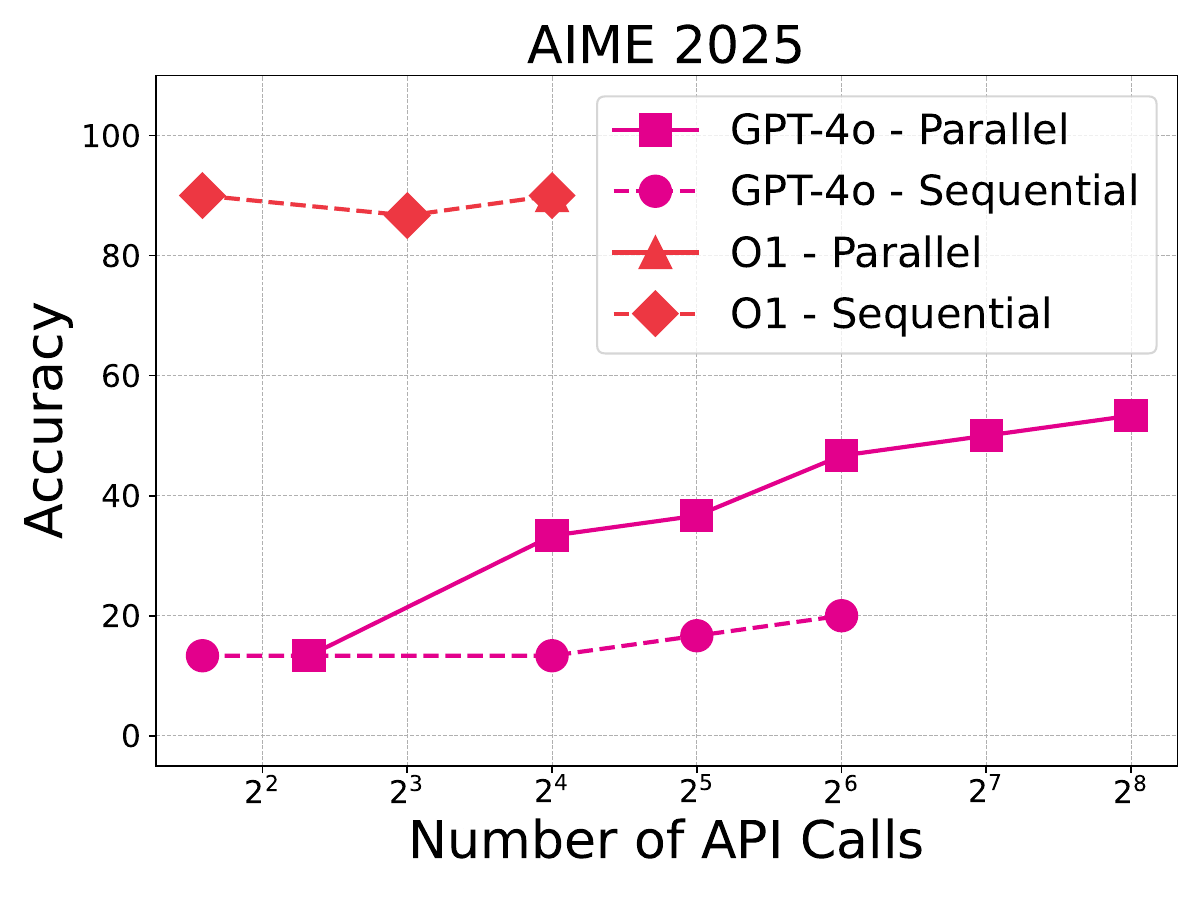}
    \end{subfigure}
    \begin{subfigure}[b]{0.32\textwidth}
        \centering
        \includegraphics[width=\textwidth]{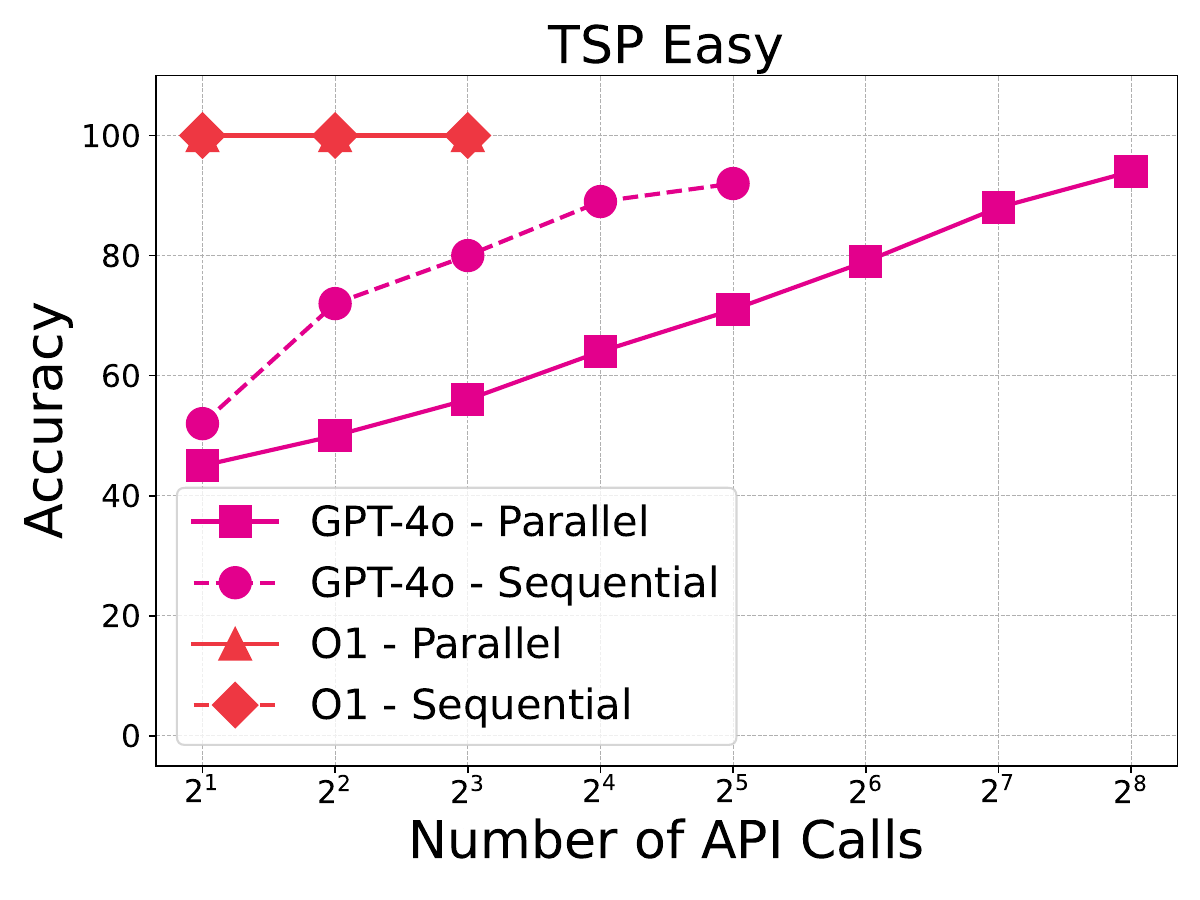}
    \end{subfigure}
    \begin{subfigure}[b]{0.32\textwidth}
        \centering
        \includegraphics[width=\textwidth]{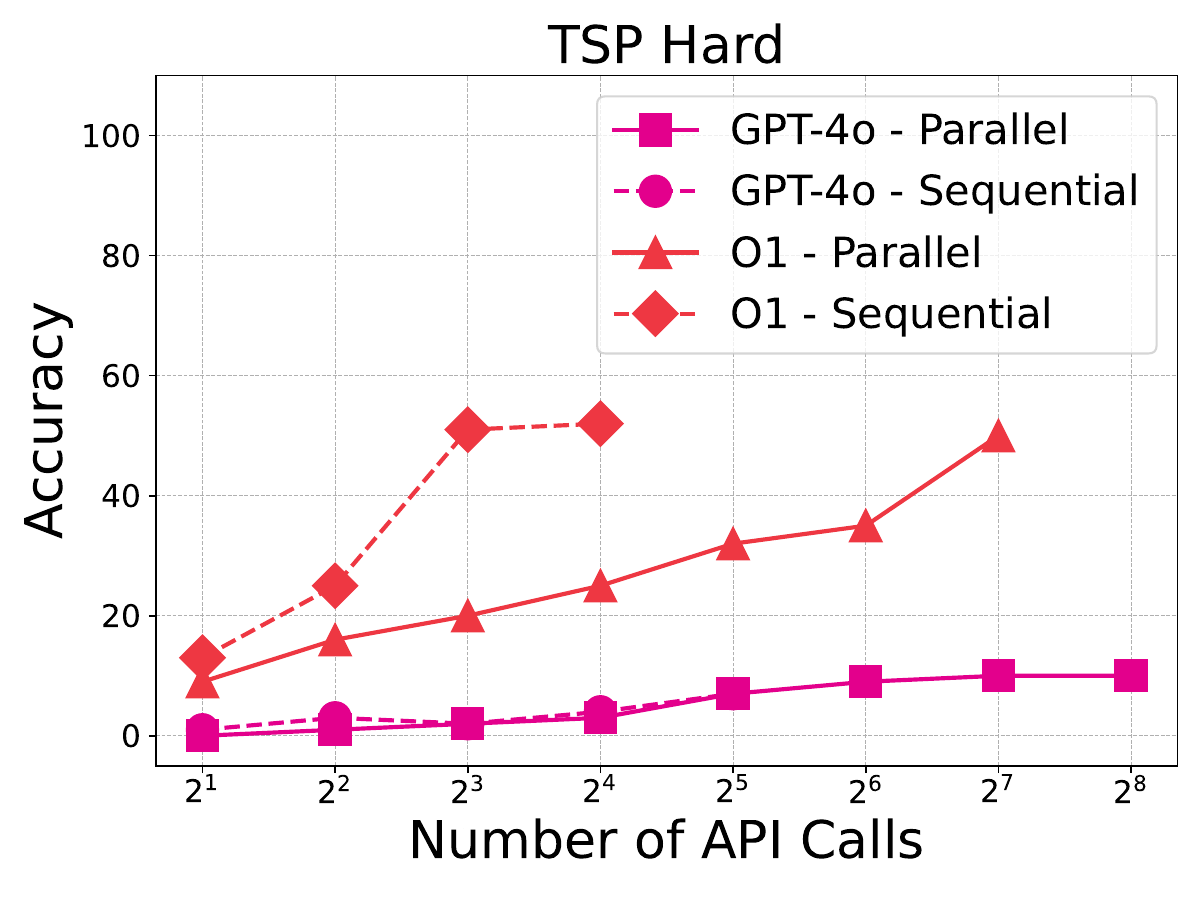}
    \end{subfigure}    
    \caption{Parallel and sequential scaling on AIME 2025 and TSP (best-of-n). The effectiveness of each approach highly depends on the downstream task. On AIME 2025, parallel scaling is more efficient than sequential scaling. On TSP however, sequential scaling appears to be more efficient. Scaling up is not helpful when the questions are extremely difficult.}
    \label{fig:scaleup}
\end{figure}

We investigated the effect of superscaling on performance through experiments on AIME 2025 and TSP, and two representative models: \OOne, as a model tuned for inference-time scaling, and \GPTFourO as a conventional model. Our goal was to measure how superscaling could improve GPT-4o’s performance on these tasks. For TSP, we selected two sets of different difficulty levels with 100 instances each. We evaluated both models by reporting the best-of-n accuracy under two superscaling settings: parallel and sequential (see Section \ref{benchmeth}). In the parallel setting, GPT-4o was scaled up to 256 calls, while in the sequential setting it was scaled up to 32 calls. 

The results (shown in Figure \ref{fig:scaleup}) indicate that superscaling substantially benefits GPT-4o on both AIME 2025 and the TSP easy dataset, with accuracy on the TSP easy set improving from 42\% to 95\%. On AIME 2025, \GPTFourO's best-of-n accuracy increases linearly with the log of model calls. Notably, \GPTFourO’s accuracy after superscaling nearly matches that of \OOne on the easy TSP instances, suggesting that the benefits of superscaling depend on problem complexity. In contrast, the TSP hard set did not show significant improvement even after superscaling, indicating that some tasks may remain challenging for conventional models regardless of test-time scaling efforts. Both superscaling methods consistently improve O1's performance, and the sequential approach with a hybrid verifier benefits this reasoning model more than parallel scaling with a perfect verifier, showcasing major encouragement for further scaling of even current reasoning models and \OOne's ability to adjust upon self-feedback. 

Furthermore, our comparison of parallel and sequential superscaling with \GPTFourO reveals that while parallel superscaling yields better gains for AIME 2025 than sequential superscaling, the latter provides meaningful improvements for the TSP easy dataset. A major difference between these benchmarks is the fact that in AIME, the success of the sequential approach also fundamentally relies on the ability of critic (the model itself in this case) to find flaws in the solution and give useful feedback. Naturally, it may be more difficult for \GPTFourO to do this for math than for easy TSP problems (smaller graphs).
\ifcolmsubmission
\vspace{-2mm}
\fi
\ifcolmfinal
\vspace{-2mm}
\fi
\section{Related Work}
Inference-time computation has led to many recent improvements in the performance of language models on reasoning tasks using longer generations \citep{wang2023selfconsistencyimproveschainthought, wei2022chain, yao2023treethoughtsdeliberateproblem}. Training models to take advantage of inference-time scaling is typically done through Reinforcement Learning (RL), where the model is optimized using a reward signal based on the correctness of its generated outputs. \cite{guo2025deepseek} showed that the chain-of-thought length can increase significantly during RL-based training. Self-training or distillation (if a strong teacher model is available) via supervised fine-tuning on reasoning traces has also been shown to be an effective alternative \citep{zelikman2022star, muennighoff2025s1, guo2025deepseek}. These approaches allow the model to learn to produce long reasoning chains without the computational overhead of full RL training.

Beyond explicit training, several other techniques have been proposed to improve model accuracy by leveraging additional compute during inference \citep{welleck2024from}. 
For example, sampling-based methods simulate test-time compute scaling by sampling generations from the same model more than once and selecting the final output using strategies such as majority voting \citep{wang2023selfconsistencyimproveschainthought, naik2024diversitythoughtimprovesreasoning}. Feedback-based methods to test-time compute provide step-wise feedback \citep{shinn2023reflexion, li2023making} or outcome-based feedback \citep{madaan2023self} to refine the model's generation. For example, \cite{zhao2025samplescrutinizescaleeffective} study the scalability of a simple sampling and self-verification technique and report a performance boost for Gemini v1.5 Pro beyond that of o1-preview. They also observe that self-verification continues to improve performance with scale even after majority-vote aggregation saturates. Finally, other approaches use explicit tree search over reasoning paths. For example, \cite{yao2023treethoughtsdeliberateproblem} and \cite{hao2023reasoning}, respectively, apply a global stepwise BFS / DFS search and Monte Carlo Tree Search over multiple reasoning paths.

Previous work on the evaluation of cost-accuracy trade-offs in test-time scaling has suggested the existence of inference scaling laws, with error rates steadily decreasing until saturation as the inference-compute increases, and emphasized the need for better verifiers in sampling-based strategies \citep{wu2025inferencescalinglaws, brown2024large}. It has also been shown that the effectiveness of different scaling approaches depends on the difficulty of the problem \citep{snell2024scaling, chen2024aremorellmcalls}. 
\section{Conclusion}
We present an extensive, empirical study of reasoning capabilities of nine foundation models across eight diverse benchmarks focusing on evaluating complex tasks that benefit from step-by-step problem solving. Going beyond aggregate performance and ranking, we analyze performance-cost tradeoffs, disaggregations, and failure patterns. Our results highlight that inference-time scaling improves performance but varies by domain and task complexity. Token use variability leads to cost nondeterminism and verification and feedback mechanisms hold untapped potential for improving model accuracy and reliability. Future directions include developing robust verifiers and adaptive token allocation strategies to enhance efficiency. Our findings offer insights into strengths, limitations, and paths for advancing inference-time scaling in large language models.
\ifcolmpreprint
\section*{Acknowledgements}
We would like to thank Ahmed Awadallah, Ece Kamar, Eric Horvitz, Rafah Hosn, Saleema Amershi for valuable discussions and guidance throughout the whole timeline of the project. We would also like to thank several colleagues and collaborators that have worked and brainstormed with us on different evaluation efforts, and have informed design and scientific choices we have made in this work: Adam Fourney, Arindam Mitra, Dimitris Papailiopoulos, Eduardo Salinas, Eric Price, Eric Zhu, Gagan Bansal, Gustavo de Rosa, James Woffinden-Luey, Katie Weissenfels, Michael Harrison, Oleg Losinets, Olli Saarikivi, Piero Kauffmann, Sahaj Agarwal, Shital Shah, Suriya Gunasekar, Vaish Shrivastava, Yanan Cai, and Xavier Fernandes. 
\fi
\ifcolmsubmission
\section*{Acknowledgements}
We would like to thank Ahmed Awadallah, Ece Kamar, Eric Horvitz, Rafah Hosn, Saleema Amershi for valuable discussions and guidance throughout the whole timeline of the project. We would also like to thank several colleagues and collaborators that have worked and brainstormed with us on different evaluation efforts, and have informed design and scientific choices we have made in this work: Adam Fourney, Arindam Mitra, Dimitris Papailiopoulos, Eduardo Salinas, Eric Price, Eric Zhu, Gagan Bansal, Gustavo de Rosa, James Woffinden-Luey, Katie Weissenfels, Michael Harrison, Oleg Losinets, Olli Saarikivi, Piero Kauffmann, Sahaj Agarwal, Shital Shah, Suriya Gunasekar, Vaish Shrivastava, Yanan Cai, and Xavier Fernandes. 
\fi
\clearpage
\section*{Reproducibility Statement}
\ifcolmsubmission
All experiments in this work were performed using a unified software framework for LLM evaluation which we will open to the community upon publication. Our framework enables reproducibility by storing all experiment configuration parameters, including prompt templates, pre-processing and post-processing operations, and evaluation metrics in text format for each individual experiment. We have included these config files and prompts in our Github repository for transparency and reproducibility. Github repository link is omitted to preserve anonymity. 
\fi
\ifcolmpreprint
All experiments in this work were performed using Eureka ML Insights, a unified and open-source software framework for LLM evaluation. Our framework enables reproducibility by storing all experiment configuration parameters, including prompt templates, pre-processing and post-processing operations, and evaluation metrics in text format for each individual experiment. We have included these config files and prompts in our Github repository for transparency and reproducibility.
\fi
\ifcolmfinal
All experiments in this work were performed using Eureka ML Insights, a unified and open-source software framework for LLM evaluation. Our framework enables reproducibility by storing all experiment configuration parameters, including prompt templates, pre-processing and post-processing operations, and evaluation metrics in text format for each individual experiment. We have included these config files and prompts in our Github repository for transparency and reproducibility.
\fi

Some of the key inference parameters that we used consistently in all experiments can be found in Table~\ref{tab:models}. Since the scope of the paper is to study inference-time scaling, all our experiments are conducted at a high temperature to ensure generation diversity. For \ROne, we use the recommended temperature by the model creators, which is 0.6 for complex reasoning.

The datasets used in this study are all publicly available. See Table~\ref{tab:datasets_repro} for links to access each of the datasets. 
\ifcolmsubmission
To preserve anonymity, the links to our contributed TSP and 3SAT datasets will be made available upon publication.
\fi
\ifcolmpreprint
The links to our contributed TSP and 3SAT datasets will be made available upon dataset release.
\fi

\paragraph{Note on experimental results.} Our experiments on \GeminiPro for the Maze and SpatialMap benchmarks were interrupted after four runs as the model was softly deprecated upon the release of Gemini 2.5 Pro on March 25, 2025. All other benchmarks instead include results for five runs for \GeminiPro. 

\ifcolmsubmission
Additionally, due to restrictive rate limiting for \ClaudeSonnetThinking (only 2-3 calls per minute) and the large size of some benchmarks, we were unable to complete five runs for Omni-MATH, Maze, and SpatialMap. All presented results for Omni-MATH are across three runs, while we do not currently report results for \ClaudeSonnetThinking on Maze and SpatialMap as only a single run was complete. We plan to complete and update all the above prior to publication. For all other benchmarks (AIME, GPQA, 3SAT, TSP, BA-Calendar) results for \ClaudeSonnetThinking include five runs.
\fi
\ifcolmpreprint
Additionally, due to restrictive rate limiting for \ClaudeSonnetThinking (only 2-3 calls per minute) and the large size of some benchmarks, we were unable to complete five runs for Omni-MATH, Maze, and SpatialMap. All presented results for Omni-MATH are across three runs, while we do not currently report results for \ClaudeSonnetThinking on Maze and SpatialMap as only a single run was complete. We plan to complete and update all the above in the next version of this report. For all other benchmarks (AIME, GPQA, 3SAT, TSP, BA-Calendar) results for \ClaudeSonnetThinking include five runs.
\fi
\begin{table}[htbp]
  \centering
  \small
  \caption{List of models studied in this paper and corresponding temperature and maximum token limits used for all experiments.}
    \begin{tabular}{lcrc}
    \toprule
    \bfseries{Model} & \bfseries{temp.} & \bfseries{max token} & \bfseries{reasoning} \\
    \hline
    \ClaudeSonnet 2024-10-22 \citep{ClaudeSonnet} & 1.0     & 4,096  & n \\
    \hline
    \ClaudeSonnetThinking 2025-02-19 \citep{Claude37Sonnet} & 1.0     & 65,536 & y \\
    \hline
    \ROne \citep{guo2025deepseek} & 0.6   & 65,536 & y \\
    \hline
    \GeminiPro Exp 2025-02-05 \citep{Gemini2Pro} & 1.0     & 4,096  & n \\
    \hline
    \GeminiFlash Exp 2025-01-21 \citep{GeminiFlash}& 1.0     & 32,768 & y \\
    \hline
    \OOne  2024-12-17  \citep{jaech2024openai}& NA    & NA    & y \\
    \hline
    \OThree 2025-01-31 (high) \citep{O3mini} & NA    & NA    & y \\
    \hline
    \GPTFourO 2024-08-06 \citep{hurst2024gpt} & 1.0     & 4,096  & n \\
    \hline
    \LlamaThreeOneLarge \citep{dubey2024llama} & 1.0     & 4,096  & n \\
    \bottomrule
    \end{tabular}%
  \label{tab:datasets_repro}%
\end{table}%

\begin{table}[htbp]
  \centering
  \small
  \caption{List of datasets studied in this paper and where to find them.}
    \begin{tabular}{lcrc}
    \toprule
    \bfseries{Dataset} & \bfseries{Link} \\
    \hline
    AIME 25~\citep{AIME25} &   \url{https://huggingface.co/datasets/lchen001/AIME2025}  \\
    \hline
    AIME 83-24~\citep{AIME8324} & \url{https://huggingface.co/datasets/di-zhang-fdu/AIME_1983_2024} \\
    \hline
    Omni-MATH~\citep{gao2024omni} & \url{https://huggingface.co/datasets/KbsdJames/Omni-MATH}     \\
    \hline
    GPQA$\mathbin{\Diamond}$~\citep{rein2024gpqa} & \url{https://huggingface.co/datasets/Idavidrein/gpqa}   \\
    \hline
    BA-Calendar~\citep{butt2024benchagents} & \url{https://huggingface.co/datasets/microsoft/ba-calendar}    \\
    \hline
    TSP-Opt (new benchmark) & To be released    \\
    \hline
    3SAT-Search (new benchmark) & To be released    \\
    \hline
    Maze~\citep{wang2024picture} & \url{https://huggingface.co/datasets/microsoft/VISION_LANGUAGE}  \\
    \hline
    SpatialMap~\citep{wang2024picture} & \url{https://huggingface.co/datasets/microsoft/VISION_LANGUAGE}  \\
    \bottomrule
    \end{tabular}%
  \label{tab:models}%
\end{table}%
\section*{Ethics Statement}
This work studies the impact of inference-time scaling on a diverse set of complex tasks that can benefit from step-by-step solutions. The work however does not include other types of problems that require social or commonsense reasoning, or reasoning about ethics and safety in complex social situations in the real world. While there have been informal statements about how inference-time scaling can benefit these problems as well, it is not clear whether such improvements in recent models originate from inference-time scaling and extended scratchpads or rather from enhanced RLHF training. Disentangling these effects is important for better understanding the dynamics of different post-training stages and can only be conducted via ablation studies that have access to the different models before and after training for inference-time scaling, as well as before and after RLHF tuning. 

A similar open question is whether current techniques can also address issues with information fabrication and lack of factuality. Although better reasoning skills could help with eliciting information in retrieval augmented generation (RAG) scenarios, studies that rigorously quantify such effects are still lacking. 

Lastly, The technical terminology for describing inference-time scaling effects is still evolving. In several cases, researchers have described extended and longer step-by-step generations as longer ``chains of thought'', and the process itself as ``thinking'' or reasoning. However, such terminology carries the risk of anthropomorphizing model behavior~\citep{devrio2025taxonomy}, which is largely considered harmful in the community since it fuels human over-reliance on models (i.e. human reliance on models even when they are incorrect)~\citep{kim2024m,passi2022overreliance}. In this work, we distinguish between models that have been tuned for inference-time scaling vs. not, and interchangeably refer to models with lengthened step-by-step scratchpads as reasoning models given the implicit assumption that such models are generally better at tasks that require more complex reasoning, as we show to be the case in this study.
\clearpage
\bibliography{colm2025_conference}
\bibliographystyle{colm2025_conference}
\clearpage
\appendix
\section{AIME - High-School Exam for Olympiad Qualification}
\label{sec:aime}
\noindent\textbf{Motivation:} AIME, or American Invitational Mathematics Examination, is a high-school mathematics competition held every year since 1983. It has been widely used to evaluate ``reasoning'' capabilities of foundation models and test-time scaling techniques. Thus, it is important to obtain a holistic understanding of how different models and test-time scaling methods perform on AIME. 
\\
\noindent\textbf{Benchmark description:} We leverage two AIME instances. One is a subset\footnote{\url{https://huggingface.co/datasets/di-zhang-fdu/AIME_1983_2024}} of questions collected from 1983 to 2024, which contains 933 questions in total. Another is the 2025 new exam containing 30 questions in total\footnote{\url{https://huggingface.co/datasets/lchen001/AIME2025}} , released on February 2025. Each question is an open-form math problem, and the correct answer is an integer. With the recent 2025 edition of the competition in February, differences between model performance in 2025 and in previous years is often also used as a proxy to generalization skills in the math domain.
\begin{figure}[t]
    \centering
    \begin{subfigure}[b]{0.42\textwidth}
        \centering
        \includegraphics[width=\textwidth]{figures/AIME_charts/analysis_report/NumericMatch_result_overall_accuracy_bar_chart.pdf}
    \end{subfigure}
    \begin{subfigure}[b]{0.42\textwidth}
        \centering
        \includegraphics[width=\textwidth]{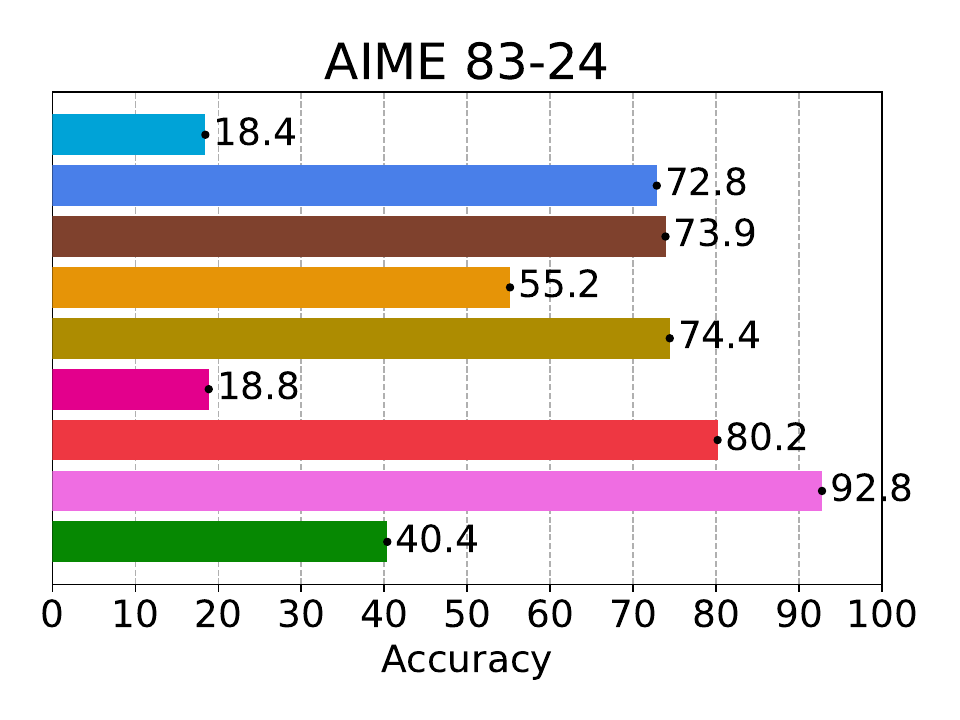}
    \end{subfigure}
        \begin{subfigure}[b]
    {\textwidth}
        \centering
\includegraphics[width=\textwidth]{figures/model_legend.png}
    \end{subfigure}

    \caption{Overall model performance for AIME 2025 and AIME 83-24. }
    \label{fig:aime_both_aggregate_Accuracy}
\end{figure}
\begin{figure}[t]
    \centering
    \includegraphics[width=1.0\linewidth]{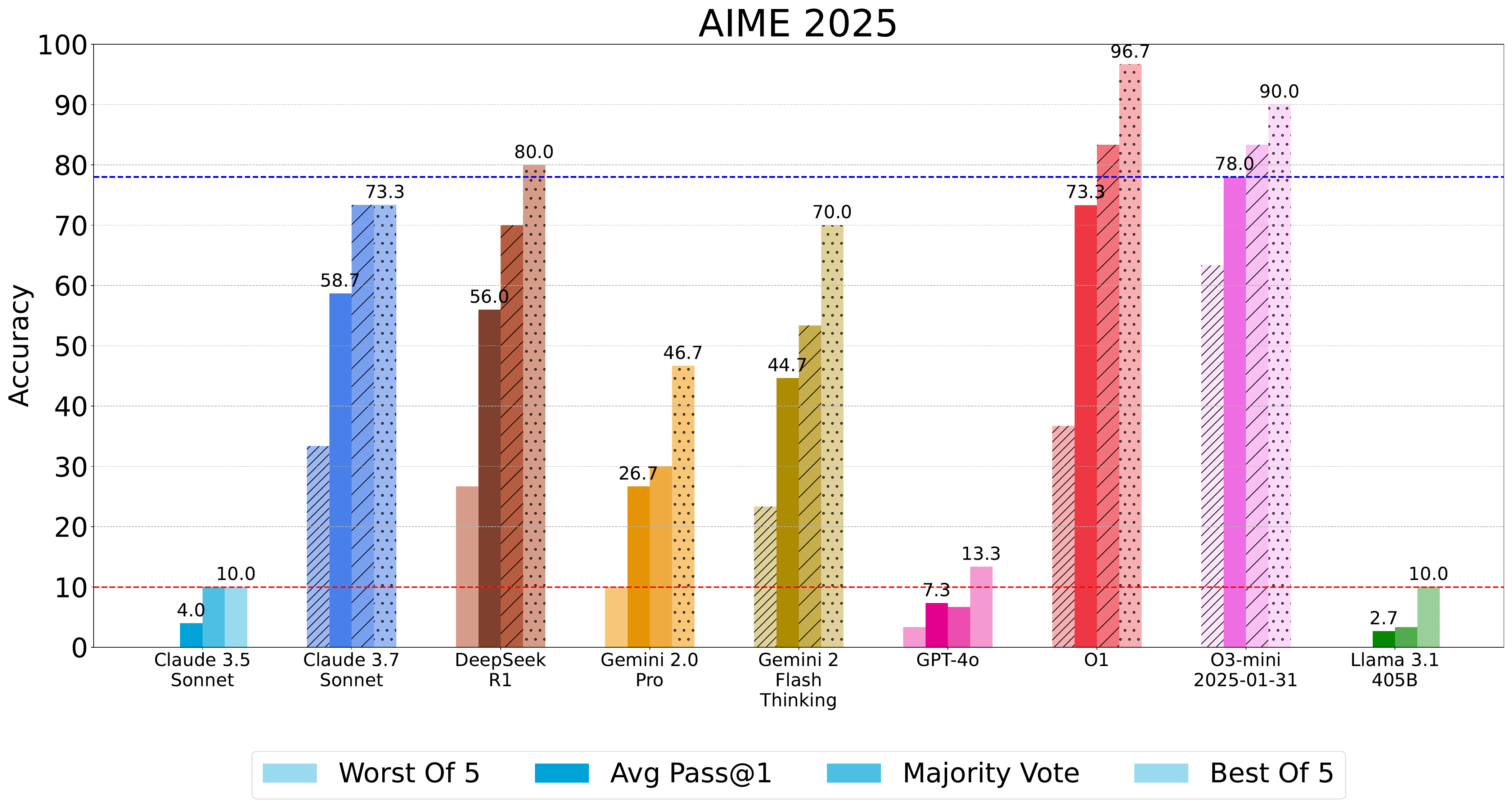}
    \caption{Results on AIME 2025 with different aggregations by parallel scaling over 5 runs. The \textcolor{red}{red} line indicates the lowest best-of-5 accuracy observed across all models, while the \textcolor{blue}{blue} line represents the highest average pass@1 accuracy.}
    \label{fig:aime_parallel}
\end{figure}

\begin{figure}[t]
    \centering
    \includegraphics[width=1.0\linewidth]{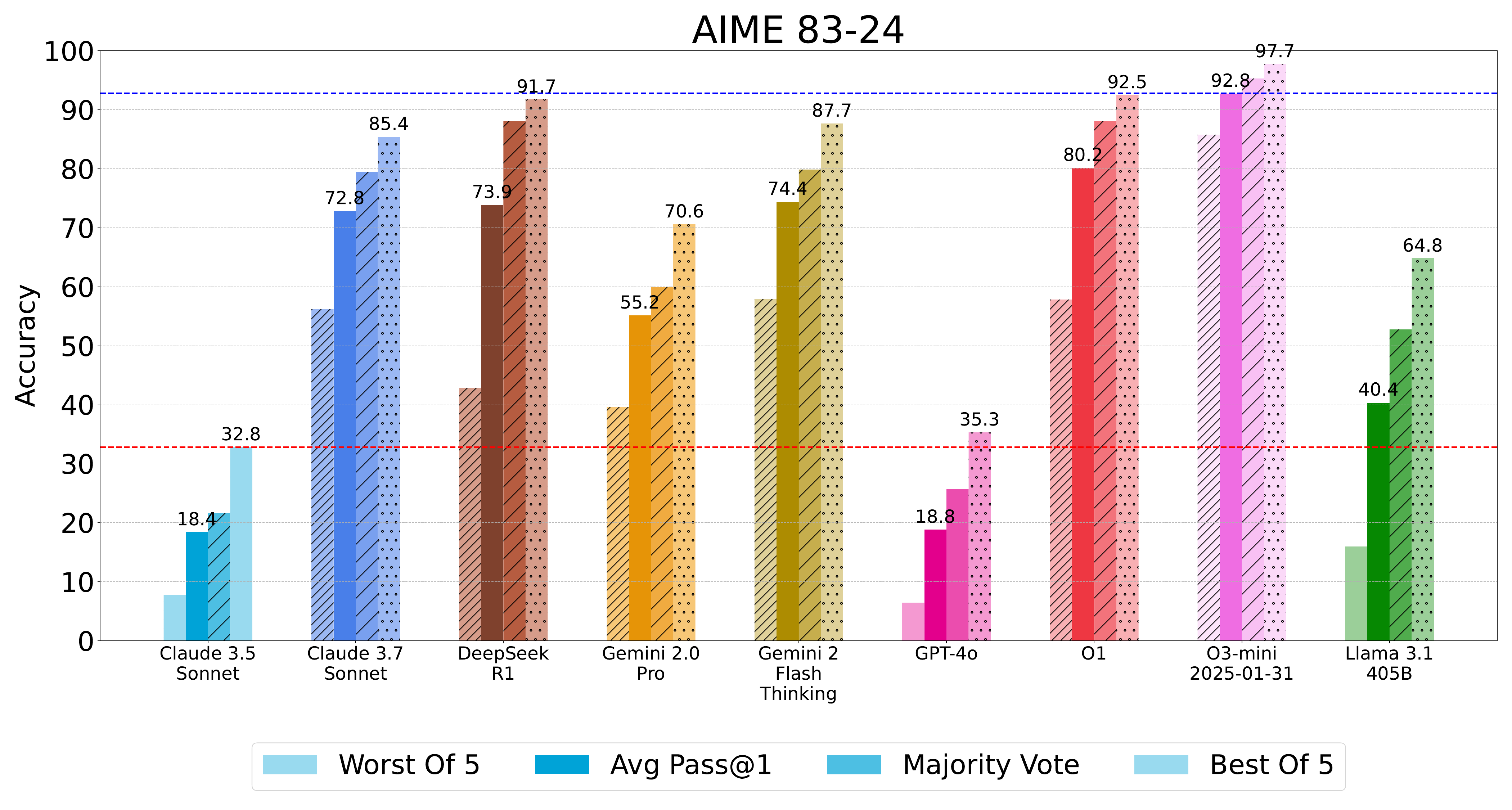}
    \caption{Results on AIME 83-24 with different aggregations by parallel scaling over 5 runs. The \textcolor{red}{red} line indicates the lowest best-of-5 accuracy observed across all models, while the \textcolor{blue}{blue} line represents the highest average pass@1 accuracy.}
    \label{fig:aime_parallel_8324}
\end{figure}

\subsubsection*{Main takeaways}
\noindent\fbox{%
    \parbox{\textwidth}{%
        \begin{itemize}[leftmargin=*]
\item Across all models, inference-time scaling's performance drops substantially on the newly released test. From conventional models in particular, the average of five runs using Llama 3.1 405B was 40\% on questions collected from 1983 to 2024, but only 1\% on the 2025 questions. Models like \OOne, \OThree, \ROne, \GeminiFlash and \ClaudeSonnetThinking also exhibit 7\%-30\% drop in performance, with \OOne showing the smallest drop (7\%). This suggests that existing inference-time scaling methods are also likely to overfit on the development datasets. 
    \item All models benefit from best-of-5 verification in AIME 2025, including reasoning models, which shows that there is still remaining opportunity for further improvement. This suggests that leveraging a high-quality verifier can substantially improve the existing test-time scaling approaches. 
    \item Longer generation is not always better. For example, \ROne consumes tokens 10 times more than \ClaudeSonnetThinking (Figure~\ref{fig:all_in_one_accuracy_tokens}), but its accuracy is even slightly lower. How to perform reasoning ``efficiently'' remains an open question.
    \item Equipped with a high-quality aggregator, test-time scaling's performance can scale log-linearly with the amount of test-time computation without model retraining or fine-tuning. In fact, the best-of-n's accuracy increases linearly with respect to the log of model calls with GPT-4o.
        \end{itemize}
    }%
}
\clearpage
\section{Omni-MATH - Olympiad Math}
\label{sec:omnimath}
\vspace{-5mm}
\begin{figure}[t]
    \centering
    \begin{subfigure}[b]{0.48\textwidth}
        \centering
        \includegraphics[width=\textwidth]{figures/Omni_Math/analysis_report/OmniMath_correctness_overall_accuracy_bar_chart.pdf}
    \end{subfigure}
    \begin{subfigure}[b]{0.42\textwidth}
        \centering
        \includegraphics[width=\textwidth]{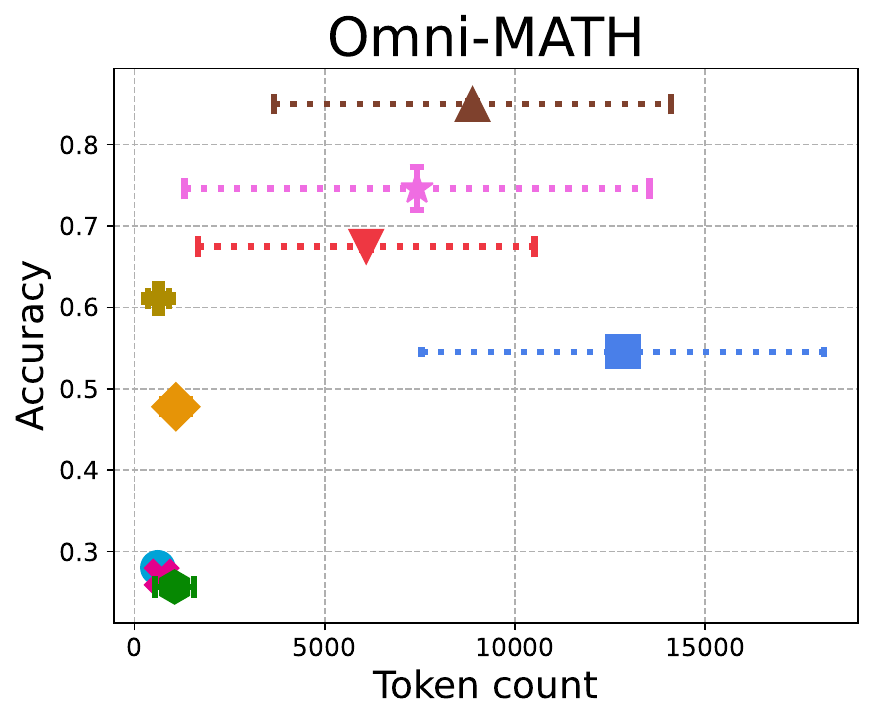}
    \end{subfigure}
        \begin{subfigure}[b]
    {\textwidth}
        \centering
\includegraphics[width=\textwidth]{figures/accuracy_tokens_legend_2lines.png}
    \end{subfigure}
    \caption{Omni-MATH overall performance and token usage.}
    \label{fig:omni_math_overall}
\end{figure}
\noindent\textbf{Motivation:} The AIME benchmark has been widely utilized for evaluating the mathematical reasoning capabilities of models. However, this prevalent use raises concerns about models potentially overfitting to this specific dataset. To address this issue, we evaluate on an additional math benchmark Omni-MATH, a comprehensive benchmark designed to assess large language models' (LLMs) mathematical reasoning abilities across a broader spectrum of problems. By incorporating Omni-MATH, we aim to study the generalization of reasoning models to diverse mathematical datasets, as it encompasses a larger and more varied collection of competition-level problems.  \\
\noindent\textbf{Benchmark description:} Omni-MATH is a meticulously curated dataset comprising 4,428 competition-level mathematical problems, specifically tailored to evaluate LLMs' proficiency in Olympiad-level reasoning. Unlike existing benchmarks, Omni-MATH focuses exclusively on mathematics, offering a nuanced analysis of model performance across various disciplines and complexity levels. The problems are categorized into 33 distinct sub-domains and span 10 difficulty levels, reflecting a hierarchical classification of mathematical domains. The dataset sources its problems from a wide range of international competitions, ensuring a diverse and challenging set of questions. Each problem is accompanied by a detailed solution, facilitating comprehensive evaluation and analysis.
\\
\noindent\textbf{Model performance:} Figure~\ref{fig:omni_math_overall} presents overall results for the Omni-MATH dataset. Here, we observe slight variations in model performance rankings when compared to AIME. While O1 and O3-mini-high are the best performing models on AIME, we see that Deepseek R1 outperforms them to be the best performing model on Omni-MATH. This indicates better generalization with the R1 model to diverse and open-ended math problems. The trends of non-reasoning models are similar to AIME with GPT 4o, Claude 3.5 Sonnet and Llama 3.1 405B performing on-par with each other. Aggregate token usage show similar trends to model performance, with Deepseek R1, O1 and O3-mini-high using orders of magnitude more tokens than their non-reasoning counterparts. These models also have the largest variance in token usage. Breaking down performance by different topics, we observe that the reasoning models have larger performance boosts in categories like Number Theory and Algebra, while they lag slightly behind in Geometry and Discrete Math. 

\begin{figure}
    \centering
    \includegraphics[width=1.0\linewidth]{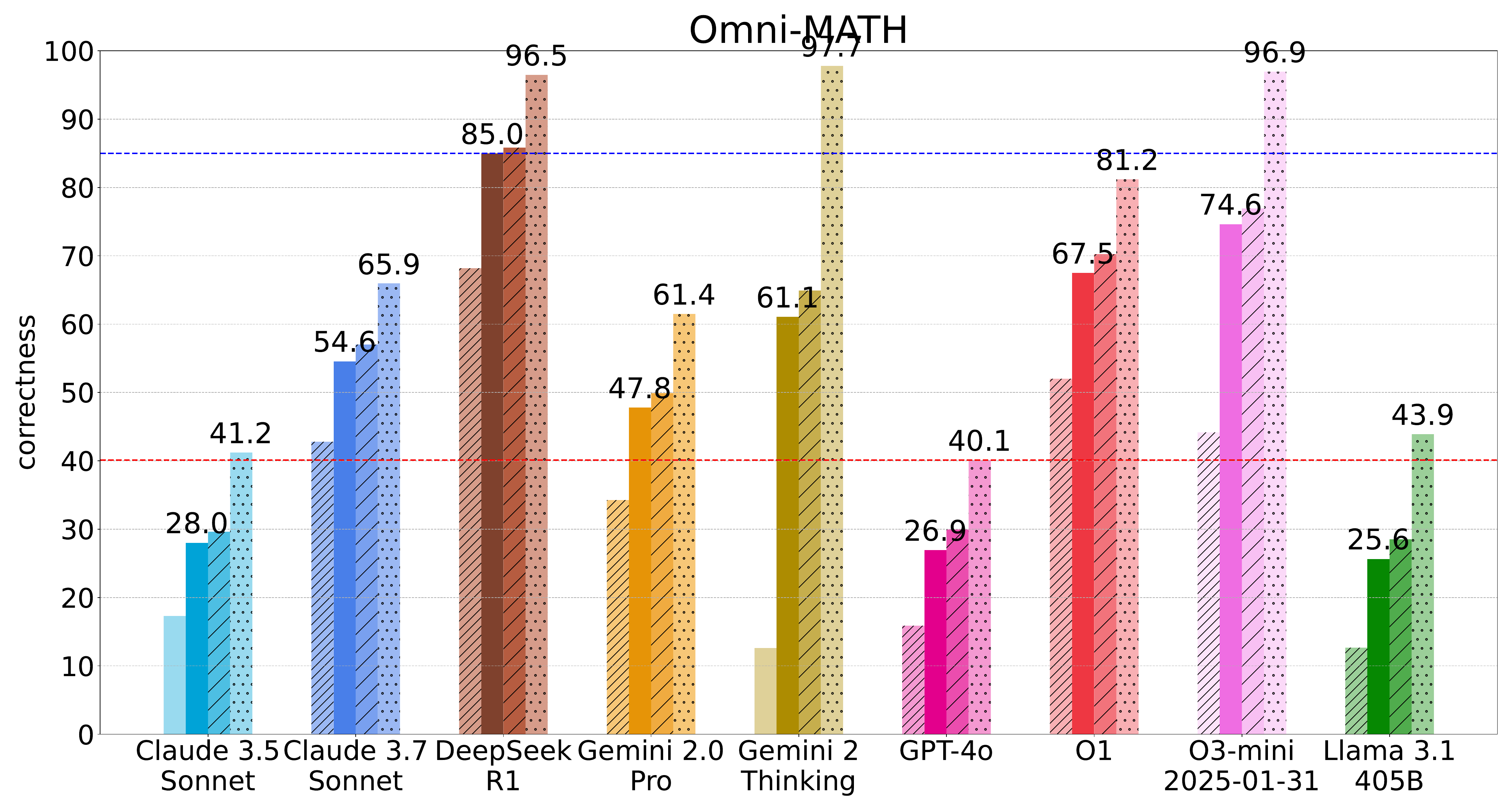}
    \includegraphics[width=0.8\textwidth]{figures/agg_legend.png}
    \caption{Results on Omni-MATH with different aggregations by parallel scaling over 5 runs. The \textcolor{red}{red} line indicates the lowest best-of-5 accuracy observed across all models, while the \textcolor{blue}{blue} line represents the highest average pass@1 accuracy.}
    \label{fig:omni_math_inf_parallel}
\end{figure}

\begin{figure}[t]
    \centering
    \begin{subfigure}[b]{0.45\textwidth}
        \centering
        \includegraphics[width=\textwidth]{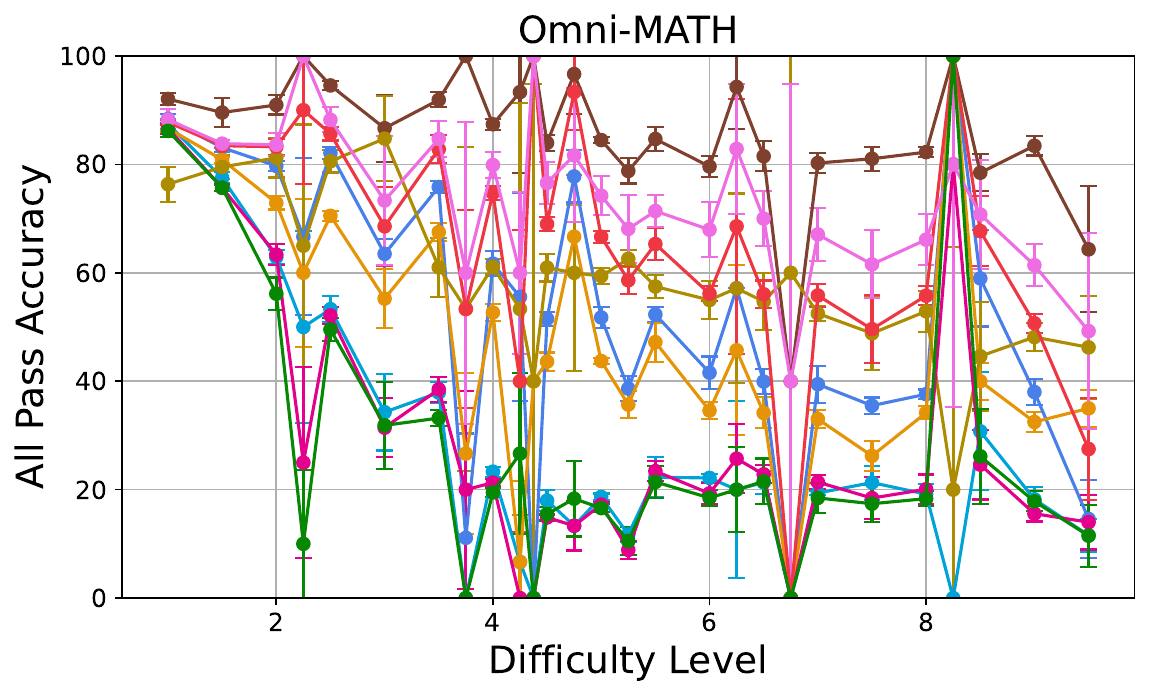}
    \end{subfigure}
    \begin{subfigure}[b]{0.45\textwidth}
        \centering
        \includegraphics[width=\textwidth]{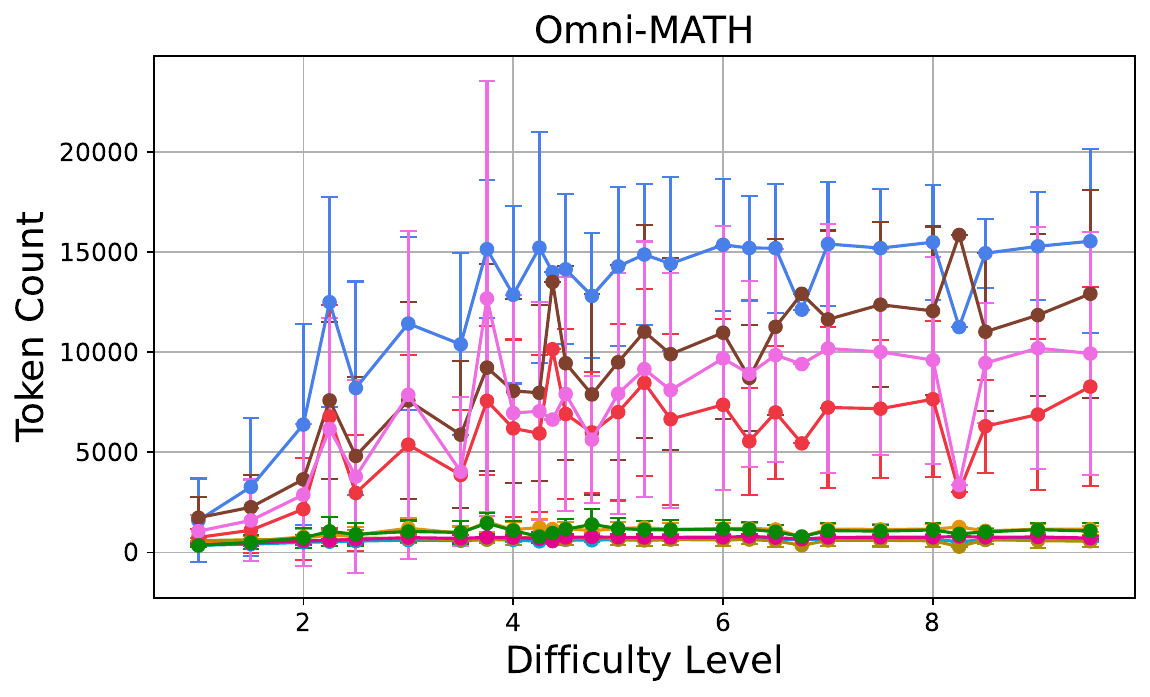}
    \end{subfigure}
        \begin{subfigure}[b]
    {\textwidth}
        \centering
\includegraphics[width=\textwidth]{figures/model_legend.png}
    \end{subfigure}

    \caption{Omni-MATH performace and token usage by problem difficulty level.}
    \label{fig:omni_math_tok_perf}
\end{figure}

\noindent\textbf{Performance vs. token usage tradeoffs:} Figure~\ref{fig:omni_math_tok_perf} breaks down performance and token usage by problem difficulty. We reproduce graphs presented in \OOne~\citep{jaech2024openai} and \OThree~\citep{O3mini}, where reasoning models increase token use for harder problems and correspondingly see a declining trend in model performance. In contrast, non-reasoning models have sharper decline in model performance with a flat token usage irrespective of difficulty level indicating that these models are unable to adapt to problem difficulty.
\\

\begin{figure}[t]
    \centering
    \includegraphics[width=1.0\linewidth]{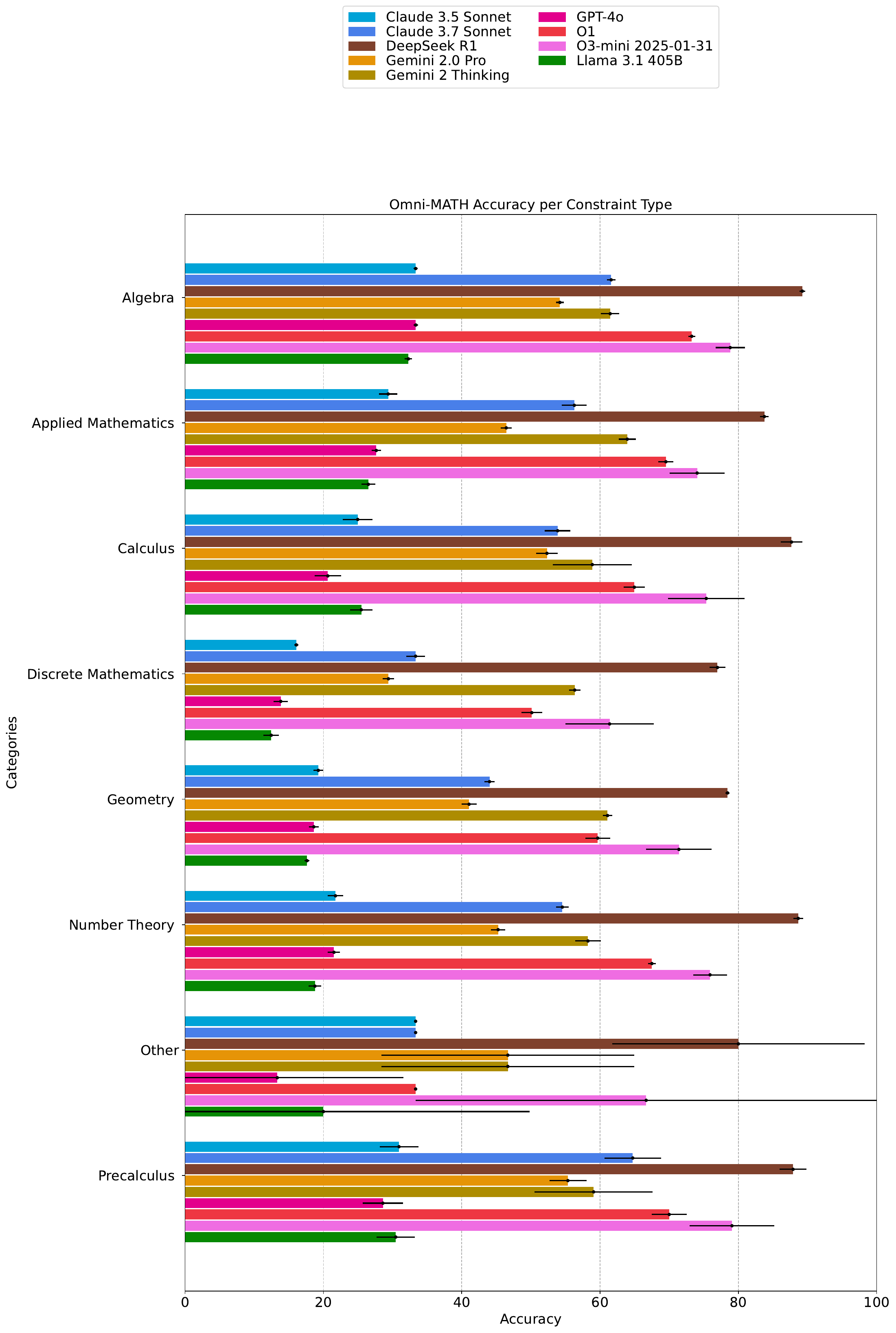}
    \caption{Omni-MATH topic-level accuracy.}
    \label{fig:omni_math_const_acc}
\end{figure}

\subsubsection*{Main takeaways}
\noindent\fbox{%
    \parbox{\textwidth}{%
        \begin{itemize}[leftmargin=*]
        \item Reasoning models significantly outperform non-reasoning models, among which \ROne is better than \OThree. These models are extremely good at Calculus and Number Theory problems but lag in Geometry and Discrete Math.
        \item There is a wide gap between the best reasoning model (highest pass@1) and a hypothetical model potentially be trained to verify and select the best outcome from the model with the lowest best-of-n, suggesting that additional math specific training is essential to equip base models to reason about more complex problems.
    \item  Gemini Flash thinking provides best tradeoff of token cost v/s accuracy - it provides better reasoning performance with significantly lower costs.
        \end{itemize}
    }%
}
\clearpage
\section{GPQA Diamond - Scientific Reasoning}
\label{sec:gpqa}
\vspace{-2mm}
\begin{figure}[t]
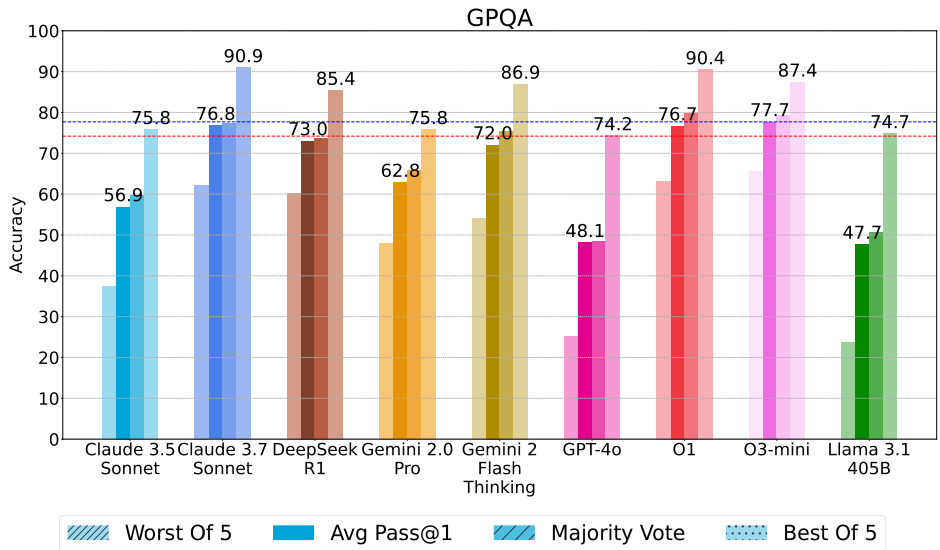

    \centering
    \includegraphics[width=0.9\linewidth]{figures/GPQA_charts/inf_analysis_report/ExactMatch_result_inference_comp_accuracy_bar_chart.pdf}
    \includegraphics[width=0.8\linewidth]{figures/agg_legend.png}
    \caption{Results on GPQA with different aggregations by parallel scaling over 5 runs. The \textcolor{red}{red} line indicates the lowest best-of-5 accuracy observed across all models, while the \textcolor{blue}{blue} line represents the highest average pass@1 accuracy. The \textcolor{red}{red} line indicates the lowest best-of-5 accuracy observed across all models, while the \textcolor{blue}{blue} line represents the highest average pass@1 accuracy.}
    \label{fig:gpqa_parallel}
\end{figure}

\begin{figure}[t]
    \centering
    \begin{subfigure}[b]{0.45\textwidth}
    \includegraphics[width=\linewidth]{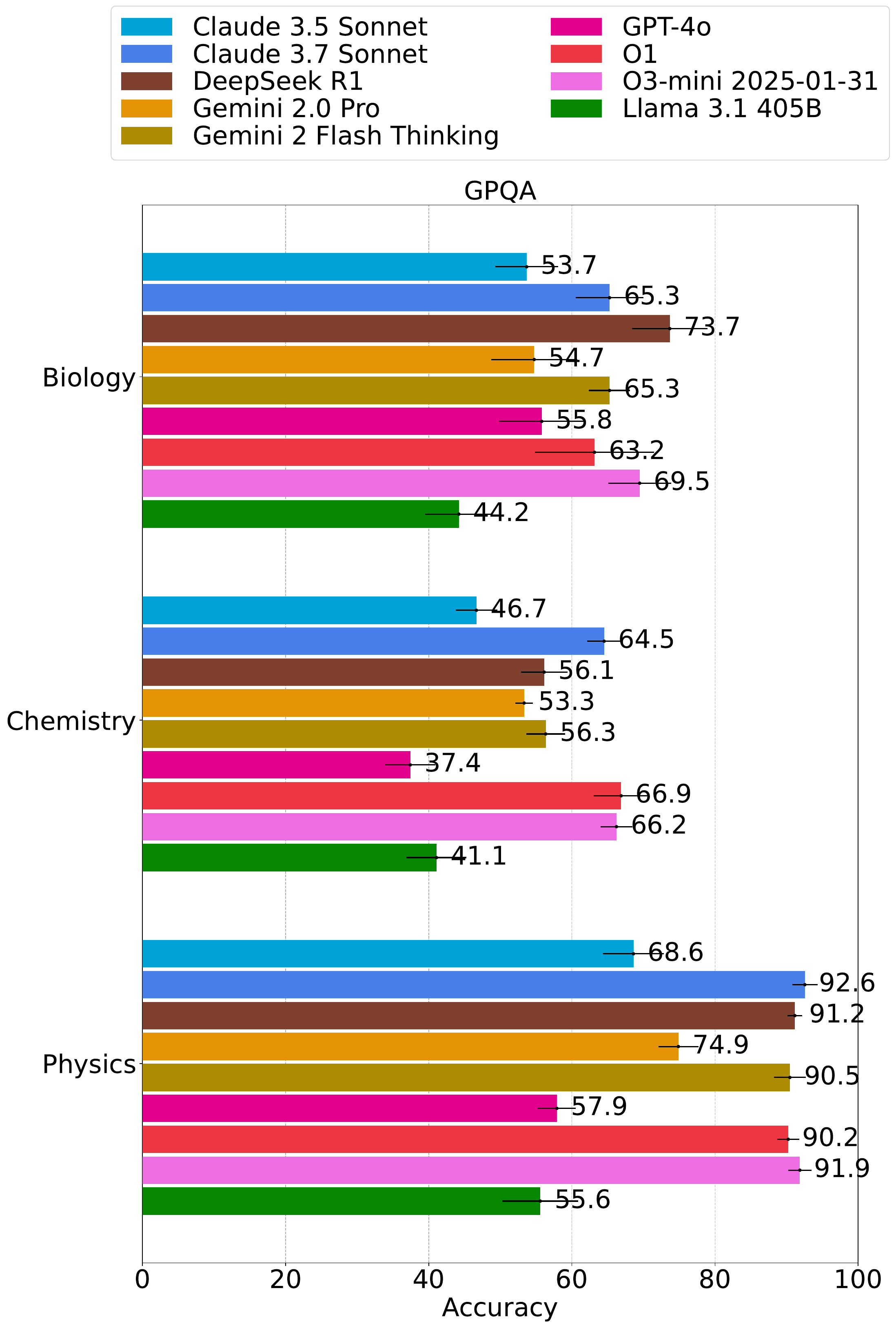}
    \end{subfigure}
        \begin{subfigure}[b]{0.45\textwidth}
    \includegraphics[width=\linewidth]{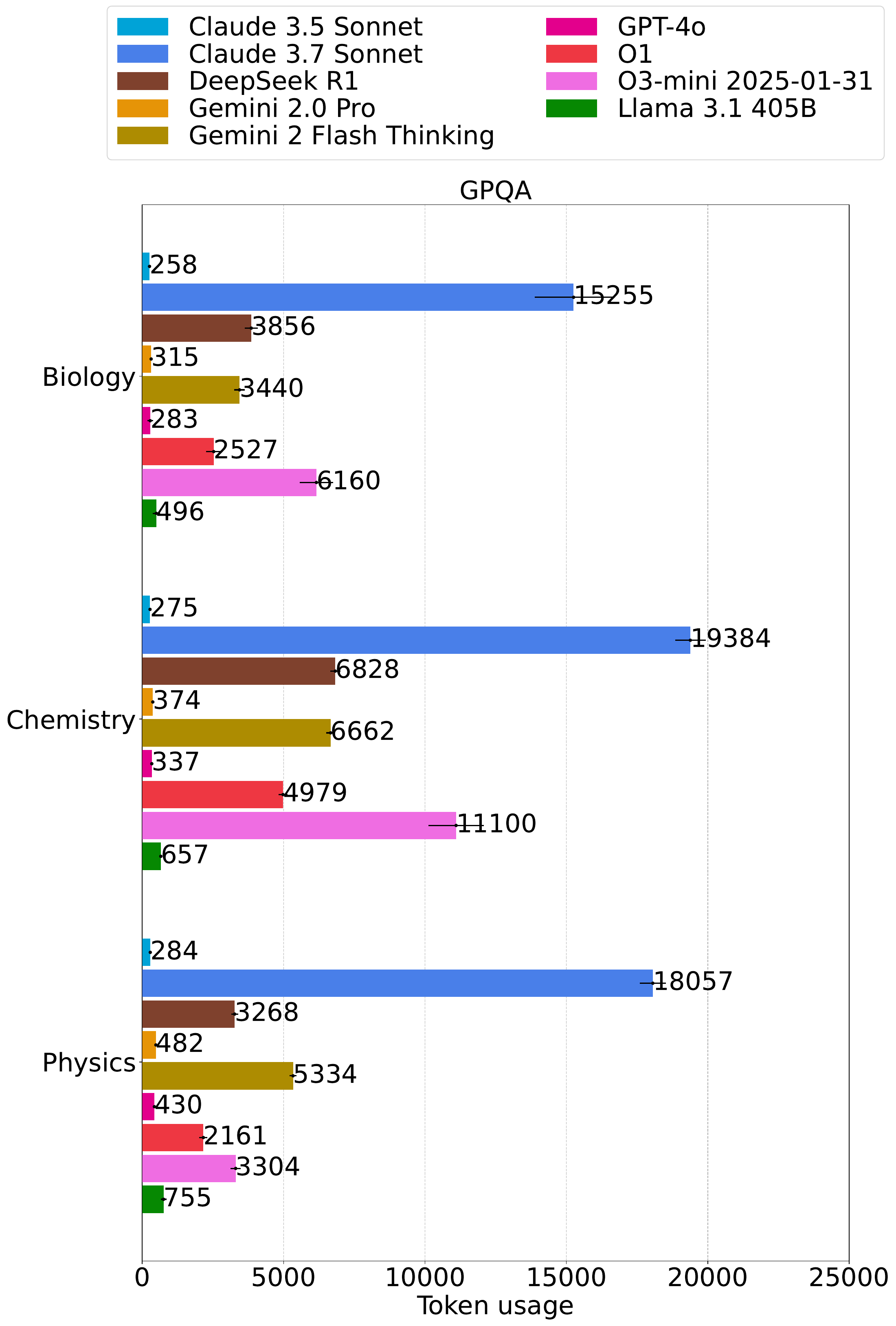}
    \end{subfigure}
    \caption{GPQA accuracy and token usage by high-level domain. Standard deviations for token usage are computed across five repeats, within the same high-level domain.}
    \label{fig:gpqa_domain}
\end{figure}
\noindent\textbf{Motivation:} GPQA was first introduced by \cite{rein2024gpqa} to assess the models' scientific knowledge in physics, biology, and chemistry. However, since it consists of challenging problems for which people with a corresponding PhD in the domain can only achieve up to 74\% accuracy (in the whole set), the benchmark has been recently used to also demonstrate that step-by-step scratchpads can generalize beyond math and coding. We include GPQA here to draw parallels with previous reports and landmark results, but also to check the consistency of generalization claims from math to more general scientific reasoning.

\noindent\textbf{Benchmark description:} From the initial set, we use GPQA Diamond, for which 2/2 experts involved in question writing agree on the problem definition and answer and $\leq$ non experts can solve the problem correctly. This accounts for a total of 198 questions for the diamond subset, where experts achieve an accuracy of 81.3\%. Despite the challenging nature of the benchmark, there are several caveats to keep in mind in this analysis. First, the benchmark is relatively small. In particular there are fewer than 100 questions per domain (86 in Physics, 93 in Chemistry, 19 in Biology). Second, even though the problems are deemed as challenging, there is no available difficulty level assigned to each question for calibration. 

\noindent\textbf{Model performance:} As shown in Figure~\ref{fig:all_in_one}, \ClaudeSonnetThinking, \OOne, and \OThree perform very similarly to each other, in a 76\%-78\% range. They are followed by \ROne and \GeminiFlash, which perform in a 72\%-73\% range. However, when we break down performance by high-level domain beyond overall model ranking, all reasoning models seem to benefit a lot from step-by-step solutions in Physics, but they still lag behind in Chemistry and Biology (Figure~\ref{fig:gpqa_domain}). In fact, the gap between Physics and other domains is more than 25\%. A possible explanation for this can be attributed to the fact that many of the problems in the Physics domain require several simpler mathematical steps as part of the solution, while the Biology and Chemistry problems seem less dependent on math skills and potentially more dependent on knowledge or domain-specific steps (e.g., breaking down a chemical reaction). The finding indicates that current inference-scaling methods may not always generalize as well for other scientific domains.  

In addition, when looking at parallel scaling effects for 5 runs (Figure~\ref{fig:gpqa_parallel}), we observe that the conventional-to-reasoning gap (i.e. the gap between the red and blue line) is very small. This indicates that even a conventional model is highly likely to produce an inference path that is as accurate as the best reasoning model. This also indicates that current improvements could have been replicated with simpler post-training and RL techniques, that do not require fine-grained reflection, but rather reflection on whole inference paths of conventional models. At the same time, given that none of the current models performs well outside of Physics, breakthroughs in other domains beyond Physics will still require more than harvesting several inference path.

\noindent\textbf{Performance vs. token usage tradeoffs:} Figure~\ref{fig:all_in_one_accuracy_tokens} shows that \ClaudeSonnetThinking spends ~3x more tokens than \OThree, which in turn spends ~2x more tokens than \OOne, while all these models perform in a very similar accuracy range. This indicates that token efficiency is still an area that requires significant optimization, and that lengthened generations do not always lead to a better result. The finding is also relevant within generations from the same model. Figure~\ref{fig:gpqa_domain} shows that all reasoning models spend more tokens on Chemistry problems. Yet, this is not sufficient for being accurate, but it rather seems a symptom of models struggling with finding good solutions.
\vspace{-3mm}
\subsubsection*{Main takeaways}
\noindent\fbox{%
    \parbox{\textwidth}{%
        \begin{itemize}[leftmargin=*]
        \item Inference-time scaling does not benefit all domains equally. All reasoning models perform more than 25\% better in Physics than Biology and Chemistry.
        \item Longer generation traces for Biology and Chemistry do not lead to higher accuracy for reasoning models.
        \item There is a very narrow gap between the worst observed best-of-5 score of conventional models and the average of 5 runs for reasoning models. This shows that having access to a stronger verifier at post training time that can extract good full inference paths from conventional models would lead to a  model that performs similarly to the state of the art in reasoning models today.
        \end{itemize}
    }%
}

\section{3SAT - Satisfiability}
\label{sec:3st}
\noindent\textbf{Motivation:}  Algorithmic problems provide a precise and structured way of assessing specific reasoning skills in models, unlike domains such as mathematics, where the exact skills being tested can be ambiguous. Additionally, algorithmic tasks allow for easy manipulation and clear control over problem difficulty, making them ideal for benchmarking reasoning capabilities in a systematic manner.

One fundamental algorithmic skill that we expect reasoning models to possess is search --- the ability to systematically enumerate potential solutions until the correct one is found. To rigorously evaluate this capability, we use the classic 3SAT problem~\citep{karp2009reducibility}. In the search version of 3SAT, we are given a Boolean formula in conjunctive normal form, where each clause consists of exactly three literals, and the task is to find an assignment of variables that satisfies all clauses. Since the search version of 3SAT is NP-Hard, solving difficult instances requires exponential time unless P=NP. This makes it a natural benchmark for assessing (exhaustive) search capabilities, as even the best-known algorithms must effectively enumerate solutions in hard cases, where search space pruning is not effective.

Beyond serving as a testbed for search, 3SAT serves as a fundamental building block for solving many real-world constraint satisfaction problems. SAT solvers are widely used in hardware verification (checking circuit correctness), scheduling (allocating resources under constraints), and software testing (symbolic execution and bug detection). Thus, assessing a model's performance on 3SAT also provides insight into its broader applicability to real-world problems.

\noindent\textbf{Benchmark description:} We use randomly generated 3SAT instances. Each clause is constructed by first selecting three distinct variables uniformly at random and then independently negating each variable with probability 0.5. The difficulty of randomly generated 3SAT instances primarily depends on two factors:
\begin{enumerate}[leftmargin=*,itemsep=0em]
    \item The number of variables ($n$). Larger instances require higher computational effort.
    \item The ratio of the number of clauses ($m$) to the number of variables ($n$). A low clause-to-variable ratio leads to underconstrained problems, which typically have many satisfying solutions and are easier to solve. Conversely, a high clause-to-variable ratio results in overconstrained problems, usually unsatisfiable and easier for algorithms that detect inconsistencies quickly.
\end{enumerate}

It has been observed that the hardest random 3SAT instances occur around a critical clause-to-variable ratio of approximately 4.26 \citep{mitchell1992hard,  kirkpatrick1994critical, cheeseman1991really}. Instances around this threshold pose significant difficulties for SAT solvers, both empirically and theoretically, as indicated in classic studies on phase transitions in random satisfiability problems (see the excellent exposition by \cite{achlioptas2009random} for more details).

Based on this, we generate the benchmark by varying the number of variables from 4 to 15. For each variable count, we generate:
\begin{enumerate}[leftmargin=*,itemsep=0em]
        \item 20 hard instances at the critical threshold (\( m/n \approx 4.26 \)).
        \item 20 easy underconstrained instances at half the threshold (\( m/n \approx 2.13 \)).
        \item 20 easy overconstrained instances at double the threshold (\( m/n \approx 8.52 \)).
        \item 20 uniquely satisfiable hard instances, where the instance is at the hard threshold (\( m/n \approx 4.26 \)) but with exactly one satisfying assignment. These are the most challenging for tested models.
\end{enumerate}
In total, we evaluate on 960 instances across different difficulty levels and variable counts.

A similar evaluation on random 3SAT instances was recently conducted by \cite{hazra2024can}, focusing primarily on GPT-4o. Their results show that GPT-4o struggles significantly on hard instances near the critical threshold. Our evaluation includes both general-purpose language models and models specifically trained for reasoning. We find that reasoning-focused models perform substantially better than their non-reasoning counterparts, particularly on the most challenging cases.

\begin{figure}[t]
    \centering
    \includegraphics[width=0.48\linewidth]    
    {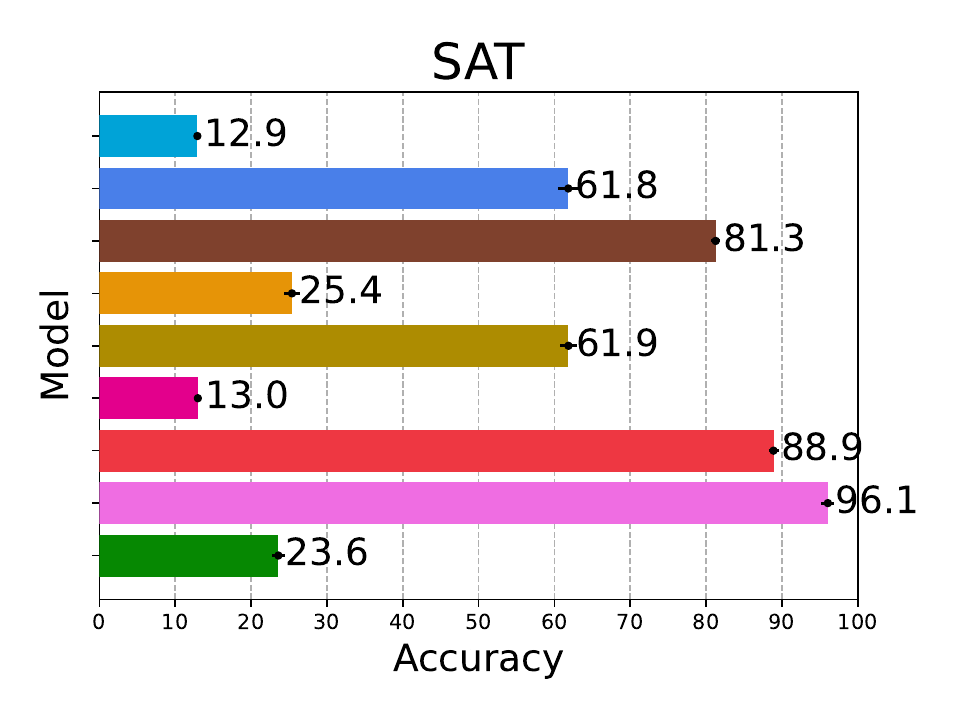}    
    \includegraphics[width=0.42\linewidth]{figures/NPHard_charts/SAT_Analysis_2025-03-19/usage_report/accuracy_vs_tokens_data_point_aggregates.pdf}
    \\
    \includegraphics[width=\textwidth]{figures/model_legend.png}
    \caption{SAT overall performance and token usage. The left figure shows overall model performance across nine models. The right figure shows pareto tradeoff between accuracy and token usage for all benchmarks.}
    \label{fig:sat_overall_acc_token_usages}
\end{figure}

\noindent\textbf{Model performance:} Figure \ref{fig:sat_overall_acc_token_usages} presents the mean accuracy achieved by different models. The O3-mini model performs best with an accuracy of 96.1\%, followed by the O1 and DeepSeek R1 models, which achieve 88.9\% and 81.3\%, respectively. There is a clear performance gap between models that use test-time scaling and those that do not. For instance, Claude 3.5 Sonnet achieves only 12.9\% accuracy—significantly lower than its test-time scaling counterparts.

\begin{figure}[t]
    \centering
    \includegraphics[width=0.48\linewidth]{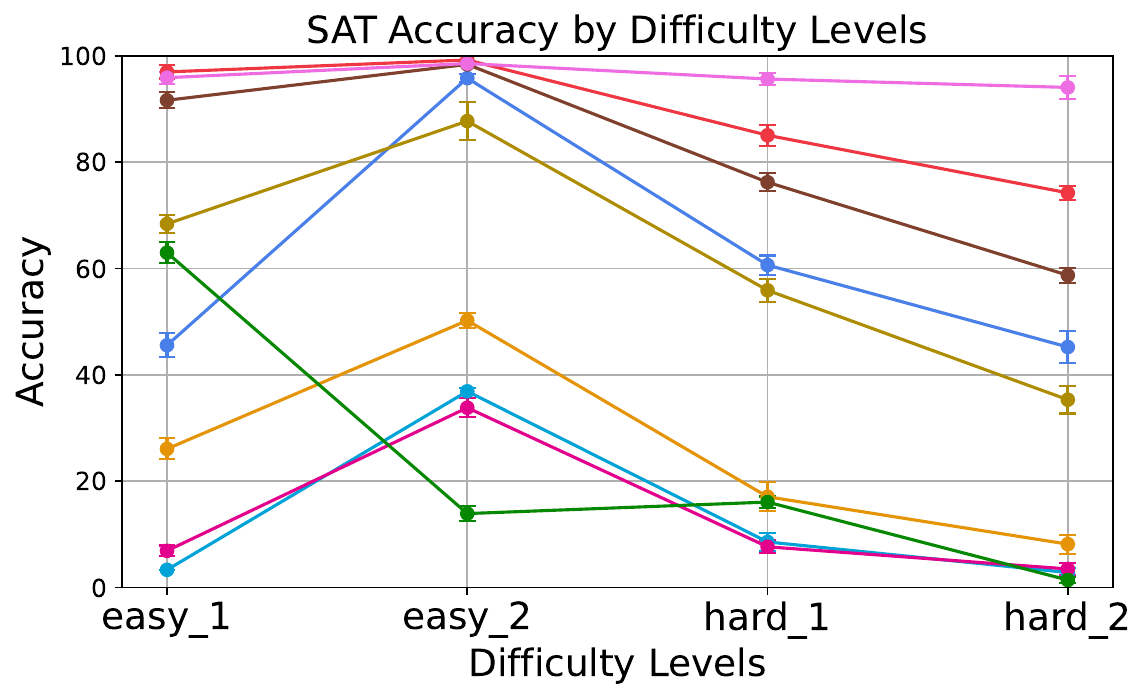}
    \includegraphics[width=0.48\linewidth]{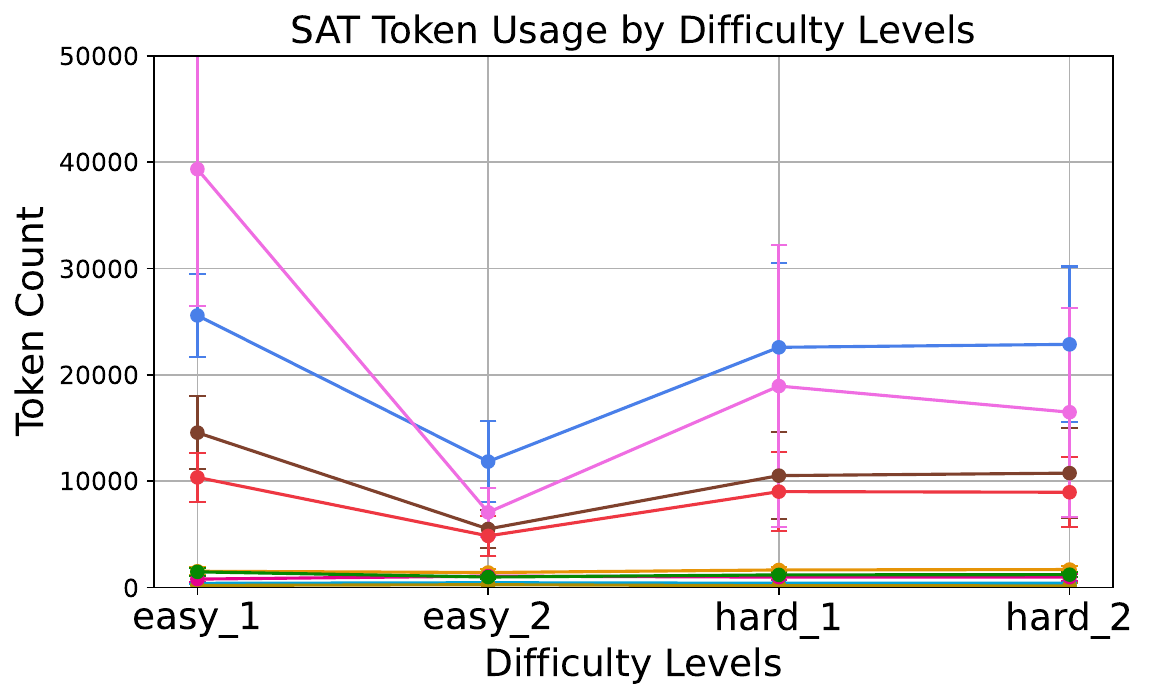}
    \\
    \includegraphics[width=\textwidth]{figures/model_legend.png}    
    \caption{SAT accuracy and token usage by difficulty level. There are four levels corresponding to: easy\_1 (easy overconstrained), easy\_2 (easy underconstrained), hard\_1  (hard multiple solutions), and hard\_2 (hard single solution). The left figure shows accuracy, while the right figure shows token usage at each level. Note that test-time scaling models consistently outperform non-test-time scaling models across all four difficulty levels. Moreover, non-test-time scaling models perform worse on easy overconstrained problems than on easy underconstrained problems. A possible explanation is that non-test-time scaling models often produce a true/false result even when the problem is unsatisfiable.}
    \label{fig:sat_accuracy_tokens_difficulty_ver1}
\end{figure}

\begin{figure}[t]
    \centering    
    \includegraphics[width=0.49\linewidth]{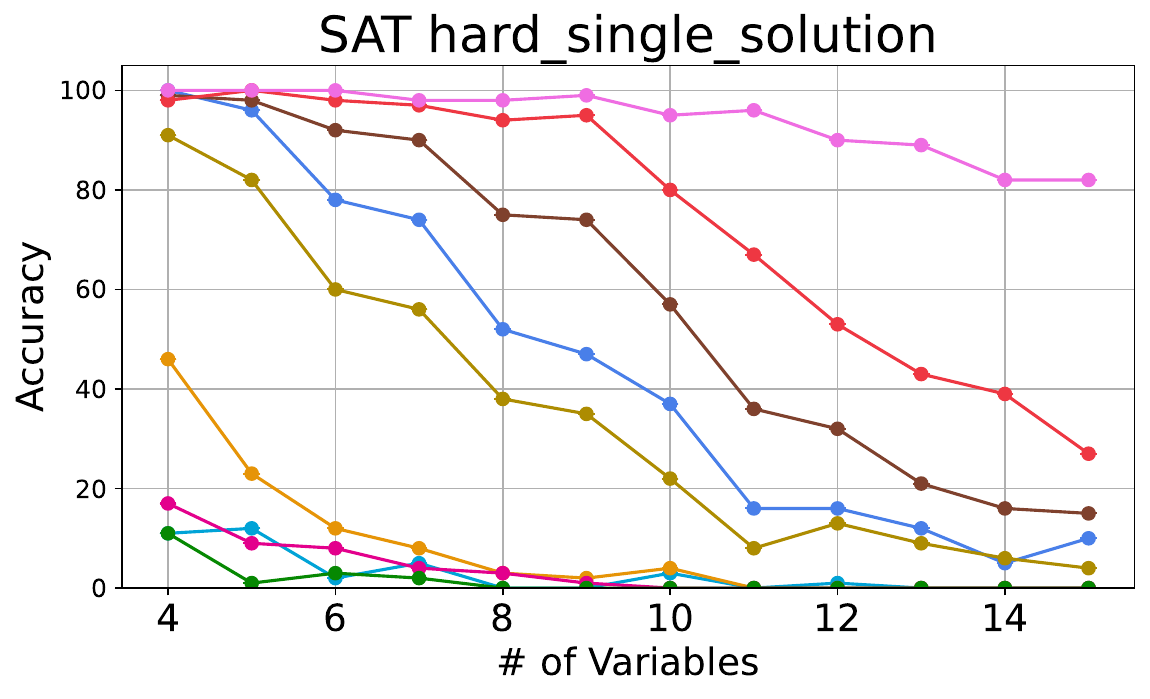}
    \includegraphics[width=0.49\linewidth]{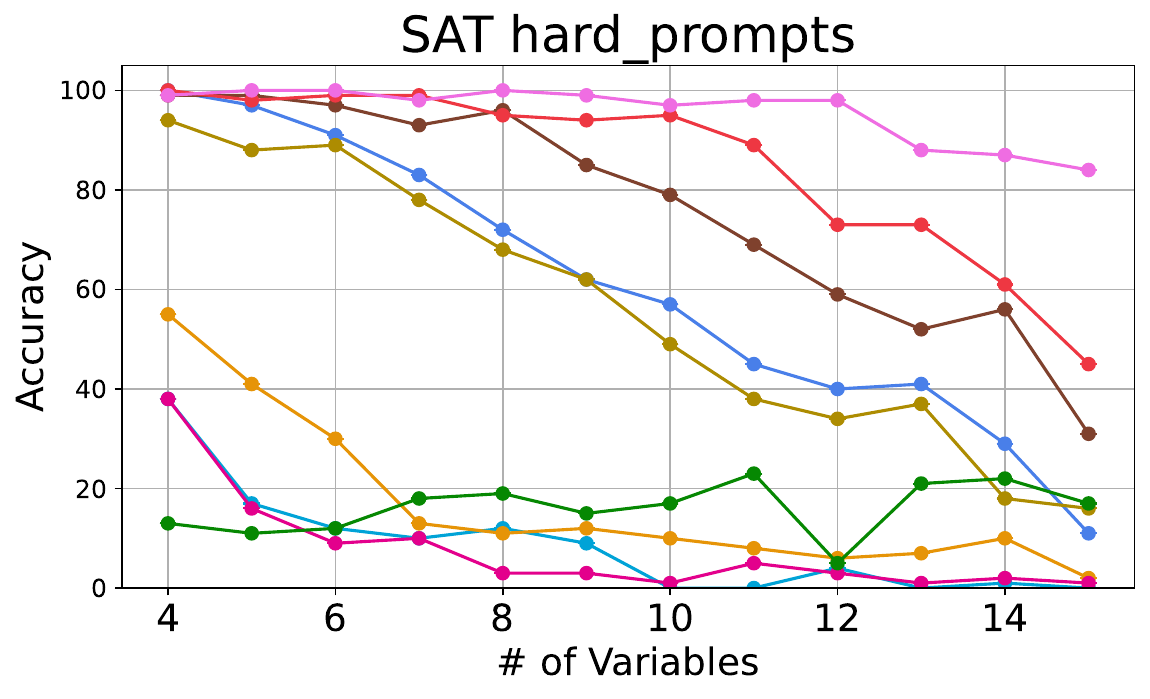}    
    \\
    \includegraphics[width=0.49\linewidth]{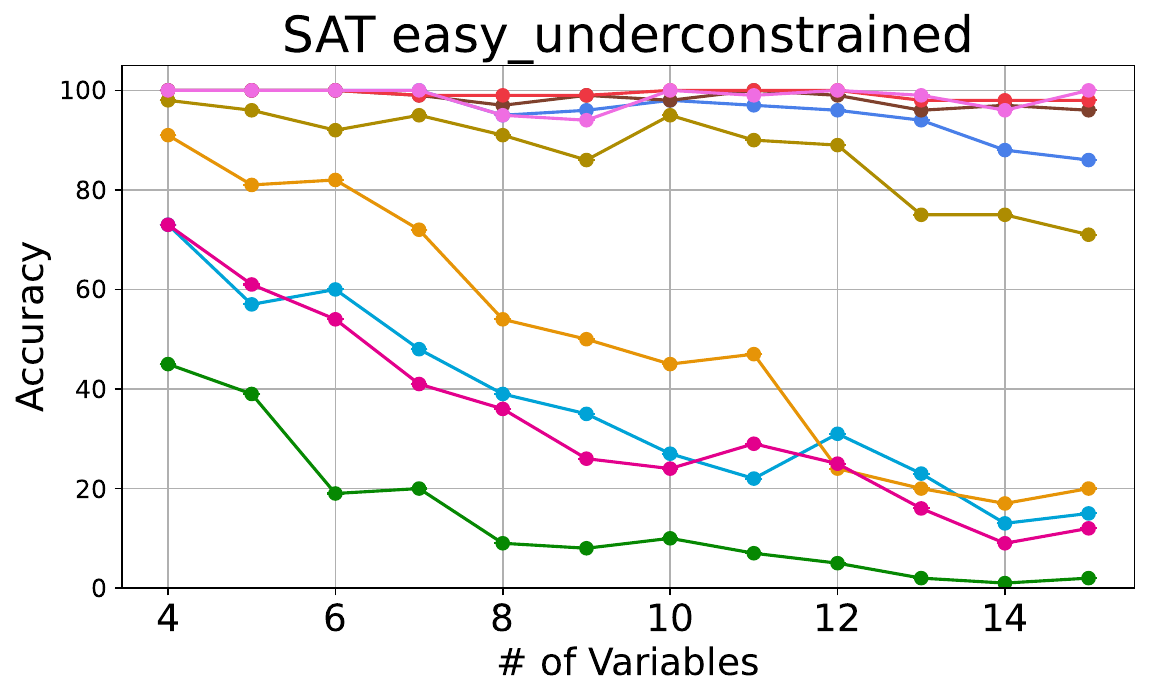}
    \includegraphics[width=0.49\linewidth]{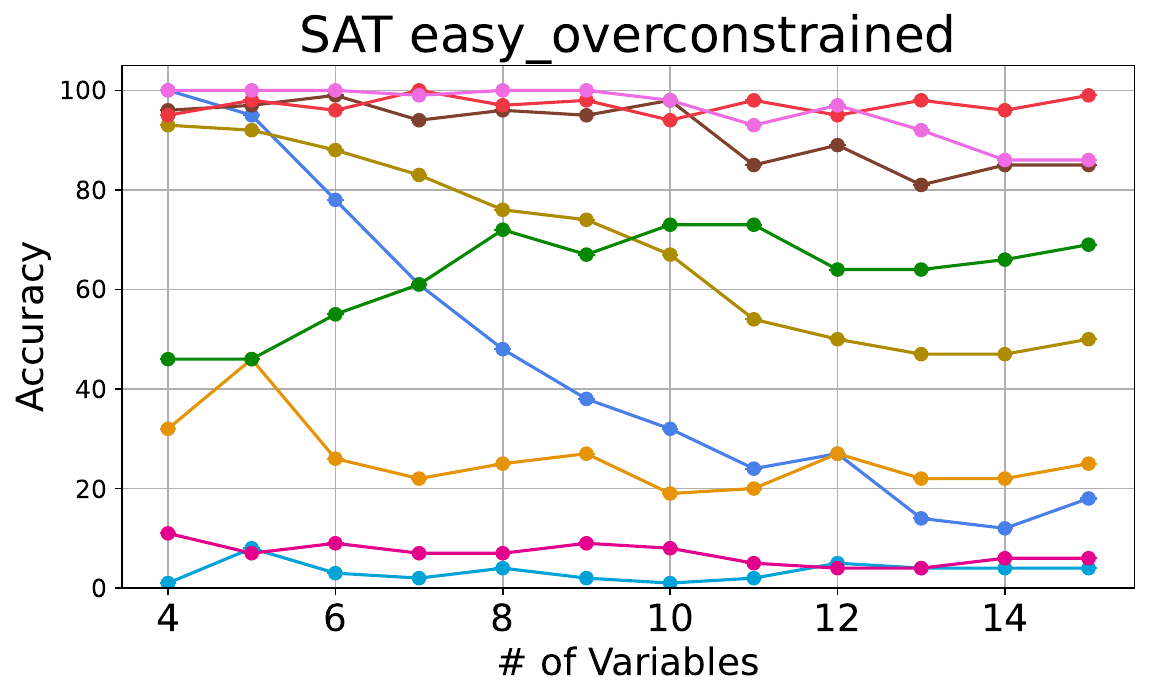}  
    \\
    \includegraphics[width=\textwidth]{figures/model_legend.png}
    \caption{3SAT accuracy across four difficulty levels—easy underconstrained, easy overconstrained, hard (single solution), and hard (multiple solutions). Each figure plots accuracy against the number of variables. Notably, even test-time scaling models experience a significant performance drop once the number of variables exceeds 10 in the hard-solution setting.}
    \label{fig:sat_accuracy_variables_difficulty_ver2}
\end{figure}

Figures \ref{fig:sat_accuracy_tokens_difficulty_ver1} shows model accuracy across different difficulty levels. There are four levels corresponding to: easy\_1 (easy overconstrained), easy\_2 (easy underconstrained), hard\_1  (hard multiple solutions), and hard\_2 (hard single solution). The left figure shows accuracy, while the right figure shows token usage at each level. Note that test-time scaling models consistently outperform non-test-time scaling models across all four difficulty levels. Moreover, non-test-time scaling models perform worse on easy overconstrained problems than on easy underconstrained problems. A possible explanation is that non-test-time scaling models often produce a true/false result even when the problem is unsatisfiable. By contrast, test-time scaling models use additional tokens to verify solutions, leading to more accurate outcomes. Additionally, we also highlight SAT accuracy with respect to number of variables in Figure \ref{fig:sat_accuracy_variables_difficulty_ver2}. 
Notably, even test-time scaling models experience a significant performance drop once the number of variables exceeds 10 in the hard-solution settings.


\noindent\textbf{Performance vs. token usage tradeoffs:} Figure \ref{fig:sat_accuracy_tokens_difficulty_ver1} shows average token usage for different models. Test-time scaling models generally take more tokens than no test-time scaling models. More difficult problems requires more tokens. However, more tokens do not necessarily mean higher accuracy even for test-time scaling models probably due to increase in difficulty levels.

\begin{figure}[t]
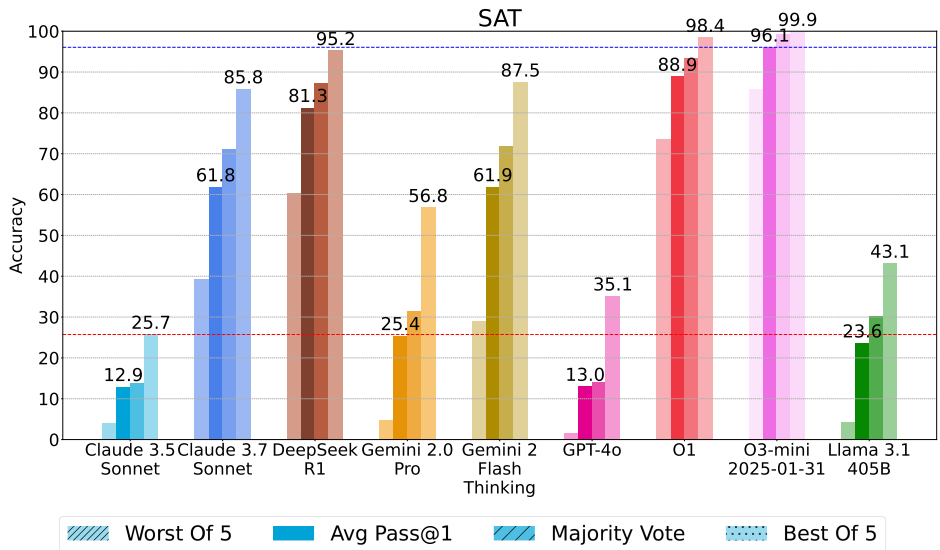

    \centering
    \includegraphics[width=0.9\linewidth]{figures/NPHard_charts/SAT_Analysis_2025-03-19/bon_analysis_report/NPHardSATMetric_result_inference_comp_accuracy_bar_chart.pdf}
    \includegraphics[width=0.8\textwidth]{figures/agg_legend.png}
    \caption{Results on 3SAT with different aggregations by parallel scaling over 5 runs. The \textcolor{red}{red} line indicates the lowest best-of-5 accuracy observed across all models, while the \textcolor{blue}{blue} line represents the highest average pass@1 accuracy. 
    }
    \label{fig:sat_bofn_wofn_parallel}
\end{figure}

\noindent\textbf{Scaling effects:} Figure \ref{fig:sat_bofn_wofn_parallel} illustrates the Best-of-N, Worst-of-N, and average performance for various models. A key observation is the substantial improvement in accuracy under the Best-of-5 setting, with most models showing gains of 10 to 15 percentage points. This suggests that the correct answer is often present among the top 5 responses. Similarly, the Worst-of-5 performance reveals a drop of a similar magnitude, highlighting the variability in model outputs.



\subsubsection*{Main takeaways}
\noindent\fbox{%
    \parbox{\textwidth}{%
        \begin{itemize}[leftmargin=*]
        
        \item \textbf{Impact of test-time scaling and model ranking.} O3-mini consistently performs best, followed by O1 and DeepSeek R1. There is a large performance gap between test-time scaling models and no test-time scaling models. 
        \item \textbf{Token usage, difficulty levels and accuracy}. Test-time scaling models generally take more tokens than no test-time scaling models. More difficult problems requires more tokens. However, more tokens do not necessarily mean higher accuracy even for test-time scaling models probably due to increase in difficulty levels. Additionally, test-time scaling models are using high number of tokens even for relatively easy problems for easy overconstrained setting.

        

        \end{itemize}
    }%
}
\section{TSP - Traveling Salesman Problem}
\label{sec:tsp}
\noindent\textbf{Motivation:}
Along with 3SAT, another algorithmic problem we consider is the Travelling Salesman Problem (refer to the discussion in Appendix \ref{sec:3st} for the motivation behind considering algorithmic problems). TSP is an NP-Hard problem where, given a connectivity graph of cities along with distances between each pair, the goal is to find the shortest possible route that visits each city exactly once and returns to the starting city. This is an optimization problem that tests the model’s ability in combinatorial optimization, as it requires reasoning over many possible tour combinations to identify the one with minimum total cost. 

In contrast to TSP, for which we consider the optimization version, we considered the search version for 3SAT. Therefore, in 3SAT, given a candidate solution, it is easy to verify whether it satisfies the formula. But in the case of TSP, verifying whether a solution is optimal is as hard as finding the optimal one. As we show further, the differences in the difficulty of verifying TSP and 3SAT solutions are also reflected in how robust reasoning models are as problem difficulty increases.

\noindent\textbf{Benchmark description:} 
Each TSP instance we consider is a complete graph, where each city is represented as a node, and each pair of cities is connected by an edge weighted by the distance between them. To facilitate our study, we construct a dataset consisting of 800 TSP instances, spanning eight difficulty levels. Each level varies in terms of the number of nodes in the graph and the distribution of edge weights, with Level 1 containing 6 nodes and Level 8—representing the most challenging cases—containing 13 nodes. For each level, we include 100 unique instances. Ground-truth solutions for all instances are obtained using brute-force search, which exhaustively evaluates all possible permutations of city visits to identify the path with minimal total length. Note that previous work~\citep{fan2023nphardeval} has also considered the TSP problem for evaluation, but it uses approximate solutions instead of exact solutions as ground truth. 

\begin{figure}[t]
    \centering
    \includegraphics[width=0.48\linewidth]    
    {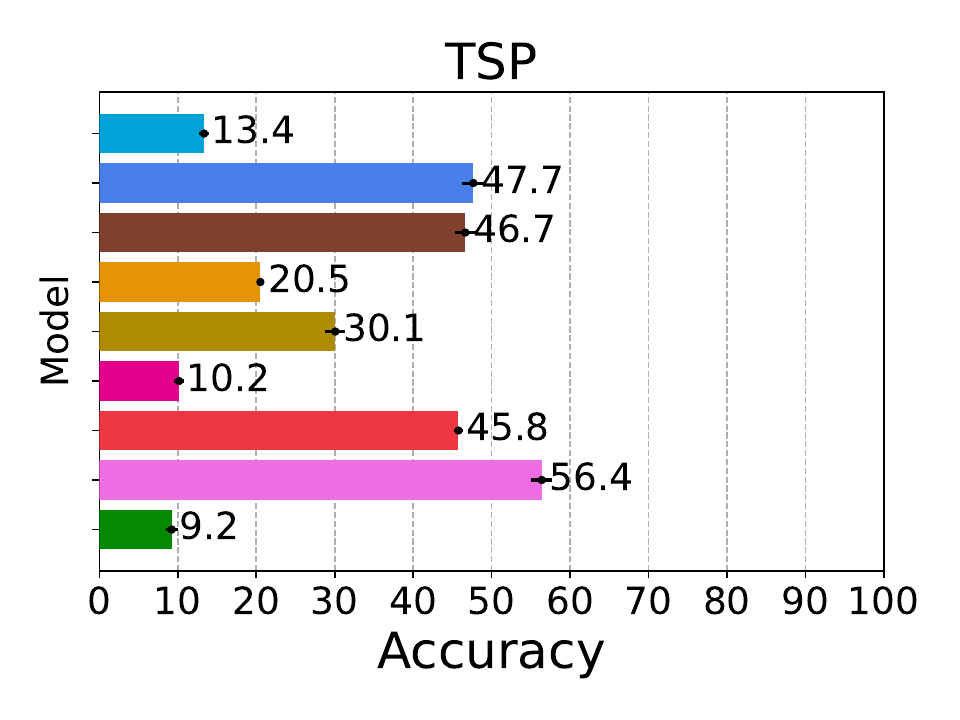}    
    \includegraphics[width=0.42\linewidth]{figures/NPHard_charts/TSP_Analysis_2025-03-18/usage_report/accuracy_vs_tokens_data_point_aggregates.pdf}
    \\
    \includegraphics[width=\textwidth]{figures/model_legend.png}
    \caption{TSP overall performance and token usage. The left figure shows overall model performance across nine models. The right figure shows pareto tradeoff between accuracy and token usage for all benchmarks.}
    \label{fig:tsp_overall_acc_token_usages}
\end{figure}

\noindent\textbf{Model performance:} Figure \ref{fig:tsp_overall_acc_token_usages} presents the mean accuracy achieved by different models. The O3-mini model performs best with an accuracy of 56.4\%, followed by the Claude 3.7 Sonnect, DeepSeek R1 and O1 models, which achieve 47.7\%, 46.7\% and 45.8\% respectively. There is a clear performance gap between models that use test-time scaling and those that do not. For instance, the best-performing non-test-time scaling model, Claude 3.5 Sonnet, achieves only 13.4\% accuracy—significantly lower than its test-time scaling counterparts.

\begin{figure}
    \centering
    \includegraphics[width=1.0\linewidth]{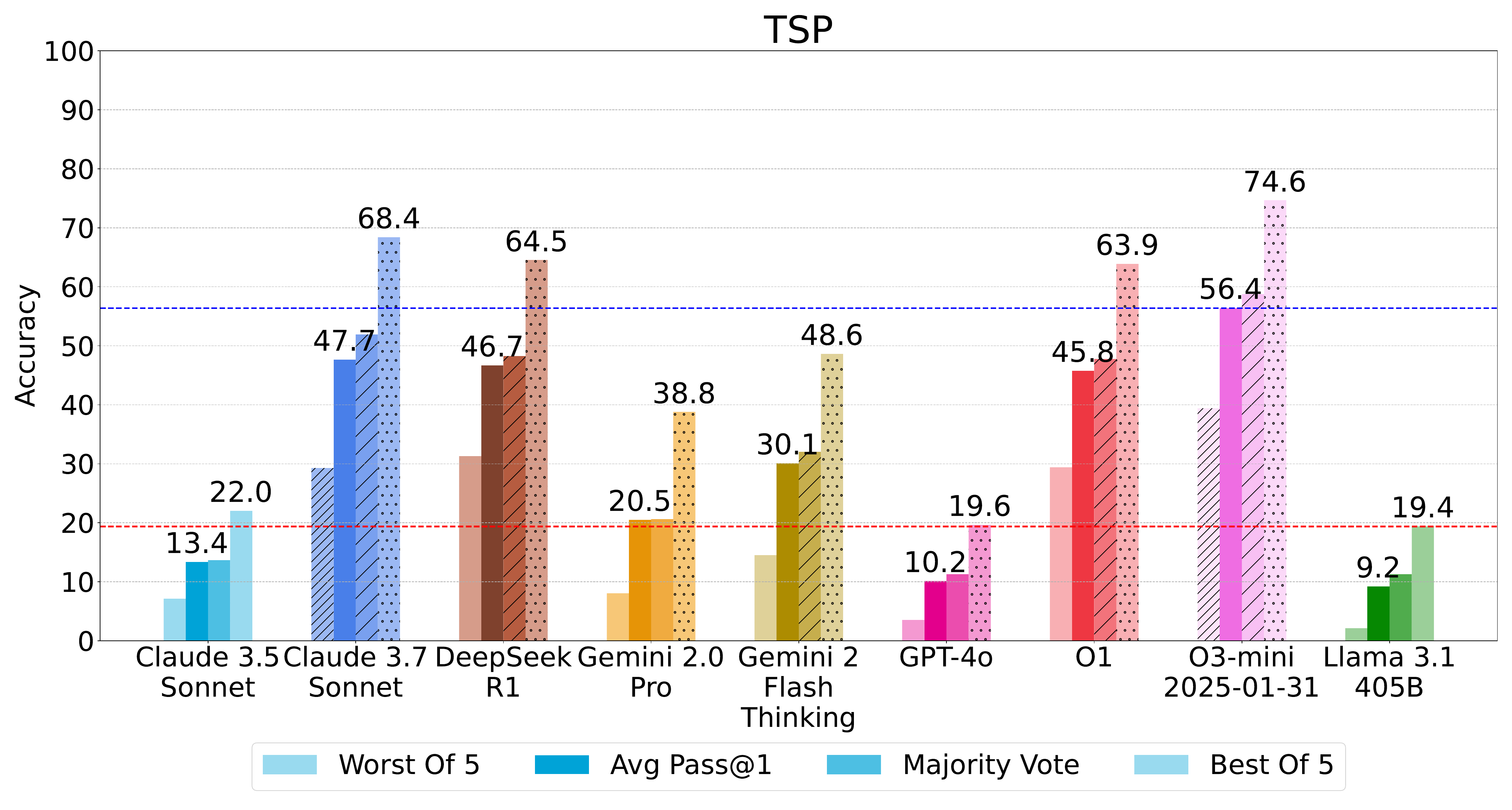}
    \caption{Results on TSP with different aggregations by parallel scaling over 5 runs. The \textcolor{red}{red} line indicates the lowest best-of-5 accuracy observed across all models, while the \textcolor{blue}{blue} line represents the highest average pass@1 accuracy. Note that there is a large gap (almost 35\%) between the best reasoning model and a hypothetical model that can potentially be trained to verify and select the best outcome from the model with the lowest best-of-5 (i.e. GPT-4o).
    }
    \label{fig:tsp_cal_inf_parallel}
\end{figure}

Figure \ref{fig:tsp_cal_inf_parallel} illustrates the Best-of-N, Worst-of-N, and average performance for various models. A key observation is the substantial improvement in accuracy under the Best-of-5 setting, with most models showing gains of 10 to 15 percentage points. This suggests that the correct answer is often present among the top 5 responses. Similarly, the Worst-of-5 performance reveals a drop of a similar magnitude, highlighting the variability in model outputs.

Figure \ref{fig:tsp_token_accuracy_difficulty} shows model accuracy across different difficulty levels. Test-time scaling models outperform non-scaling models, particularly on easier levels. However, even with test-time scaling, performance declines significantly on more challenging instances. After difficulty level 5—corresponding to graphs with 10 nodes—all models begin to struggle, underscoring the increasing complexity of the problem.

\noindent\textbf{Performance vs. token usage tradeoffs:} Figure \ref{fig:tsp_token_accuracy_difficulty} shows average token usage for different models. Test-time scaling models generally take more tokens than no test-time scaling models. More difficult problems requires more tokens. However, more tokens do not necessarily mean higher accuracy even for test-time scaling models probably due to increase in difficulty levels.

\noindent\textbf{Superscaling effects:} Figure \ref{fig:scaleup} shows benefits of superscaling. Scaling up the number of runs (e.g., from 5 to 256) can lead to significant further gains in overall accuracy even for GPT-4o which is not a test time scaling model.
\clearpage




\subsubsection*{Main takeaways}
\noindent\fbox{%
    \parbox{\textwidth}{%
        \begin{itemize}[leftmargin=*]
        \item \textbf{Impact of test-time scaling and model ranking.} O3-mini consistently performs best, followed by Claude 3.7 Sonnet, O1 and DeepSeek R1. There is a large performance gap between test-time scaling models and no test-time scaling models. 
        \item \textbf{Difficulty level vs. accuracy.} Test-time scaling helps to improve accuracy on TSP problems. This is generally observed on easier difficulty levels. However, even with test-time scaling, models still struggle on the most difficult problems.
        \item \textbf{Token usage, difficulty levels and accuracy}. Test-time scaling models generally take more tokens than no test-time scaling models. More difficult problems requires more tokens. However, more tokens do not necessarily mean higher accuracy even for test-time scaling models probably due to increase in difficulty levels.

        \item  \textbf{Impact of verification.} Test time scaling models generally perform much better than no test-time scaling models on SAT problems than TSP problems, which could be attributed to the fact that verification in SAT is easier than in TSP (even for LLMs).

        \item \textbf{Impact of superscaling.} Scaling up the number of runs can lead to significant further gains in overall accuracy even for GPT-4o which is not a test time scaling model. It is also encouraging to see that there exists ample potential even for further improving \OOne.
                
        \end{itemize}
    }%
}
\clearpage
\section{BA-Calendar - Planning}
\label{sec:bacalendar}
\begin{figure}[t]
    \centering
    \begin{subfigure}[b]{0.45\textwidth}
        \centering
        \includegraphics[width=\textwidth]{figures/BA_Calendar/analysis_report/BACalendarMetric_all_correct_overall_accuracy_bar_chart.pdf}
    \end{subfigure}
    \begin{subfigure}[b]{0.45\textwidth}
        \centering
        \includegraphics[width=\textwidth]{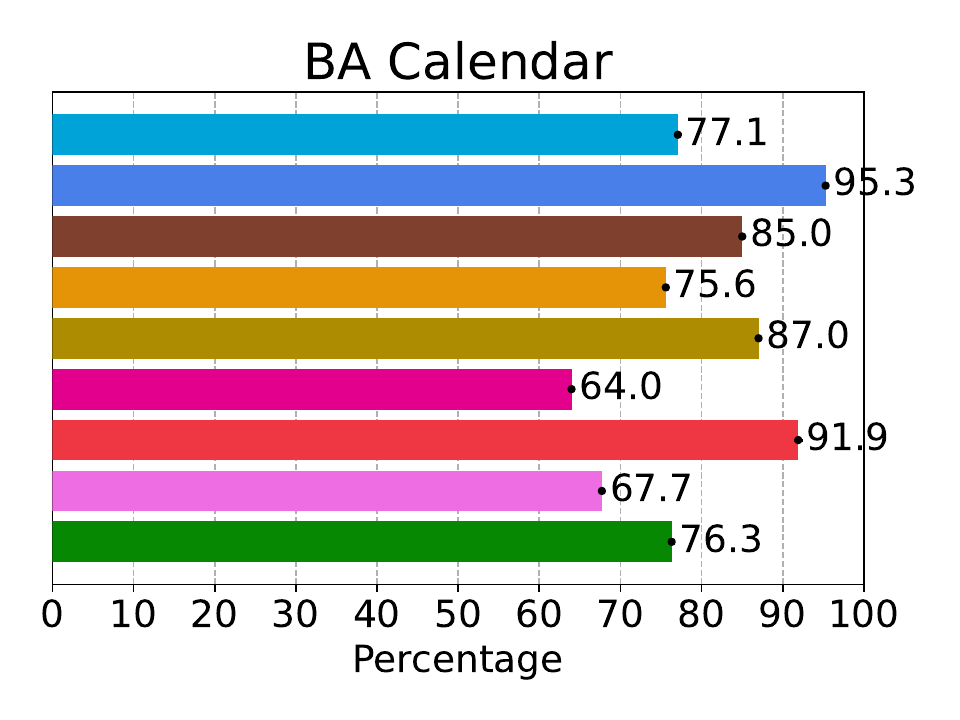}
    \end{subfigure}
        \begin{subfigure}[b]
    {\textwidth}
        \centering
\includegraphics[width=\textwidth]{figures/model_legend.png}
    \end{subfigure}

    \caption{BA-Calendar Metrics.}
    \label{fig:ba_cal_overall}
\end{figure}

\noindent\textbf{Motivation:} While prior evaluations of reasoning models have primarily focused on mathematical and STEM-related benchmarks, it is equally crucial to assess their ability to generalize across different domains. In particular, planning and scheduling require sophisticated reasoning over multiple constraints, making them an important area of evaluation. To address this gap, we evaluate on BA-Calendar, a benchmark designed to test models' proficiency in planning tasks that necessitate handling and satisfying multiple complex constraints. This benchmark is particularly relevant to real-world applications, as effective calendar planning is a fundamental aspect of office productivity and organizational workflows.

\noindent\textbf{Benchmark description:} BA-Calendar is a planning benchmark generated via BenchAgents \cite{butt2024benchagents}, a framework that systematically leverages large language models (LLMs) to automate benchmark creation for complex capabilities while maintaining high-quality data and evaluation metrics. The benchmark comprises a diverse set of calendar planning problems that require models to process and satisfy various constraints, such as participant availability, buffer time between events, task prioritization, and scheduling feasibility. Unlike traditional benchmarks focused on structured logic or single-task reasoning, BA-Calendar evaluates the ability of LLMs to navigate interconnected constraints dynamically, reflecting real-world decision-making challenges in professional and collaborative settings. By assessing performance on BA-Calendar, we gain deeper insights into models' practical utility in workplace environments, particularly in assisting with scheduling, resource management, and coordination tasks.

\noindent\textbf{Model performance:} Overall pass all accuracy in Figure~\ref{fig:ba_cal_overall} shows that the reasoning models like O1, Claude 3.7 Sonnet and Deepseek-R1 perform well on the task with $\ge$80\% accuracy, while non-reasoning models like \GPTFourO, \ClaudeSonnet or \LlamaThreeOneLarge struggle and perform with less that 50\% accuracy. Performance on fraction passed indicates that most models are able to satisfy $\ge$70\% of constraints present in a scheduling problem, but non-reasoning models are unable to reliably satisfy all constraints, often missing a few constraints. This showcases a strength of reasoning models which are able to verify and re-attempt a problem to reach a solution which satisfies all (or more) constraints. 
\begin{figure}[t]
    \centering
    \includegraphics[width=1.0\linewidth]{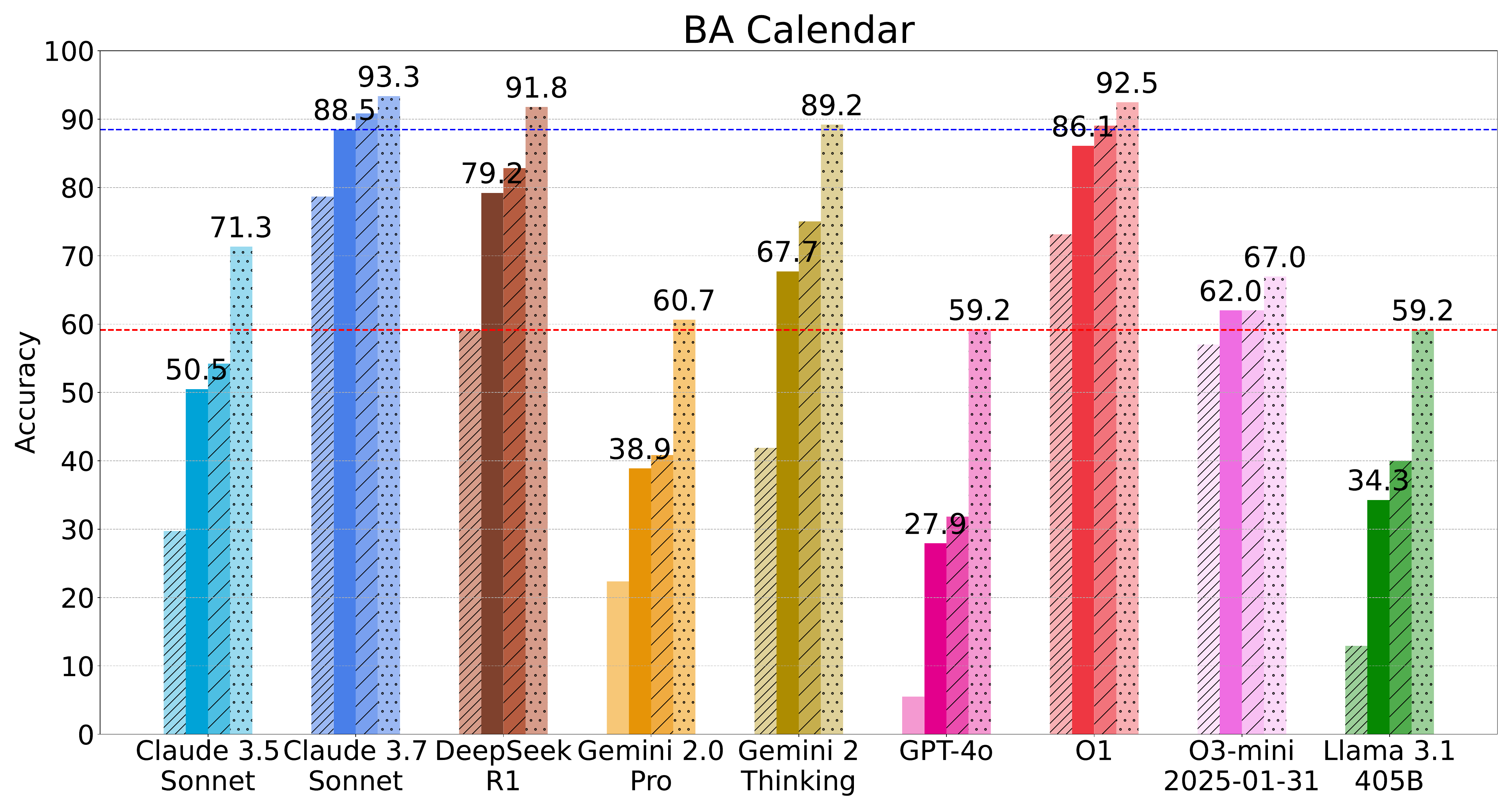}
    \includegraphics[width=0.8\textwidth]{figures/agg_legend.png}
    \caption{Results on BA-Calendar with different aggregations by parallel scaling over 5 runs. The \textcolor{red}{red} line indicates the lowest best-of-5 accuracy observed across all models, while the \textcolor{blue}{blue} line represents the highest average pass@1 accuracy.}
    \label{fig:ba_cal_inf_parallel}
\end{figure}

\begin{figure}[t]
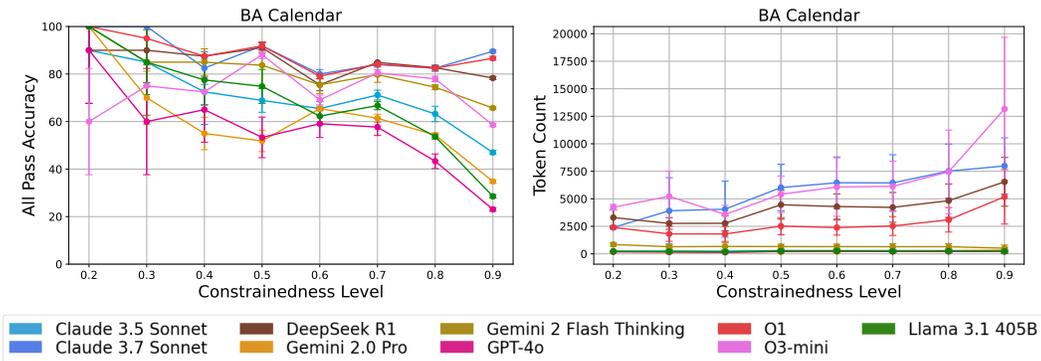

    \centering
    \begin{subfigure}[b]{0.48\textwidth}
        \centering
        \includegraphics[width=\textwidth]{figures/BA_Calendar/difficulty_report/overall_accuracy_by_constraint_level_line_chart.pdf}
    \end{subfigure}
    \begin{subfigure}[b]{0.48\textwidth}
        \centering
        \includegraphics[width=\textwidth]{figures/BA_Calendar/difficulty_report/usage_completion_by_constraint_level_line_chart.pdf}
    \end{subfigure}
        \begin{subfigure}[b]
    {\textwidth}
        \centering
\includegraphics[width=\textwidth]{figures/model_legend.png}
    \end{subfigure}

    \caption{BA-Calendar tokens v/s performance.}
    \label{fig:ba_cal_perf_tokens_constrainedness}
\end{figure}

\begin{figure}[t]
    \centering
    \includegraphics[width=0.85\linewidth]{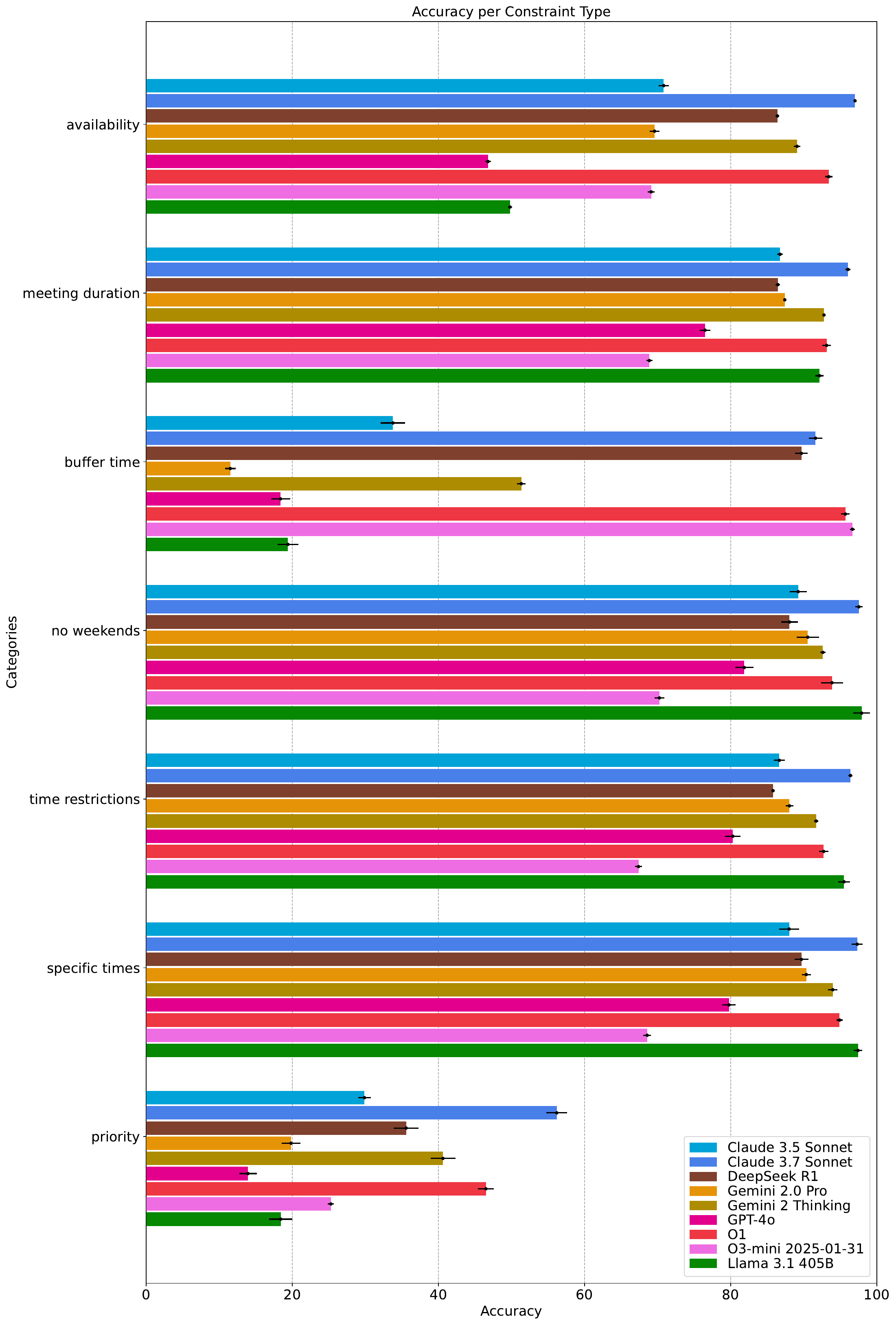}
    \caption{BA-Calendar Constraint Level accuracy.}
    \label{fig:ba_cal_const}
\end{figure}

The dis-aggregations in Figures~\ref{fig:ba_cal_const}  show performance on specific constraints and model-specific strengths and weaknesses. Models like \OOne, \ROne, \OThree show significant  improvement with respect to buffer time and priority, indicating improvements in constraints involving more complex reasoning and arithmetic. Contrary to previous evaluations in math~\citep{O3mini}, \OThree under performs on the task in pass-all accuracy, fraction passed, and in simpler constraints like meeting duration and no weekends often performing on-par or lower than non-reasoning models like \GPTFourO, showing that this distilled model even with high reasoning budget struggles to generalize to other reasoning domains. 

\noindent\textbf{Performance vs. token usage tradeoffs:} 
Figure~\ref{fig:ba_cal_perf_tokens_constrainedness} shows how different models perform in terms of pass-all accuracy under varying constrainedness (complexity). The results show a drop in performance for all models as the complexity increases, showing that as the search space for solutions increases, all models, including reasoning models, struggle to find the correct solution. Correspondingly, we see a significant increase in token usage as constrainedness level increases for reasoning models, again validating that these models can adapt their reasoning pattern to problem difficulty.
\subsubsection*{Main takeaways}
\noindent\fbox{%
    \parbox{\textwidth}{%
        \begin{itemize}[leftmargin=*]
            \item Reasoning models show substantial improvement in satisfying all constraints which was previously extremely hard. There still exists a 15 point gap in extremely hard problems.
    \item Reasoning models show most improvement in buffer time and priority constraints - with priority being a category that models still struggle with.
    \item Smaller gaps between Average Pass@1 and BestofN performance for \OOne and \ROne indicating that models are close to their best performance.
        \end{itemize}
    }%
}
\clearpage
\section{Maze - Navigation}
\label{sec:maze}
\begin{figure}[t]
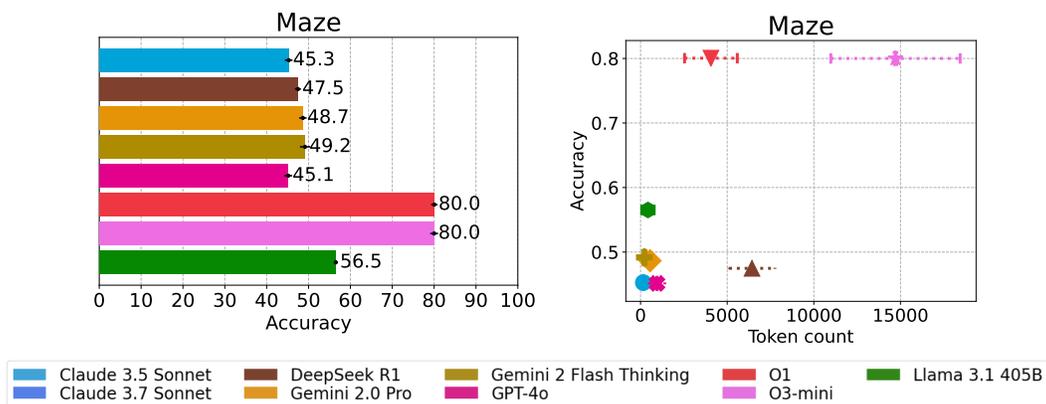

    \centering
    \includegraphics[width=0.46\linewidth]    
{figures/Maze_Charts/SubstringExistsMatch_result_overall_accuracy_bar_chart.pdf}    
    \includegraphics[width=0.40\linewidth]{figures/Maze_Charts/accuracy_vs_tokens_data_point_aggregates.pdf}
    \includegraphics[width=\textwidth]{figures/model_legend.png}
    \caption{Maze overall performance and token usage.}
    \label{fig:maze_overall_acc_token_usages}
\end{figure}

\begin{figure}
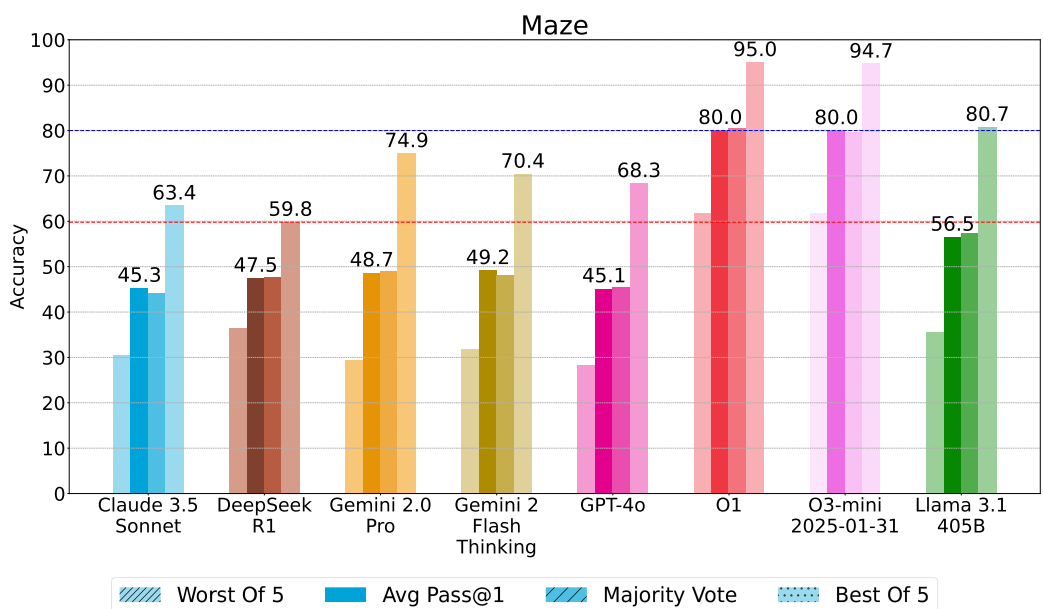

    \centering
    \includegraphics[width=1.0\linewidth]{figures/Maze_Charts/SubstringExistsMatch_result_inference_comp_accuracy_bar_chart.pdf}
    \includegraphics[width=0.8\textwidth]{figures/agg_legend.png}
    \caption{Results on Maze with different aggregations by parallel scaling over 5 runs. The \textcolor{red}{red} line indicates the lowest best-of-5 accuracy observed across all models, while the \textcolor{blue}{blue} line represents the highest average pass@1 accuracy.}
    \label{fig:maze_inf_parallel}
\end{figure}
\noindent\textbf{Motivation:} 
A key component of reasoning for LLMs is spatial reasoning. We use the benchmark of \cite{wang2024pictureworththousandwords} to measure these abilities.  This dataset is a procedurally generated synthetic dataset designed for both multimodal and text-only model capabilities.  In this work, we focus on text-only reasoning skills.

\noindent\textbf{Benchmark description:} 
The dataset consists of small mazes presented in the forms both image and ASCII code. A maze consists of colored blocks where different colors signify distinct elements: \emph{``a green block marks
the starting point (S), a red block indicates the exit (E), black blocks represent impassable walls,
white blocks denote navigable paths, and blue blocks trace the path from S to E. The objective is to
navigate from S to E following the blue path, with movement permitted in the four cardinal directions
(up, down, left, right).''}.
The task of the LLM is to answer questions, i.e., counting the number of turns from S to E and determining the spatial relationship 
between S and E. 

\begin{figure}[t!]
\centering
    \includegraphics[width=.8\textwidth]{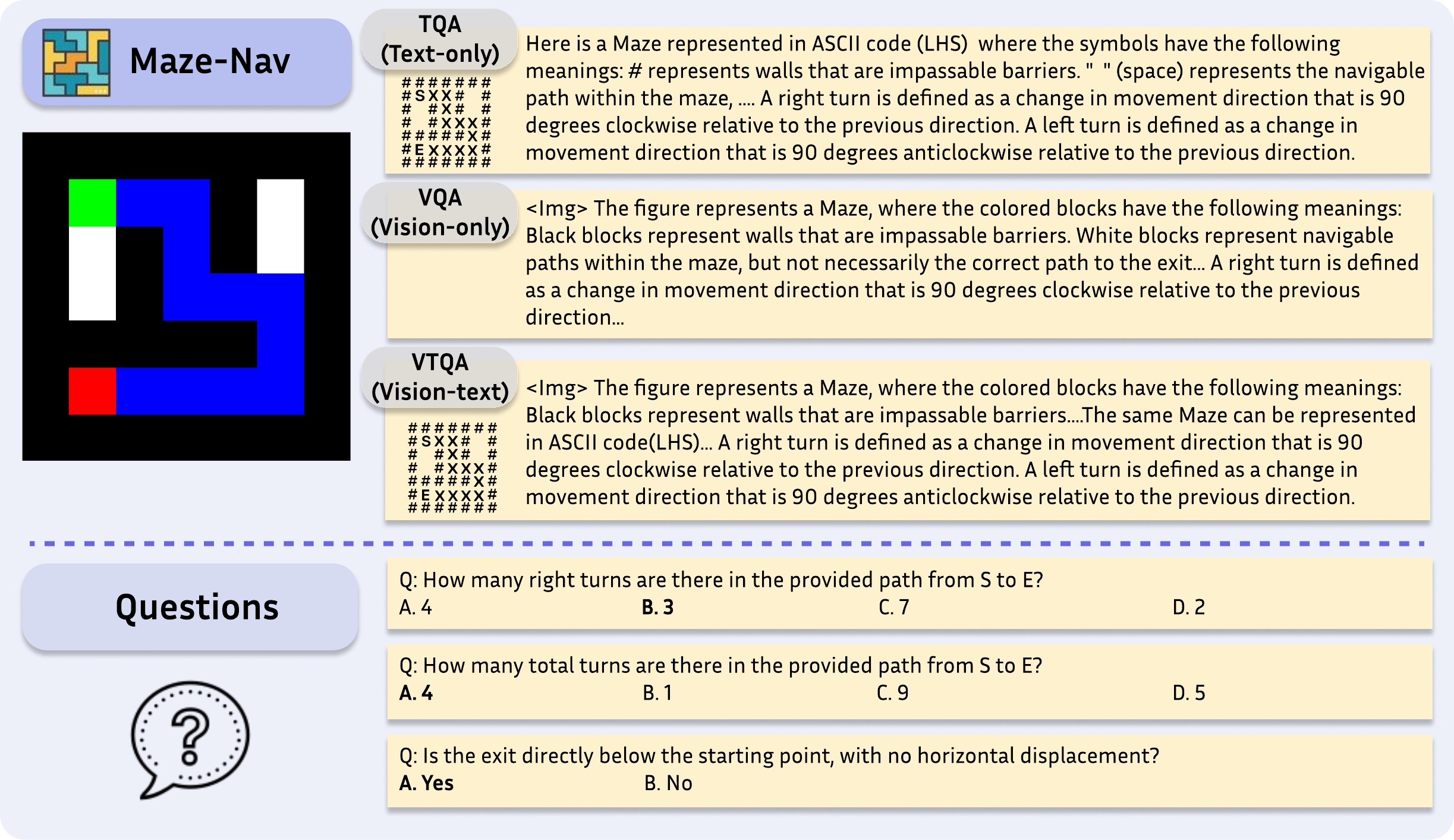}\\
\caption{Illustration of the Maze benchmark. Originally, the benchmark includes three different modalities: Text-only, Vision-only, and Vision-text. In this work, we only focus in the text-only reasoning skills.}
\label{fig:vl_maze}
\end{figure}

Each task has three conditions, with respect to the input modality, 1) text-only, input and a question, 2) vision-only, and 3) vision-text includes both text and image representations with the question. See Figure~\ref{fig:vl_maze} for an illustration of each task. We used only the text-only condition, which 1500 questions.

\noindent\textbf{Model performance:} 
Figure~\ref{fig:maze_overall_acc_token_usages} shows the mean accuracy of each model.  \OOne and \OThree have the best performance at 80\% accuracy, and then there is a very large gap between these two models and all others, which are in the 40-50\% range.  It is interesting and perhaps surprising that other test-time models such as \ROne, \ClaudeSonnetThinking, and \GeminiFlash actually perform worse than \LlamaThreeOneLarge, which is a conventional model.  This is a bit inconsistent with many of the other benchmarks in this paper.
One explanation for this could be that, as shown in Figure~\ref{fig:maze_inf_parallel} is that the conventional-to-reasoning gap for Maze is quite large, about 20\%, thus we expect test-time models to have a good opportunity for increased performance -- for this dataset, it appears \OOne and \OThree take better advantage of this opportunity. 

\noindent\textbf{Performance vs. token usage tradeoffs:} 
Figure~\ref{fig:maze_overall_acc_token_usages} shows average token usage for different models. The test-time scaling models generally take more tokens than no test-time scaling
models, with \OThree having the highest average token use and also a lot of variability in the number of tokens used.

\noindent\textbf{Scaling effects:} 
Figure~\ref{fig:maze_inf_parallel} illustrates the Best-of-N, Worst-of-N, and average performance for each model. A key observation is the substantial improvement in accuracy under the Best-of-5 setting, with most models showing gains of 15 to 25 percentage points. This
suggests that the correct answer is often present among the top 5 responses. Similarly, the Worst-of-5 performance reveals a drop of a similar magnitude, highlighting the variability in model output.

\subsubsection*{Main takeaways}
\noindent\fbox{%
    \parbox{\textwidth}{%
        \begin{itemize}[leftmargin=*]
        \item \OOne and \OThree have the best performance at 80\% accuracy and then there is a very big gap between these two models and all others, which are in the 40-50\% range.
        \item All models show a large improvement with benefit from Best-of-5, including reasoning models, which shows that there is still remaining opportunity for further improvement.
        \item Test-time scaling models generally take more tokens than no test-time scaling models. However, more tokens do not necessarily mean higher accuracy, for example with \ROne.
        \end{itemize}
    }%
}
\section{SpatialMap - Spatial Reasoning}
\label{sec:spatialmap}
\begin{figure}[t]
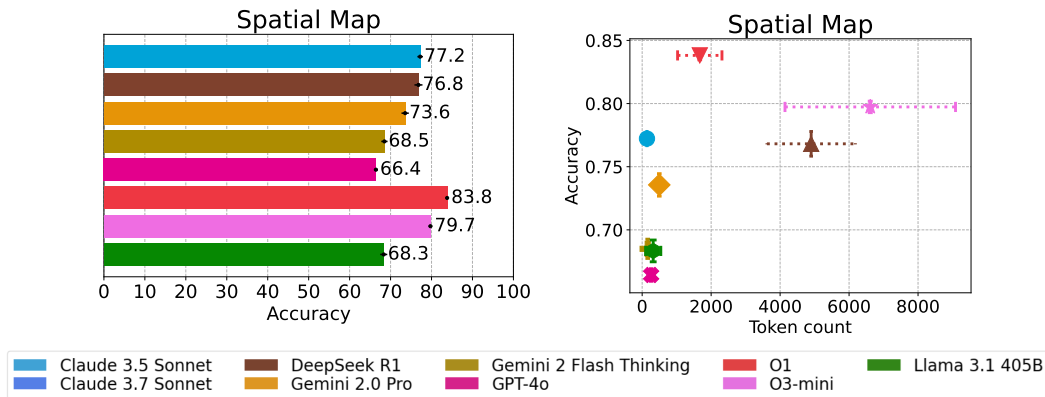

    \centering
    \includegraphics[width=0.45\linewidth]    
{figures/Spatial_Map_Charts/SubstringExistsMatch_result_overall_accuracy_bar_chart.pdf}    
    \includegraphics[width=0.40\linewidth]{figures/Spatial_Map_Charts/accuracy_vs_tokens_data_point_aggregates.pdf}
    \includegraphics[width=\textwidth]{figures/model_legend.png}
    \caption{SpatialMap overall performance and token usage.}
    \label{fig:spatial_map_overall_acc_token_usages}
\end{figure}
\begin{figure}[t]
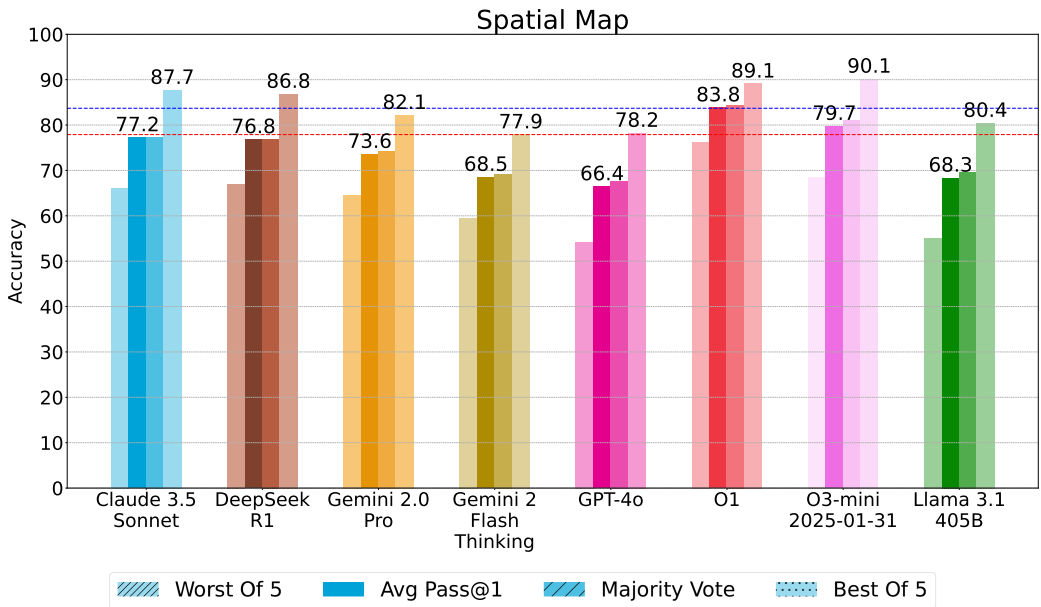

    \centering
    \includegraphics[width=1.0\linewidth]{figures/Spatial_Map_Charts/SubstringExistsMatch_result_inference_comp_accuracy_bar_chart.pdf}
    \includegraphics[width=0.8\textwidth]{figures/agg_legend.png}
    \caption{Results on SpatialMap with different aggregations by parallel scaling over 5 runs. The \textcolor{red}{red} line indicates the lowest best-of-5 accuracy observed across all models, while the \textcolor{blue}{blue} line represents the highest average pass@1 accuracy.}
    \label{fig:spatial_map_inf_parallel}
\end{figure}
\noindent\textbf{Motivation:} 
A key question for understanding reasoning capabilities of a model is what is the ability for spatial reasoning and understanding.  We use the benchmark from~\cite{wang2024pictureworththousandwords} to measure these abilities.  This dataset is a procedurally generated synthetic dataset designed to test multimodal vs. language capabilities of models.  In this work, we only focus in the text-only reasoning skills.

\noindent\textbf{Benchmark description:} 
The dataset consists of spatial relationships for random layouts of symbolic objects with text names on  white background. 
Each object is associated with a unique location name, such as Unicorn Umbrellas and Gale Gifts. To study the impact of modality,
the textual representation of each input consists of pairwise relations such as ``Brews Brothers Pub
is to the Southeast of Whale’s Watches''. The questions include asking about the spatial
relationships between two locations and the number of objects that meet specific spatial criteria.

\begin{figure}[t!]
\centering
    \includegraphics[width=.8\textwidth]{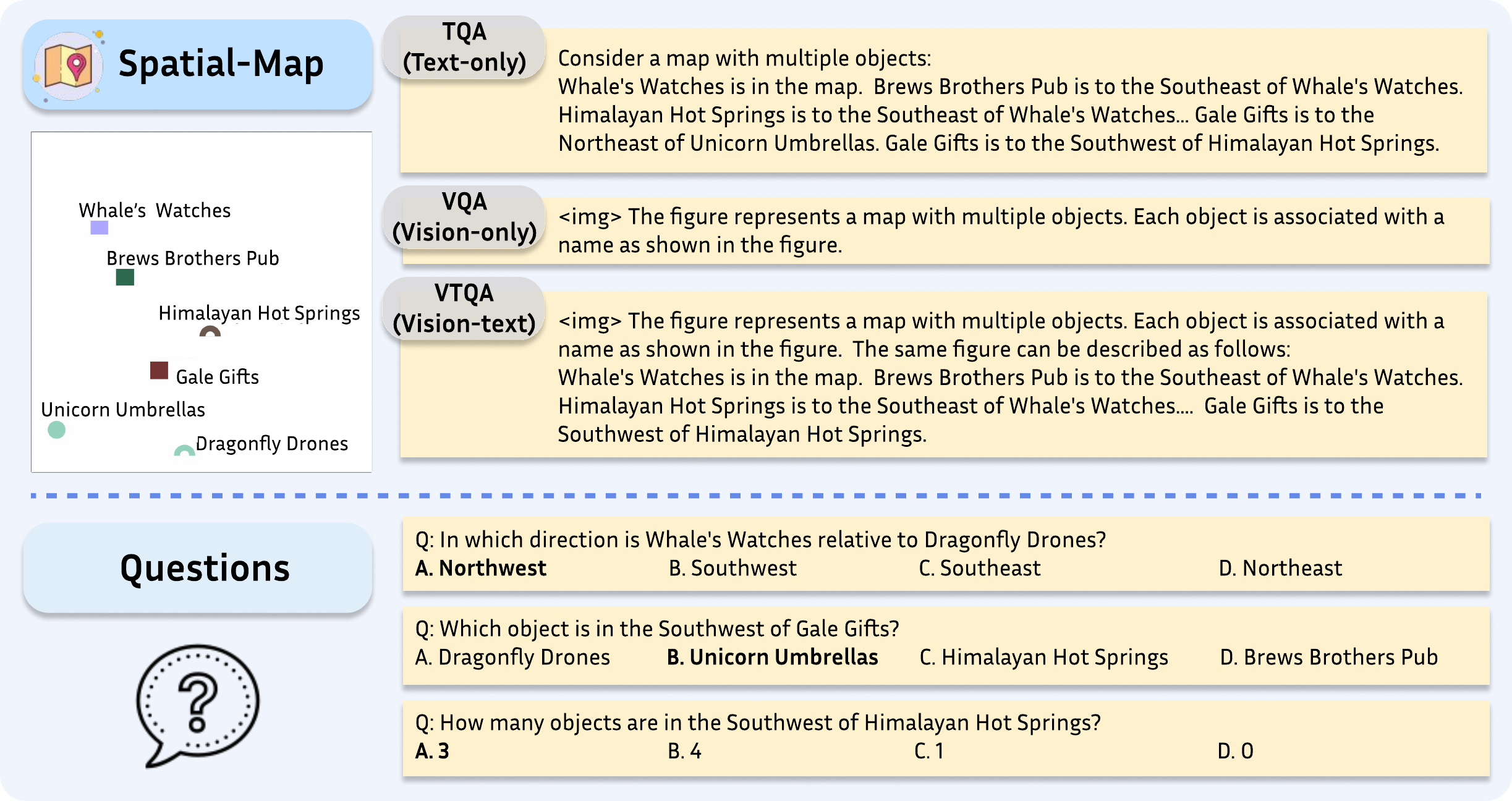}\\
\caption{Illustration of the Spatial-Map (spatial understanding) benchmark. Originally, the benchmark includes three different modalities: Text-only, Vision-only, and Vision-text. In this work, we only focus in the text-only reasoning skills.}
\label{fig:vl_spatialmap}
\end{figure}

Each task has three conditions, with respect to the input modality, 1) text-only, input and a question, 2) vision-only, and 3) vision-text includes both text and image representations with the question. See Figure~\ref{fig:vl_spatialmap} for an illustration of each task. We used only the text-only condition, which 1500 questions.

\noindent\textbf{Model performance:} 
Figure~\ref{fig:spatial_map_overall_acc_token_usages} shows the mean accuracy for each model.  \OOne has the best performance at 83.8\% accuracy with \OThree about 4\% behind.  The next best-performing models are test-time models: \ROne and \ClaudeSonnetThinking, while
\GeminiFlash performs on par with the conventional models.  Overall, the accuracy across models is not a very large spread.

\noindent\textbf{Performance vs. token usage tradeoffs:} 
Figure~\ref{fig:spatial_map_overall_acc_token_usages} shows average token usage for different models. The test-time scaling models generally take more tokens than the non-test-time scaling models, \OThree having the highest average token use and also a lot of variability in the number of tokens used.

\noindent\textbf{Scaling effects:} 
Figure~\ref{fig:spatial_map_inf_parallel} illustrates the Best-of-N, Worst-of-N, and average performance for each model. Here, the conventional-to-reasoning gap is not very large, about 6\%, thus there is less opportunity for test-time models to have an advantage. One explanation for this could be that this benchmark is nearing saturation, where the average reasoning difficulty of the questions is not high enough for test-time models to gain much of an advantage.  The Worst-of-5 performance reveals a slightly larger drop than the gain for the Best-of-5 performance.

\subsubsection*{Main takeaways}
\noindent\fbox{%
    \parbox{\textwidth}{%
        \begin{itemize}[leftmargin=*]
        \item \OOne has the best performance at 83.8\% accuracy with \OThree about 4\% behind.  The next best-performing models are test-time models: \ROne and \ClaudeSonnetThinking, while 
\GeminiFlash performs on par with the conventional models.
        \item The conventional-to-reasoning gap is not very large, about 6\%, thus there is less opportunity for test-time models to have an advantage.
        \item Test-time scaling models generally take more tokens than no test-time scaling models. However, more tokens do not necessarily mean higher accuracy, for example with \ClaudeSonnet out-performs \ROne with far fewer tokens.
        \end{itemize}
    }%
}

\section{Performance vs. token usage tradeoffs - Extended}
\label{sec:extended_length_variance}
\begin{figure}[h!]
    \centering
    \begin{subfigure}[b]{0.24\textwidth}
    \centering
    \includegraphics[width=\textwidth]{figures/GPQA_charts/accuracy_vs_tokens/claude-3-7-sonnet-20250219_distribution_stdev_correct_incorrect_mixed.pdf}
    \end{subfigure}
    \begin{subfigure}[b]{0.24\textwidth}
        \centering
        \includegraphics[width=\textwidth]{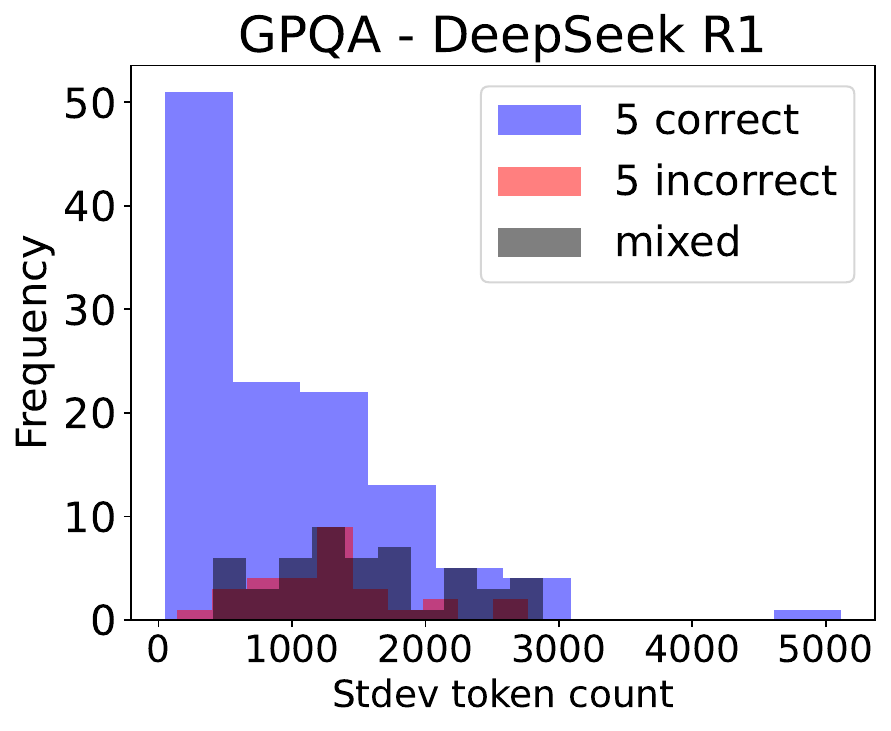}
    \end{subfigure}
        \begin{subfigure}[b]{0.24\textwidth}
        \centering
        \includegraphics[width=\textwidth]{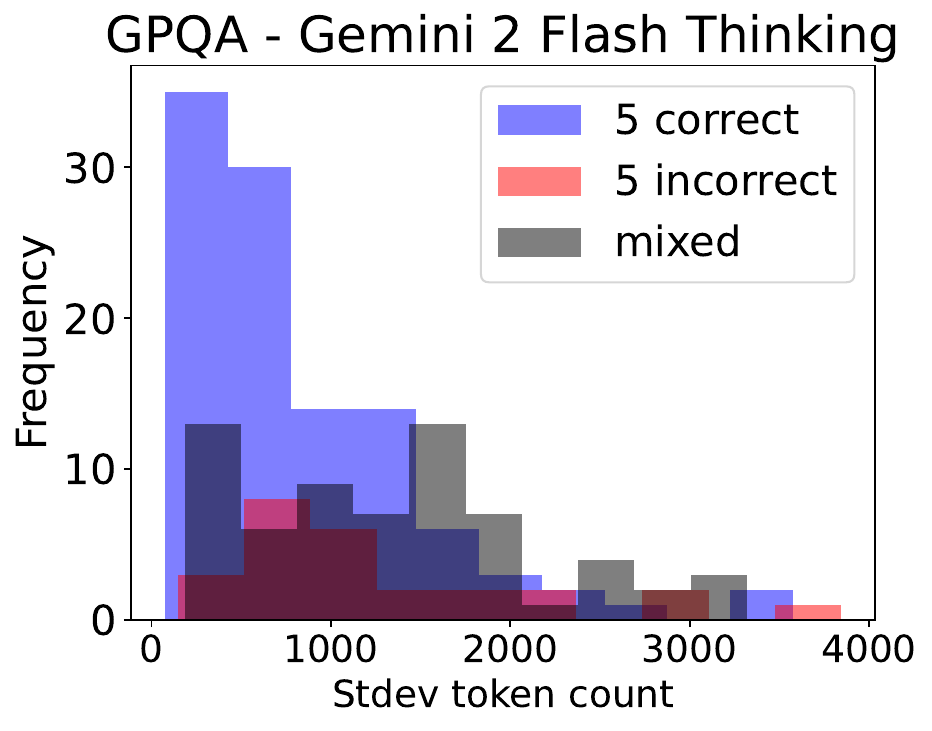}
    \end{subfigure}
    \begin{subfigure}[b]{0.24\textwidth}
        \centering
        \includegraphics[width=\textwidth]{figures/GPQA_charts/accuracy_vs_tokens/o1-20241217_distribution_stdev_correct_incorrect_mixed.pdf}
    \end{subfigure}

    \begin{subfigure}[b]{0.24\textwidth}
    \centering
    \includegraphics[width=\textwidth]{figures/Omni_Math/usage_variance/claude-3-7-sonnet-20250219_distribution_stdev_correct_incorrect_mixed.pdf}
    \end{subfigure}
    \begin{subfigure}[b]{0.24\textwidth}
        \centering
        \includegraphics[width=\textwidth]{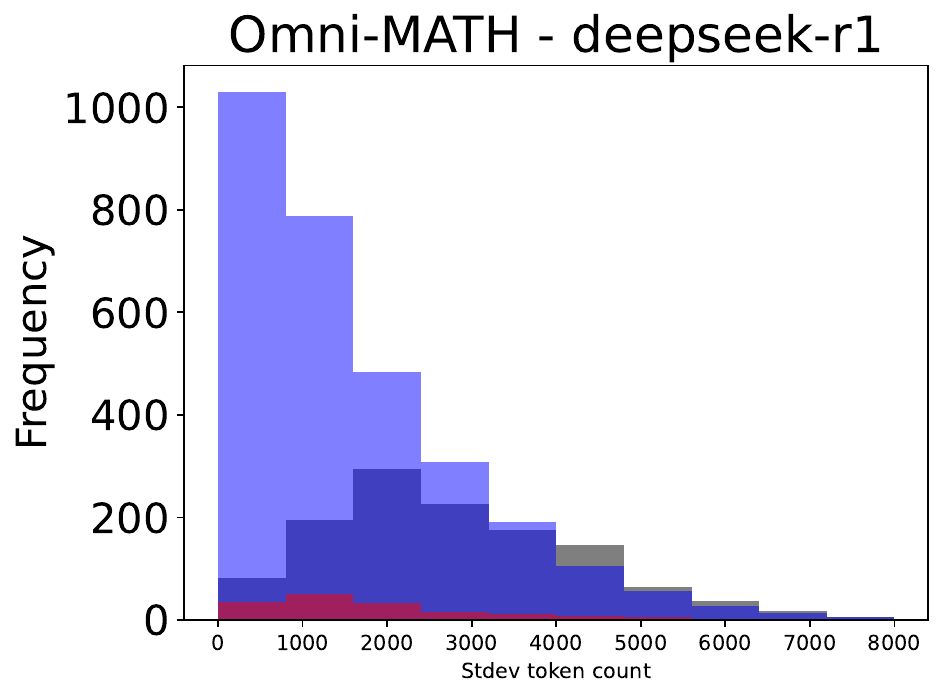}
    \end{subfigure}
        \begin{subfigure}[b]{0.24\textwidth}
        \centering
        \includegraphics[width=\textwidth]{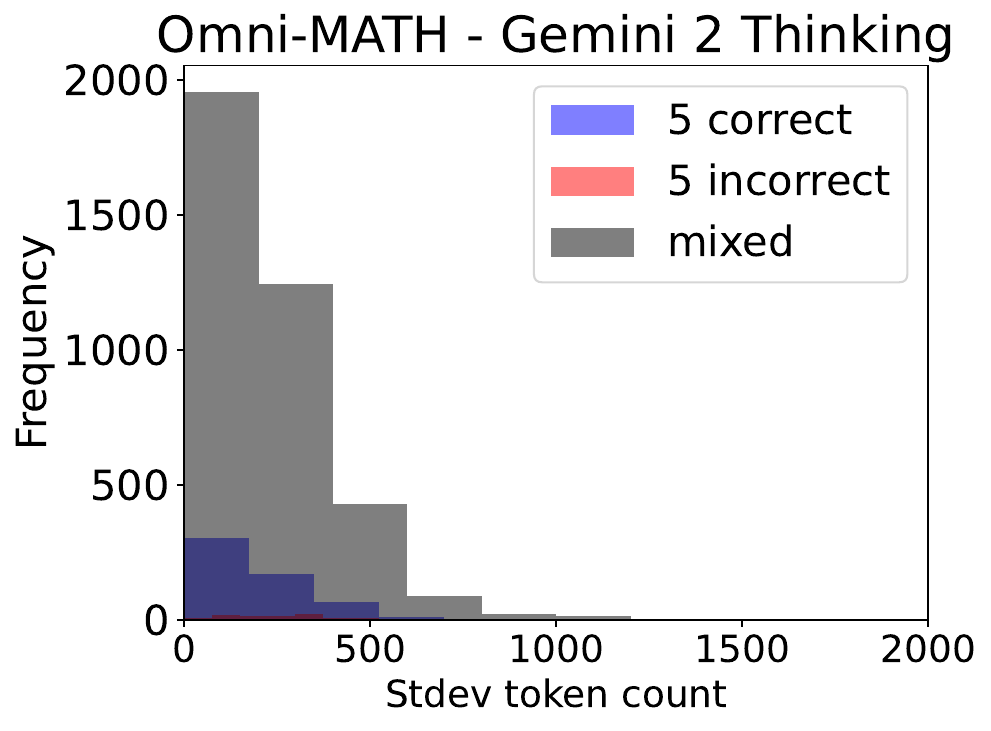}
    \end{subfigure}
    \begin{subfigure}[b]{0.24\textwidth}
        \centering
        \includegraphics[width=\textwidth]{figures/Omni_Math/usage_variance/o1-20241217_distribution_stdev_correct_incorrect_mixed.pdf}
    \end{subfigure}

   \begin{subfigure}[b]{0.24\textwidth}
   \centering
   \includegraphics[width=\textwidth]{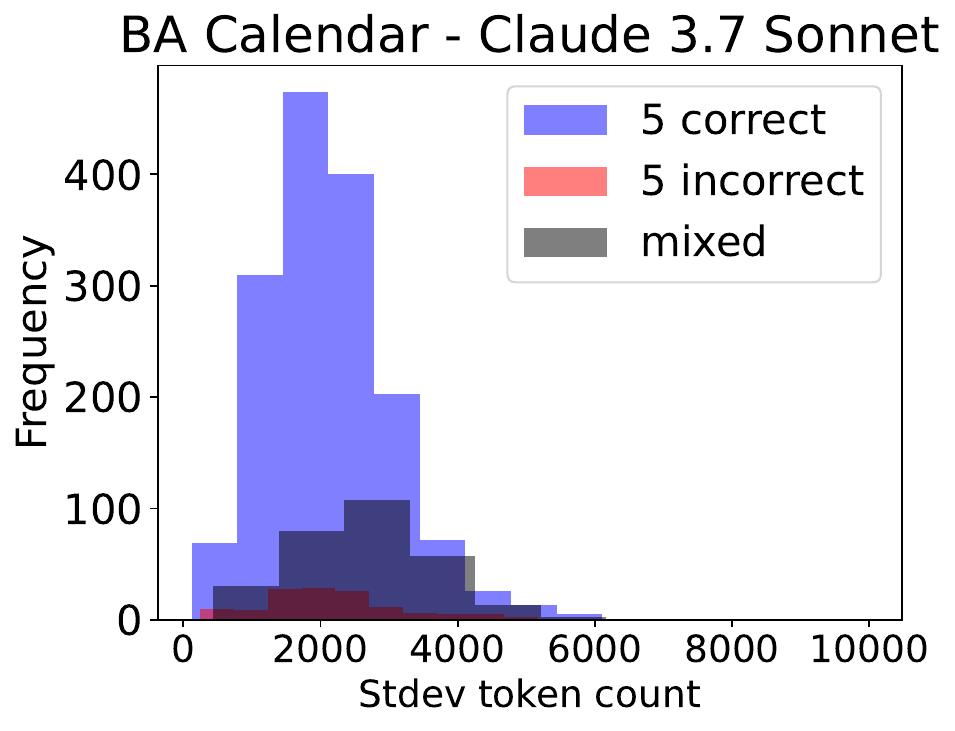}
   \end{subfigure}
   \begin{subfigure}[b]{0.24\textwidth}
       \centering
       \includegraphics[width=\textwidth]{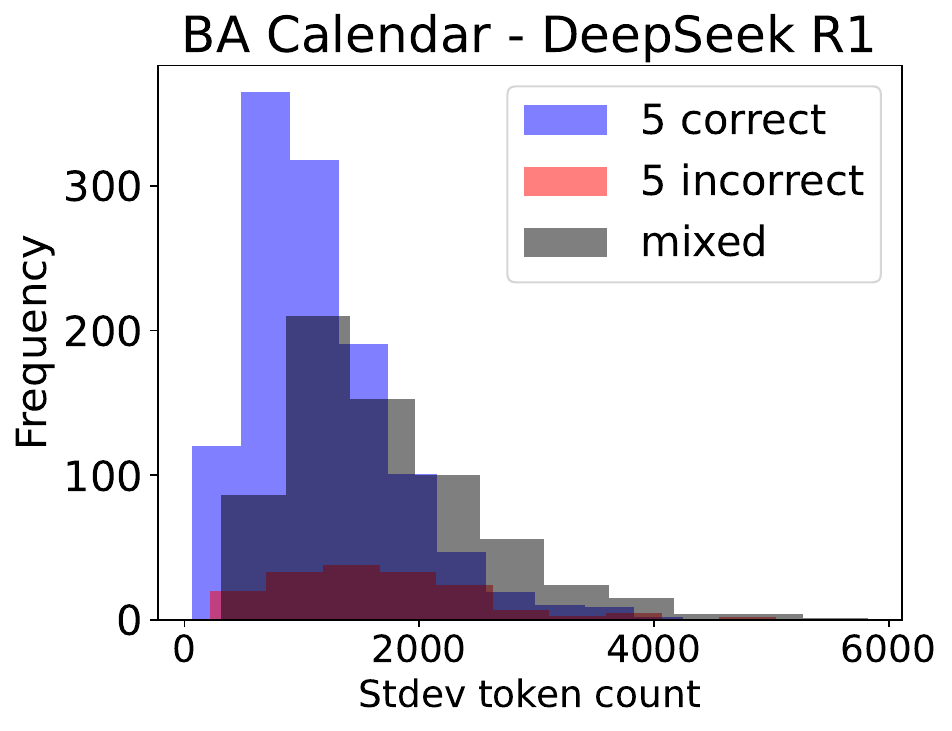}
   \end{subfigure}
       \begin{subfigure}[b]{0.24\textwidth}
       \centering
       \includegraphics[width=\textwidth]{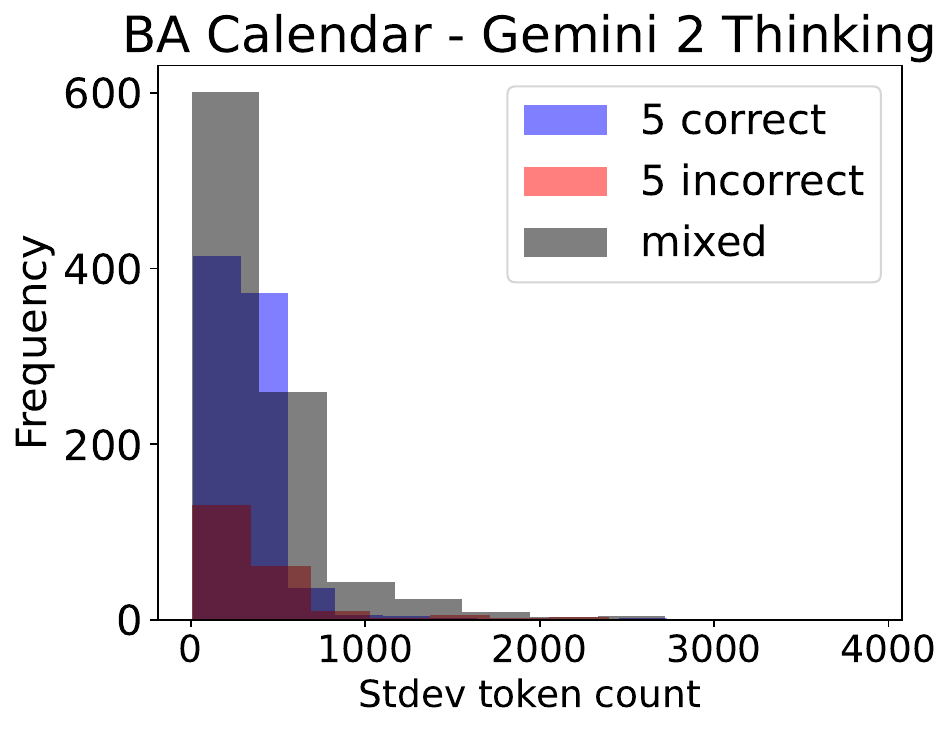}
   \end{subfigure}
   \begin{subfigure}[b]{0.24\textwidth}
       \centering
       \includegraphics[width=\textwidth]{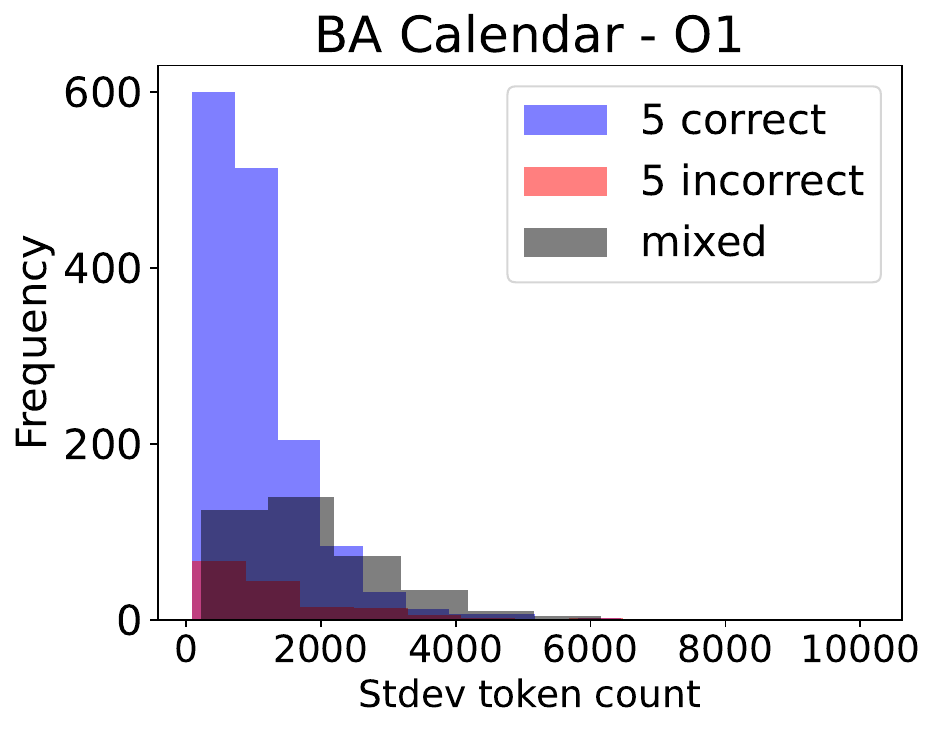}
   \end{subfigure}
    
    \caption{Distributions of the standard deviations of token usage within the same instance (5 repeats), shown for instances where the models are always correct, always incorrect, or mixed.}
\label{fig:correct_incorrect_token_usage_stdev_appendix}
\end{figure}
\clearpage
\begin{figure}[t]
    \centering
    \begin{subfigure}[b]{0.24\textwidth}
        \centering
        \includegraphics[width=\textwidth]{figures/GPQA_charts/accuracy_vs_tokens/claude-3-7-sonnet-20250219_distribution_mean_correct_incorrect_mixed.pdf}
    \end{subfigure}
    \begin{subfigure}[b]{0.24\textwidth}
        \centering
        \includegraphics[width=\textwidth]{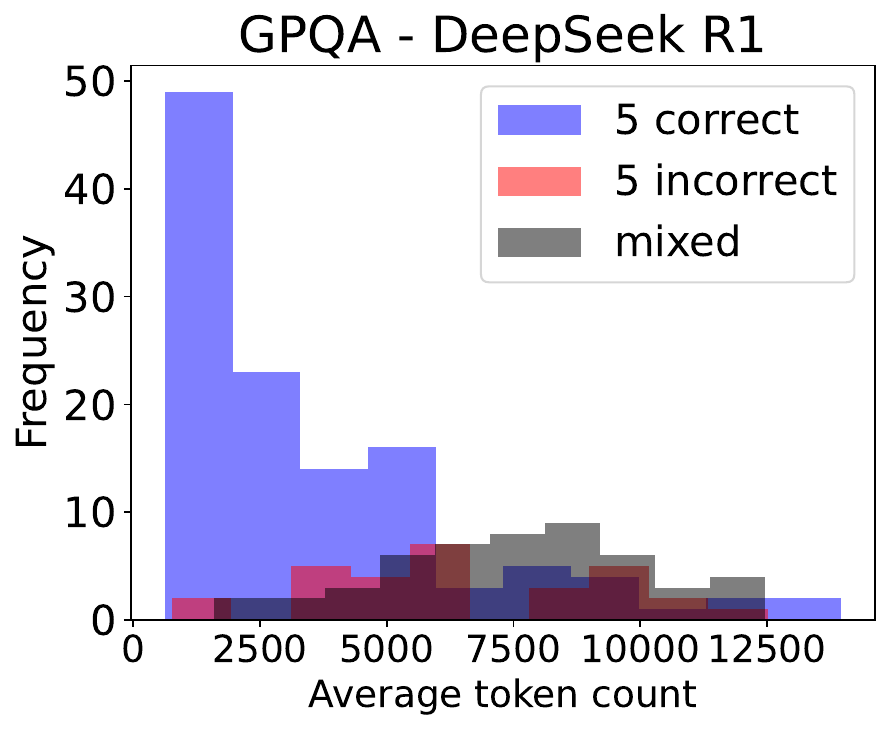}
    \end{subfigure}
        \begin{subfigure}[b]{0.24\textwidth}
        \centering
        \includegraphics[width=\textwidth]{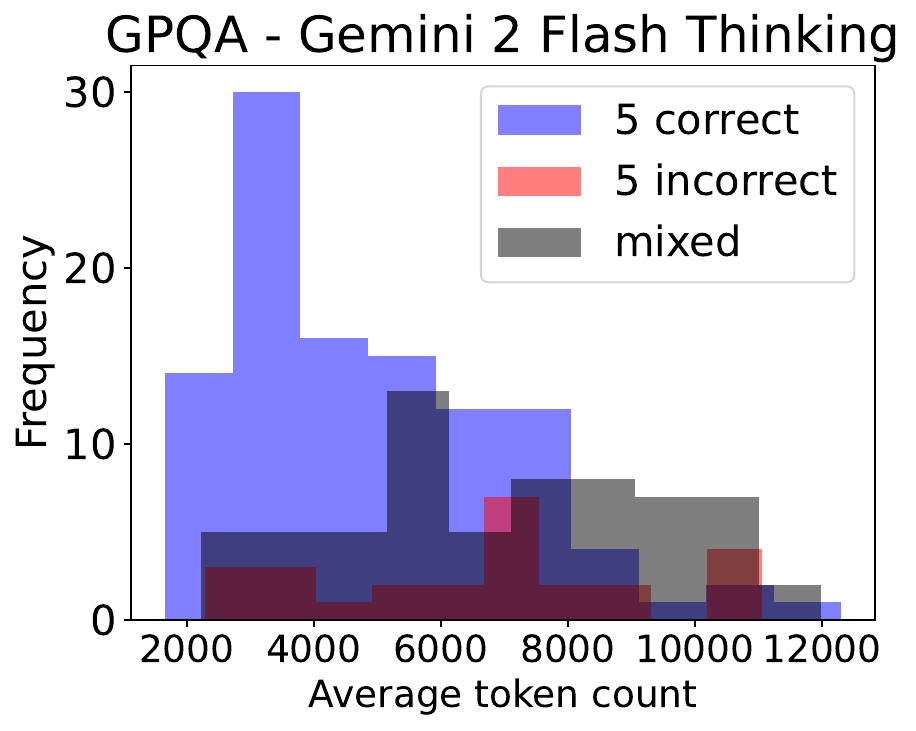}
    \end{subfigure}
    \begin{subfigure}[b]{0.24\textwidth}
        \centering
        \includegraphics[width=\textwidth]{figures/GPQA_charts/accuracy_vs_tokens/o1-20241217_distribution_mean_correct_incorrect_mixed.pdf}
    \end{subfigure}

   \begin{subfigure}[b]{0.24\textwidth}
       \centering
       \includegraphics[width=\textwidth]{figures/Omni_Math/usage_variance/claude-3-7-sonnet-20250219_distribution_mean_correct_incorrect_mixed.pdf}
   \end{subfigure}
    \begin{subfigure}[b]{0.24\textwidth}
        \centering
        \includegraphics[width=\textwidth]{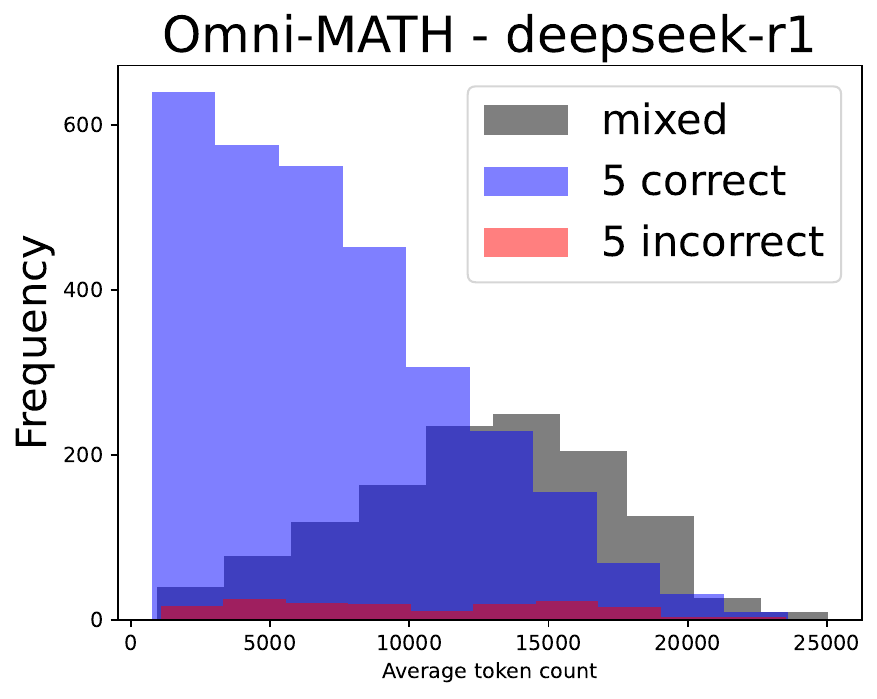}
    \end{subfigure}
        \begin{subfigure}[b]{0.24\textwidth}
        \centering
        \includegraphics[width=\textwidth]{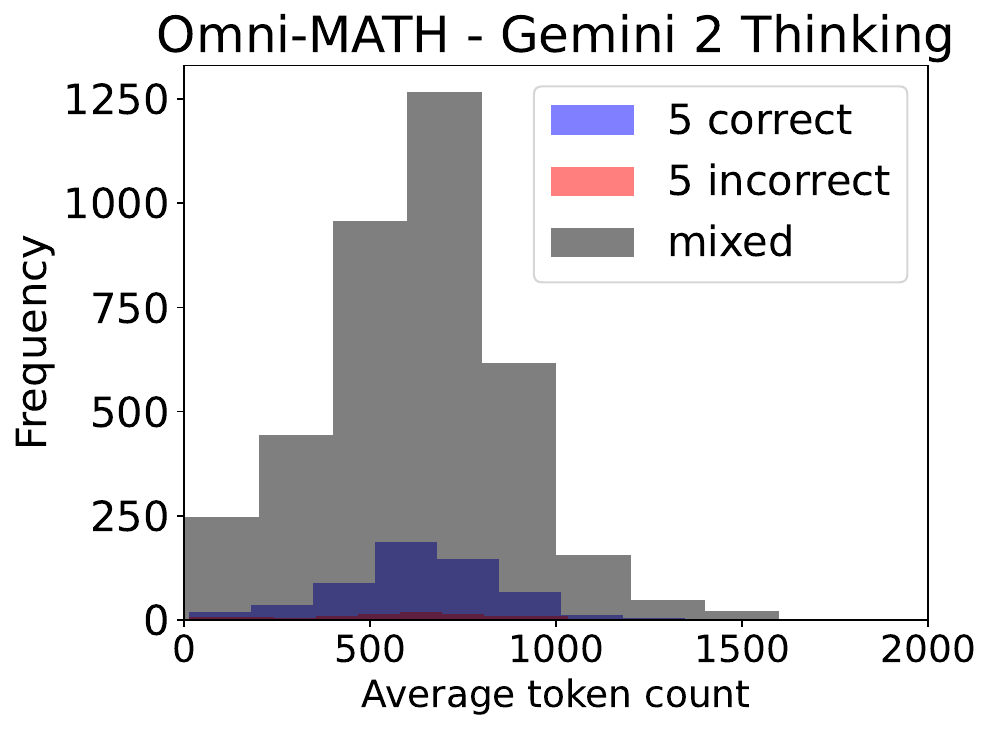}
    \end{subfigure}
    \begin{subfigure}[b]{0.24\textwidth}
        \centering
        \includegraphics[width=\textwidth]{figures/Omni_Math/usage_variance/o1-20241217_distribution_mean_correct_incorrect_mixed.pdf}
    \end{subfigure}

   \begin{subfigure}[b]{0.25\textwidth}
       \centering
       \includegraphics[width=\textwidth]{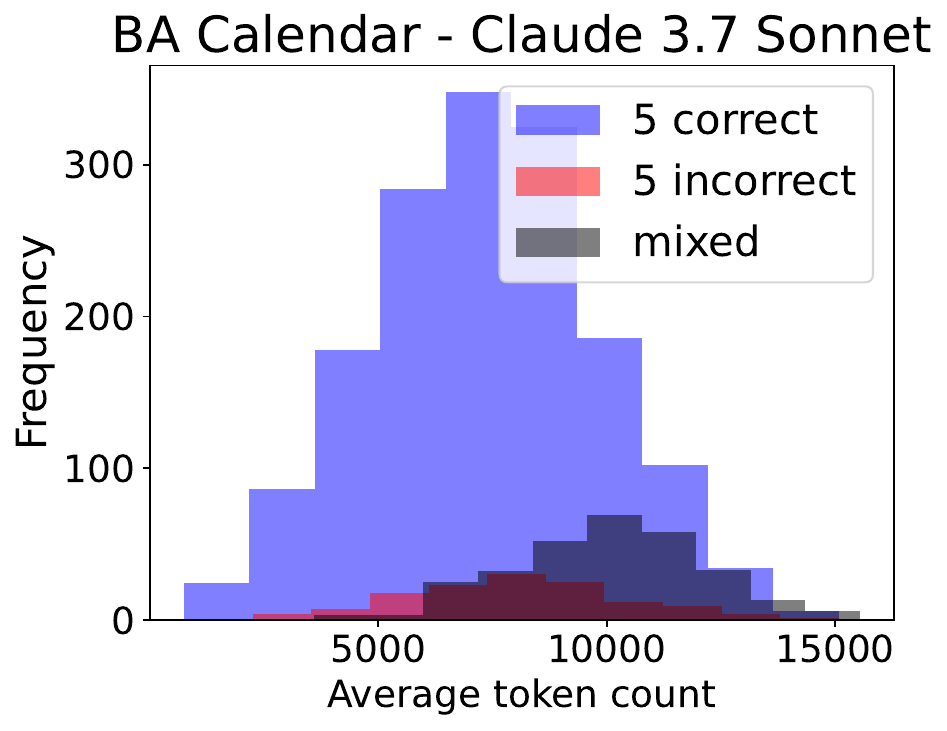}
   \end{subfigure}
   \begin{subfigure}[b]{0.24\textwidth}
       \centering
       \includegraphics[width=\textwidth]{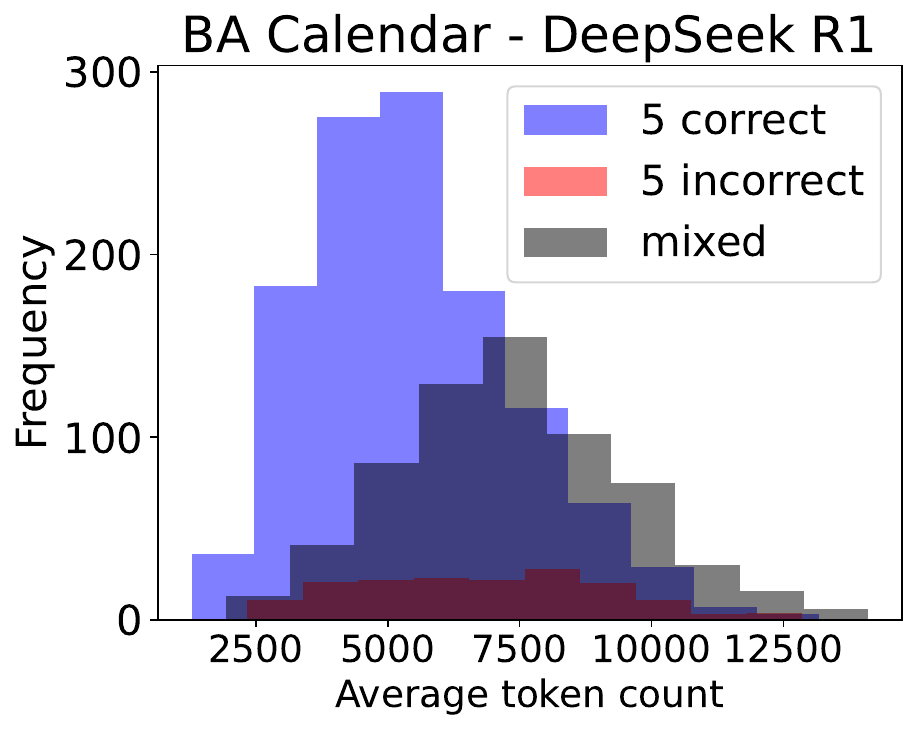}
   \end{subfigure}
       \begin{subfigure}[b]{0.24\textwidth}
       \centering
       \includegraphics[width=\textwidth]{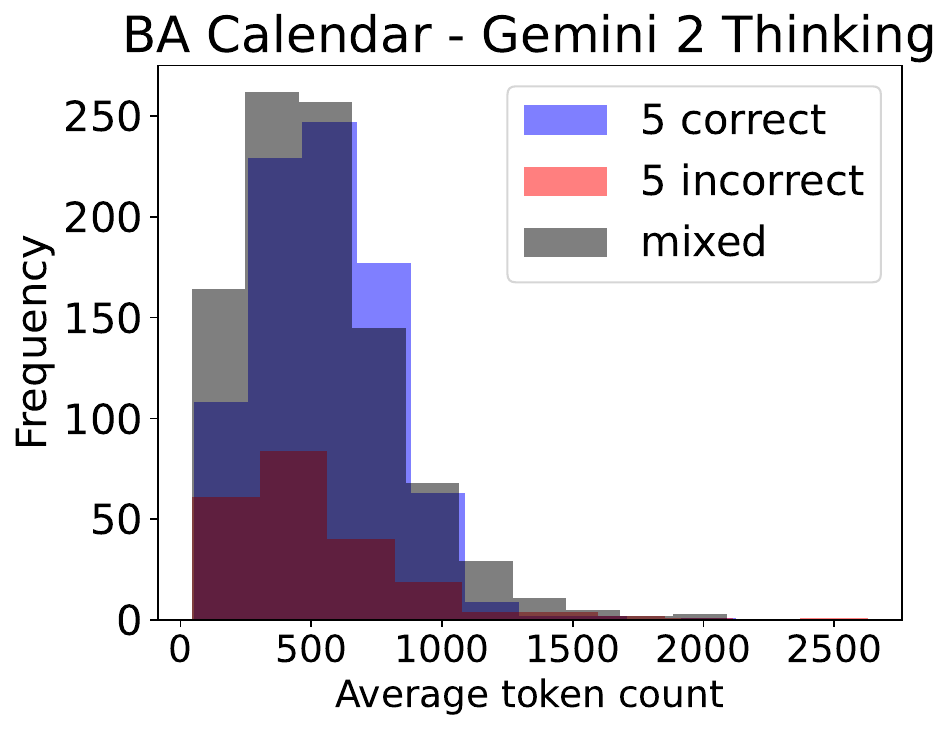}
   \end{subfigure}
   \begin{subfigure}[b]{0.24\textwidth}
       \centering
       \includegraphics[width=\textwidth]{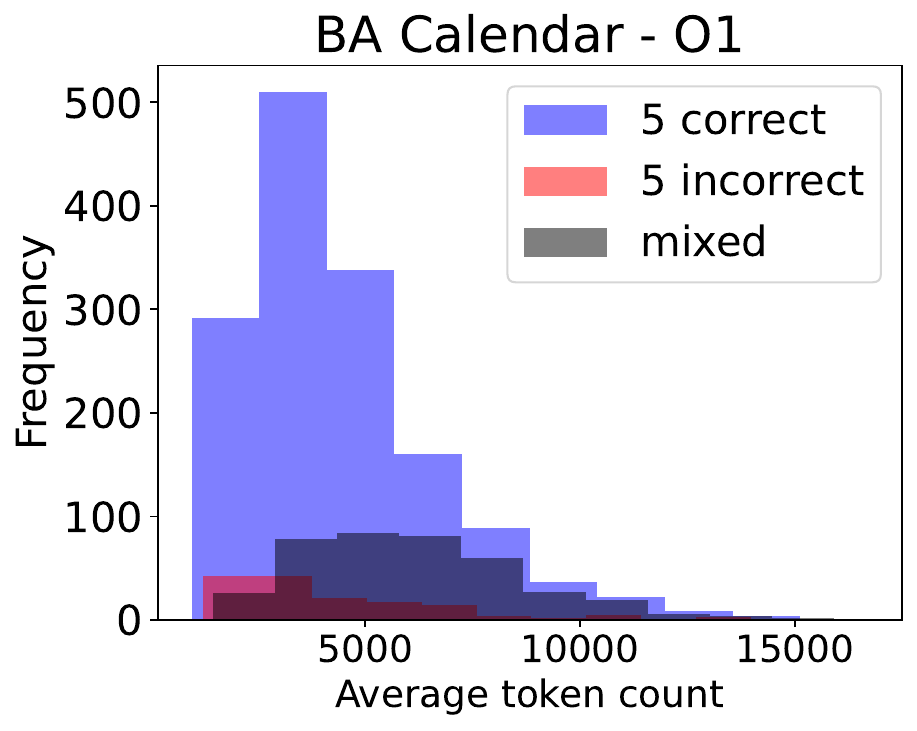}
   \end{subfigure}
    
    \caption{Distributions of average token usage, shown for instances where the models are always correct, always incorrect, or mixed. \OOne 
    has a higher concentration of ``all correct'' instances towards the shorter lengths, while for other models the ``all correct'' instances are more spread out indicating more unpredictability of token usage across instances even when the model is always correct.}
    \label{fig:correct_incorrect_token_usage_mean_appendix}
\end{figure}

\end{document}